\documentclass{winnower}
\usepackage{amsmath}
\usepackage{dsfont}
\usepackage[T1]{fontenc}

\begin{document}

\graphicspath{{figures/}}

\title{Adversarial Examples in Modern Machine Learning: A Review}


\author{Rey Reza Wiyatno \qquad Anqi Xu \qquad Ousmane Dia \qquad Archy de Berker\\
Element AI\\
Montr\'{e}al, Canada\\
{\tt\small \{rey.reza, ax, ousmane, archy\}@elementai.com}
}
\date{}

\maketitle

\begin{abstract} 

Recent research has found that many families of machine learning models are vulnerable to adversarial examples: inputs that are specifically designed to cause the target model to produce erroneous outputs.
In this survey, we focus on machine learning models in the visual domain, where methods for generating and detecting such examples have been most extensively studied.
We explore a variety of adversarial attack methods that apply to image-space content, real world adversarial attacks, adversarial defenses, and the transferability property of adversarial examples.
We also discuss strengths and weaknesses of various methods of adversarial attack and defense.
Our aim is to provide an extensive coverage of the field, furnishing the reader with an intuitive understanding of the mechanics of adversarial attack and defense mechanisms and enlarging the community of researchers studying this fundamental set of problems.

\end{abstract}

\pagebreak

\tableofcontents

\pagebreak


\pagebreak

\section{Introduction} 

Machine learning algorithms have critical roles in an increasing number of domains. Technologies such as autonomous vehicles and language translation systems use machine learning at their core. 
Since the early success of Convolutional Neural Networks (CNNs) on the ImageNet Large Scale Visual Recognition Competition (ILSVRC)~\cite{imagenet_cvpr09,ILSVRC15,Krizhevsky:2012:ICD:2999134.2999257}, deep learning~\citep{LeCun2015,Goodfellow-et-al-2016} has been successfully applied to numerous tasks including image classification~\citep{Krizhevsky:2012:ICD:2999134.2999257,He2016DeepRL,DBLP:journals/corr/HuangLW16a}, segmentation~\citep{Shelhamer:2017:FCN:3069214.3069246,he2017maskrcnn}, object tracking~\citep{held2016learning,bertinetto2016fully,Valmadre_2017_CVPR}, object detection~\citep{girshick2014rcnn,girshick15fastrcnn,NIPS2015_5638}, speech recognition~\citep{6638947,Graves:2014:TES:3044805.3045089,zhang_y+al-2016-interspeech}, language translation~\citep{Sutskever:2014:SSL:2969033.2969173,DBLP:journals/corr/GehringAGYD17,NIPS2017_7181}, and many more. 

Despite the success of modern machine learning techniques in performing various complex tasks, security in machine learning has received far less academic attention. Robustness, both to accident and to malevolent agents, is clearly a crucial determinant of the success of machine learning systems in the real world. For example,~\citet{DBLP:journals/corr/AmodeiOSCSM16} catalogue a variety of possible negative impacts from poorly secured systems, particularly emphasizing issues around sensitive applications such as medical and transportation systems. Although these authors focused on the safety of reinforcement learning algorithms in particular, many of the addressed concerns can directly applied to wide range of machine learning models.

In this review, we focus on the application of adversarial examples to supervised learning problems. We leave it to future authors to cover unsupervised learning, reinforcement learning, or other classes of problems. We also restrict our discussion to classification tasks, as opposed to regression, although many of the methods discussed below may generalize to other problem contexts.

\subsection{What can we learn from adversarial examples?} \label{What can we learn from adversarial examples?} 


The concept of an adversarial example long predates work in machine learning. Humans are famously vulnerable to perceptual illusions, which have been fruitfully used to explore the mechanisms and representations underlying human cognition. Such illusions are frequently used to elucidate the implicit priors present in human perception. For instance, the M\"uller-Lyer illusion~\citep{335346}, in which a line bracketed by two outwards-facing arrows appears longer than one bracketed by outwards facing ones is thought to reveal a ``cubeness'' prior learned by people who live in highly-rectangular environments~\citep{doi:10.1111/j.2044-8295.1978.tb01653.x}. Color and brightness constancy - our ability to perceive colors and brightness as unchanging despite variance in illumination - are richly explored in illusions such as Adelson's checkerboard illusion and Lotto's coloured cube. Such illusions can even probe the inter-individual variance in visual priors: the recently popular \enquote{dress colour} meme is thought to rely upon peoples' differing expectations of illuminating light, with people used to blue-tinted illumination seeing the dress as gold/white, and those assuming warm illumination seeing it as blue/black~\citep{BRAINARD2015R551}. 

Such illusions elucidate not only cognitive phenomena, but details of the underlying neural circuitry (for a thorough review, see~\citet{Eagleman2001VisualIA}). For example, the phenomenon of Inter Ocular transfer has provided a classic paradigm for identifying the locus of illusory effects. Based upon whether an illusory effect transfers from one eye to another, the substrate can be localized to pre-cortical circuitry (no transfer between eyes) or cortical ones, pathways after the mingling of information between the eyes~\citep{BlakemoreCampbell,Blake1981}. Similarly, adaptation-based illusions - where the appearance of an image changes over a period of prolonged viewing - have been used to predict the tuning curves of orientation-tuned neurons in visual cortex~\citep{BlakemoreCampbell}, the organization of colour representation in the lateral geniculate nucleus~\citep{HurvichOpponency}, and the sensitivity of the three colour-coding cell types in the retina~\cite{Stiles}. So effective were these techniques that they earned the nickname ``the psychologists' microelectrode'' - a precise, non-invasive way to characterize the internal workings of a blackbox system.



\begin{figure}[ht]
\centering
\includegraphics[width=0.6\textwidth]{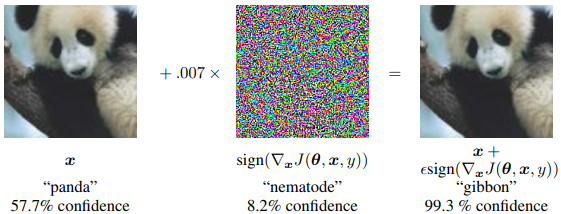}
\caption{Adversarial example generated using the Fast Gradient Sign Method (FGSM) (taken from~\citet{43405}). GoogleNet~\citep{43022} classified the original image shown on the left correctly as panda, but misclassified the image on the right as gibbon with high confidence. Note that the perturbation in the middle has been amplified for visualization purposes only.
}
\label{fig:fgsm_panda}
\end{figure}

\citet{42503} found that deep neural networks are also vulnerable to \enquote{illusions}. These \enquote{adversarial examples} are created by the addition of \enquote{hidden messages} to an image, causing a machine learning model to grossly misclassify the perturbed image. Unlike perceptual illusions in humans, which are handcrafted, such examples are algorithmically generated to fool machine learning models. Figure~\ref{fig:fgsm_panda} shows the now famous adversarial example generated using a method called the Fast Gradient Sign Method (FGSM)~\citep{43405} (see Section~\ref{FGSM}). Judging from these evidences, there is indeed a huge generalisation gap between humans and deep neural networks~\citep{Geirhos:2018:GHD:3327757.3327854}.

The parallel between perceptual illusions in humans and adversarial examples in machines extends beyond the superficial in several important ways. One explanation for the effectiveness of adversarial examples is that they push the input data off the manifold of natural inputs. This is precisely the same mechanism by which perceptual illusions in humans are generated, often exploiting mechanisms that generally improve the fidelity of perception but result in erroneous percepts when input data differs from natural inputs. As such, both approaches help to illustrate which features of the input domain are important for the performance of the system. Secondly, just as perceptual illusions have been used to interrogate the neural organization of perception, adversarial examples can help us understand the computations performed by neural networks. Early explanations focused upon the linearity of deep models~\citep{43405}, but more recent work focuses upon the entropy of the logit outputs~\citep{dogus2018intriguing}. Thirdly, recent evidence suggests a much stronger correspondence, namely that the same examples that fool neural networks also fool humans in time-limited settings~\cite{elsayed2019adversarial}. As observed in that paper, this opens up the interesting possibility that techniques for adversarial defense might draw inspiration from the biological mechanisms that render humans invulnerable to such examples under normal viewing conditions.

Although the recent interest in adversarial attacks is chiefly concerned with their application to deep learning models, the field itself precedes the era of deep learning. For example,~\citet{Huang:2011:AML:2046684.2046692} established a taxonomy of adversarial attacks that has inspired many other works in adversarial machine learning, and also described a case study on adversarial attacks upon SpamBayes~\citep{Robinson:2003:SAS:636750.636753}, a machine learning model for email spam filtering. Other works have also showed empirically that various machine learning models such as logistic regression, decision trees~\citep{DBLP:books/wa/BreimanFOS84}, k-Nearest Neighbor (kNN)~\citep{Cover:2006:NNP:2263261.2267456}, and Support Vector Machines (SVM)~\citep{Cortes:1995:SN:218919.218929} are also vulnerable to adversarial examples~\citep{43405,DBLP:journals/corr/PapernotMGJCS16}.

\subsection{Why this work?}


Since the findings of~\citet{42503}, the arms race between adversarial attacks and defenses has accelerated. For example, defensive distillation~\citep{journals/corr/PapernotMWJS15} (see Section~\ref{DISTIL}), then the state of the art defense against adversarial examples, was defeated within a year by an attack method proposed by~\citet{DBLP:journals/corr/CarliniW16a} (see Section~\ref{CW}). Similarly, defense using adversarial training~\citep{43405} (see Section~\ref{ADV_TRAINING}) once thought to be robust to whitebox attacks~\citep{45816}, was swiftly shown to rely upon a phenomenon called gradient masking~\citep{DBLP:journals/corr/PapernotMGJCS16,8406613,tramèr2018ensemble} which can be circumvented by certain types of attacks. Most recently, a variety of defenses proposed by various different groups~\citep{buckman2018thermometer,ma2018characterizing,s.2018stochastic,xie2018mitigating,samangouei2018defensegan,song2018pixeldefend,guo2018countering} 
that were accepted to the Sixth International Conference on Learning Representations (ICLR) 2018 have been shown to rely on gradient masking, and thus circumventable, shortly after the acceptance decision~\citep{obfuscated-gradients}. This further emphasizes how difficult it is to solve the adversarial examples problem in machine learning.

In this paper, we discuss various adversarial attack and defense strategies. Note that although adversarial examples exist in various domains such as computer security~\citep{10.1007/978-3-319-66399-9_4}, speech recognition~\citep{DBLP:journals/corr/abs-1801-01944,DBLP:journals/corr/abs-1801-00554,DBLP:journals/corr/abs-1801-03339}, and text~\citep{zhao2018generating}, this paper focuses on adversarial examples in the computer vision domain where adversarial examples have been most extensively studied. We particularly focus upon perceptually imperceptible and inconspicuously visible adversarial examples, which have the clearest potential for malicious use in the real world. Although other works have attempted to provide literature review on adversarial examples~\citep{8611298,8294186}, our work provides in-depth explanations of the motivation and mechanism of a wide array of attack and defense algorithms, along with a complete taxonomy and ontology of adversarial attacks and defenses, and a discussion of the strengths and weaknesses of different methods.

This paper is organized as follows. We begin by defining a list of common terms in adversarial machine learning and describing the notation used in this paper, in Section~\ref{COMMON} and~\ref{NOTATIONS}, respectively.
Section~\ref{GENERAL} provides a general introduction to adversarial examples. The taxonomy, ontology, and discussion of adversarial attack and defense methods are defined in Section~\ref{ATTACK} and~\ref{DEFENSE}, respectively. We discuss several adversarial attacks in the real world in Section~\ref{PHYSICAL}, and the transferability property of adversarial examples in Section~\ref{TRANSFER}. Finally, we conclude and suggest interesting lines of future research in Section~\ref{CONCLUSION}.

\pagebreak

\section{Common Terms and Notations} 

\subsection{Common Terms} \label{COMMON}

We provide definitions of several terms that are commonly used in the field of adversarial machine learning in Table~\ref{table:terms}.

\begin{table}[H]
\centering
\caption{Common terms in adversarial machine learning.}
\begin{tabular*}{ 0.92 \textwidth}{ll}
\hline
Common Terms & Definition \\
\hline
Adversarial example & Input to a machine learning model that is intentionally designed to \\ & cause a model to make mistake in its predictions despite resembling \\ & a valid input to a human. \\
\hline
Adversarial perturbation & Difference between a non-adversarial example and its adversarial \\ & counterpart. \\
\hline
Adversarial attacks & Methods to generate adversarial examples. \\ 
\hline
Adversarial defenses & Methods to defend against adversarial examples. \\ 
\hline
Adversarial robustness & The property of resisting misclassification of adversarial examples. \\ 
\hline
Adversarial detection & Methods to detect adversarial examples. \\ 
\hline
Whitebox attack & Attack scenario where an attacker has complete access to the \\ & target model, including the model's architecture and parameters. \\
\hline
Blackbox attack & Attack scenario where an attacker can only observe the outputs of \\ & the targeted model. \\
\hline
Transferability & A property of adversarial examples: \\ & examples specifically generated to fool a model can also be used to \\ & fool other models. \\ 
\hline
Universal attack & Attack scenario where an attacker devises a single transform such \\ &  as image perturbation that adversarially confuses the model for \\ & all or most input values (input-agnostic). \\ 
\hline
Targeted attack & Attack scenario where an attacker wants the adversaries to be \\ & mispredicted in a specific way. \\
\hline
Non-targeted attack & Attack scenario where an attacker does not care about the outcome \\ & as long as the example is mispredicted. \\
\hline
Adversarial training~\citep{43405} & Adversarial defense technique to train a model by including \\ & adversarial examples into the training set. Note that this differs from\\ & the notion of adversarial training used in Generative \\ & Adversarial Networks (GANs)~\citep{NIPS2014_5423}. \\ 
\hline
Gradient masking~\citep{8406613,tramèr2018ensemble} & Defense mechanisms which prevent a model revealing \\ & meaningful gradients, \emph{masking} or hiding the \\ & gradients of the outputs with respect to the inputs. \\
\hline
Shattered gradients~\citep{obfuscated-gradients} & When the gradients of a model are hard to compute\\ & exactly due to non-differentiable operations. \\
\hline
Stochastic gradients~\citep{obfuscated-gradients} & When the gradients of a model are obstructed due to some stochastic \\ & or random operations. \\
\hline
Obfuscated gradients~\citep{obfuscated-gradients} & A form of gradient masking which encompasses shattered gradients, \\ & stochastic gradients, vanishing, and exploding gradients. \\
\hline
Vanishing gradients & When the gradients of a model are converging to zero. \\
\hline
Exploding gradients & When the gradients of a model are diverging to infinity. \\
\hline
\end{tabular*}
\label{table:terms}
\end{table}


\subsection{Notations} \label{NOTATIONS} 


In order to make this paper easier to follow, we use the notations defined in Table~\ref{table:notations} throughout this paper unless otherwise stated.

\begin{table}[H]
\centering
\caption{Notations used in this survey.}
\begin{tabularx}{ 0.9 \textwidth}{lX}
\hline
\textbf{Notation} & \textbf{Description} \\
\hline
$X$ & A set of input data (e.g., training set). \\
\hline
$x$      	& An instance of input data (usually a vector, a matrix, or a tensor). \\
\hline
$x'$      & A modified instance of $x$, typically adversarial.    	\\
\hline
$r$      	& The additive perturbation from $x$ to $x'$, i.e., $r = x' - x$. \\
\hline
$y$      	& The true class label for the input instance $x$. \\
\hline
$t$      & An adversarial target class label for $x'$.  \\
\hline
$f(x;\theta)$ & Output of a machine learning model $f$ parameterized by $\theta$ (e.g., a neural network). For a classifier, $f(x;\theta)$ specifically refers to the softmax predictions vector. Note that $\theta$ is often omitted for notation simplicity. Also, we use the parenthesized subscript $f(x;\theta)_{(c)}$ to denote the $c$-th element of the softmax vector, i.e., the predicted likelihood for class $c$. \\
\hline
$\hat{y}(x)$      & Predicted label of a classifier model, i.e., $\hat{y}(x) = \arg \max_c f(x)_{(c)}$  \\
\hline
$Z(x)$      & Predicted logits vector of $x$ from a softmax classifier model.    		\\
\hline
$\mathcal{L}(x,y)$ & Loss function used to train a machine learning model. Note that we use $\mathcal{L}(x,y)$ instead of $\mathcal{L}(f(x;\theta),y)$ to simplify the notation. \\
\hline
$\nabla_x \mathcal{L}(x,y)$ & Derivative of the loss function $\mathcal{L}(x,y)$ with respect to $x$. \\
\hline
\end{tabularx}
\label{table:notations}
\end{table}

\pagebreak

\section{Adversarial Examples} \label{GENERAL} 

Adversarial examples are typically defined as inputs $x'$, where the differences between $x'$ and non-adversarial inputs $x$ are minimal under a distance metric $d(\cdot)$ (e.g., $d(\cdot)$ can be the $L_p$ distance), whilst fooling the target model $f$. Generally, adversarial examples seek to satisfy

\begin{equation} \label{equations:adversarial}
d(x',x) < \epsilon \ \textrm{such that} \ \hat{y}(x') \neq \hat{y}(x),
\end{equation}

\noindent where $\epsilon$ is a small constant that bounds the magnitude of the perturbations, and $\hat{y}(\cdot)$ denotes the predicted label of a classifier model (i.e., $\hat{y}(x) = \arg \max_c f(x)_{(c)}$). Note that we use $\hat{y}(x)$ to denote $\arg \max_c f(x)_{(c)}$ throughout this paper. However, some works have proposed perturbations that are visible but inconspicuous~\citep{Sharif:2016:ACR:2976749.2978392,DBLP:journals/corr/EvtimovEFKLPRS17,DBLP:journals/corr/abs-1712-09665}, so the similarity constraint between $x$ and $x'$ can be relaxed. Throughout this paper, we call these as \enquote{imperceptible} and \enquote{inconspicuous} (i.e., may be visible but not suspicious) adversarial examples, respectively.

Adversarial attacks are often categorized as either whitebox or blackbox. In whitebox attacks, the attacker is assumed to have information about the target model such as the architecture, parameters, training procedure, or the training data. On the other hand, blackbox attacks involve only access to the output of a target model, and not its internals. Blackbox attacks are more realistic assumption in the real world since an attacker rarely enjoys knowledge of the internals of the victim. However, evaluating models against whitebox attacks is important to measure the performance of the model in the worst-case scenarios. Adversarial attacks can be further categorized as targeted or non-targeted attacks. In a targeted attack, the adversarial example is designed to elicit a specific classification - like classifying all faces as belonging to George Clooney - whilst non-targeted attack only seek to generate an incorrect classification, regardless of class.

Adversarial examples exhibit an interesting phenomenon called the transferability property~\citep{42503,43405,DBLP:journals/corr/PapernotMG16,DBLP:journals/corr/LiuCLS16}. This property states that adversarial examples generated to fool a specific model can often be used to fool other models. This phenomenon will be discussed in Section~\ref{TRANSFER}.

Why are machine learning models vulnerable to these examples? Several works have argued that adversarial examples are effective because they lie in the low probability region of the data manifold~\citep{42503,DBLP:journals/corr/PapernotMG16}.~\citet{43405} pointed that deep neural networks are vulnerable to adversarial examples due to the local linearity property of these models, especially when using activation functions like the Rectified Linear Units (ReLU)~\citep{pmlr-v15-glorot11a} or Maxout~\citep{Goodfellow:2013:MN:3042817.3043084}.~\citet{43405} observed that although deep neural networks use non-linear activation functions, one often trains such networks to only operate in the linear regions of the activation functions to avoid things like the vanishing gradient problem~\citep{Hochreiter:1998:VGP:353515.355233,chapter-gradient-flow-2001,Pascanu:2013:DTR:3042817.3043083}. Furthermore,~\citet{43405} considered the fact that FGSM~\citep{43405} (see Section~\ref{FGSM}) was designed based on the linear assumption works effectively to fool deep neural networks to support their argument that neural networks behave like a linear classifier.~\citet{pmlr-v70-arpit17a} analyzed the capacity of neural networks to memorize training data and found that models with high degree of memorization are more vulnerable to adversarial examples.~\citet{DBLP:journals/corr/abs-1711-11561} argued that convolutional neural networks tend to learn the statistical regularities in the dataset rather than the high level abstract concepts.
This may be related to the transferability property; since adversarial examples are transferable between models that are trained on the \emph{same dataset}, these different models may have learned the same statistics and hence fall into the same traps. Similarly,~\citet{ilyas2019adversarial} suggested that adversarial examples exist as a byproduct of exploiting \textit{non-robust} features that exist in a dataset. Up to now, the reasons why machine learning models are vulnerable to adversarial examples are still an open research area.

\pagebreak

\section{Adversarial Attacks} \label{ATTACK} 

\begin{figure}[H]
\centering
\includegraphics[width=1.15\textwidth, angle=90]{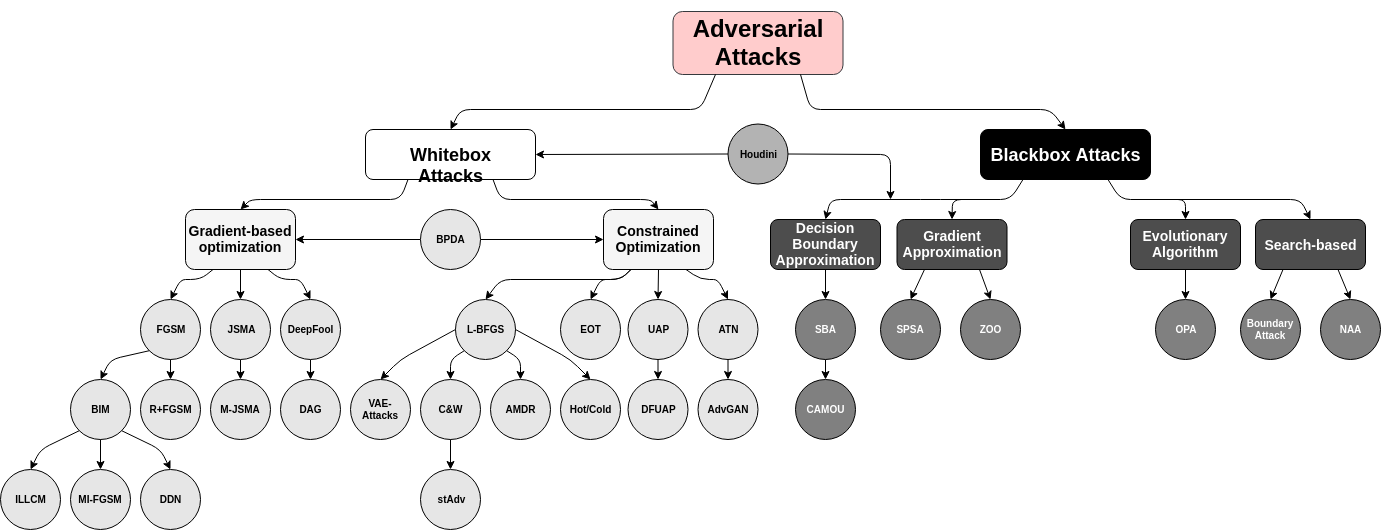}
\caption{Ontology of adversarial attacks covered in this paper.}
\label{fig:attack_ontology}
\end{figure}

In this section, we survey the adversarial attacks that exist today. Figure~\ref{fig:attack_ontology} and Table~\ref{table:attack_taxonomy} illustrate the ontology and taxonomy of adversarial attack techniques discussed in this paper, respectively.

\begin{sidewaystable}[!ph]
\centering
\caption{Taxonomy of adversarial attacks covered in this paper.}
\begin{tabular*}{ 1.0 \textwidth}{llll}
\hline
\textbf{Adversarial Attack(s)} & \textbf{Transparency} & \textbf{Specificity} & \textbf{Remarks} \\
\hline
L-BFGS~\citep{42503} & W & T, NT & Early attack on neural networks using constrained optimization method\\
\hline
FGSM~\citep{43405} & W & T, NT & A fast single-step gradient ascent attack \\
\hline
BIM~\citep{DBLP:journals/corr/KurakinGB16,45816} & W & T, NT & Iterative variants of FGSM \\
\hline
ILLCM~\citep{DBLP:journals/corr/KurakinGB16,45816} & W & T & Extension of BIM to attack models with many output classes \\
\hline
R+FGSM~\citep{tramèr2018ensemble} & W & T, NT & FGSM~\citep{43405} with random initialization, can circumvent gradient masking \\
\hline
AMDR~\citep{DBLP:journals/corr/SabourCFF15} & W & T, NT & Similar to L-BFGS but targetting feature space \\
\hline
DeepFool~\citep{DBLP:journals/corr/Moosavi-Dezfooli15} & W & NT & Efficient method to find minimal perturbation that causes misclassification \\
\hline
JSMA~\citep{DBLP:journals/corr/PapernotMJFCS15} & W & T, NT & Some variants of JSMA can fool defensive distillation \\
\hline
SBA~\citep{DBLP:journals/corr/PapernotMGJCS16} & B & T, NT & Can fool defensive distillation~\citep{journals/corr/PapernotMWJS15}, MagNet~\citep{Meng:2017:MTD:3133956.3134057}, gradient masking defenses \\
\hline
Hot/Cold~\citep{DBLP:journals/corr/RozsaRB16} & W & T & Simultaneously moving towards ``hot'' class and away from ``cold'' class \\
\hline
C\&W~\citep{DBLP:journals/corr/CarliniW16a} & W & T, NT & Can fool defensive distillation~\citep{journals/corr/PapernotMWJS15}, MagNet~\citep{Meng:2017:MTD:3133956.3134057} and various detector networks
\\
\hline
UAP~\citep{DBLP:journals/corr/Moosavi-Dezfooli16} & W & NT & Generate input-agnostic perturbations \\
\hline
DFUAP~\citep{gduap-mopuri-2018} & W & NT & Generate input-agnostic perturbations without knowing any inputs \\
\hline
VAE Attacks~\citep{DBLP:journals/corr/KosFS17} & W & T, NT & Can fool VAE~\citep{DBLP:journals/corr/KingmaW13} and potentially defenses relying on generative models \\
\hline
ATN~\citep{46527} & W & T, NT & Generate adversarial examples using neural networks \\
\hline
DAG~\citep{DBLP:journals/corr/XieWZZXY17} & W & T, NT & Can fool semantic segmentation \& object detection Models \\
\hline
ZOO~\citep{Chen:2017:ZZO:3128572.3140448} & B & T, NT & Can fool defensive distillation~\citep{journals/corr/PapernotMWJS15} and non-differentiable models \\
\hline
OPA~\citep{DBLP:journals/corr/abs-1710-08864} & B & T, NT & Uses genetic algorithm, can generate adversary by just modifying one pixel \\
\hline
Houdini~\citep{DBLP:conf/nips/CisseANK17} & W, B & T, NT & Method for attacking models directly through its non-differentiable metric \\
\hline
MI-FGSM~\citep{Dong_2018_CVPR} & W & T, NT & BIM + momentum, faster to converge and better transferability \\
\hline
AdvGAN~\citep{xiao2018generating} & W & T, NT & Generate adversarial examples using GAN~\citep{NIPS2014_5423} \\
\hline
Boundary Attack~\citep{brendel2018decisionbased} & B & T, NT & Can fool defensive distillation~\citep{journals/corr/PapernotMWJS15} and non-differentiable models \\
\hline
NAA~\citep{zhao2018generating} & B & NT & Can generate adversaries for non-sensory inputs such as text \\
\hline
stAdv~\citep{xiao2018spatially} & W & T, NT & Unique perceptual similarity objective \\
\hline
EOT~\citep{pmlr-v80-athalye18b} & W & T, NT & Good for creating physical adversaries and fooling randomization defenses \\
\hline
BPDA~\citep{obfuscated-gradients} & W & T, NT & Can fool various gradient masking defenses \\
\hline
SPSA~\citep{pmlr-v80-uesato18a} & B & T, NT & Can fool various gradient masking defenses \\
\hline
DDN~\citep{DBLP:journals/corr/abs-1811-09600} & W & T, NT & Better convergence compared to other constrained optimization methods \\
\hline
CAMOU~\citep{zhang2018camou} & B & NT & Attack in simulation using SBA~\citep{DBLP:journals/corr/PapernotMGJCS16}, can be used to attack detection model \\
\hline
\end{tabular*}
\footnotetext{W: Whitebox}
\footnotetext{B: Blackbox}
\footnotetext{T: Targeted}
\footnotetext{NT: Non-targeted}
\label{table:attack_taxonomy}
\end{sidewaystable}

\subsection{L-BFGS Attack} \label{L-BFGS} 

The L-BFGS attack~\citep{42503} is an early method designed to fool models such as deep neural networks for image recognition tasks. Its end goal is to find a perceptually-minimal input perturbation $\arg \min_{r} ||r||_2$, i.e., $r = x' - x$, within bounds of the input space, that is adversarial, i.e., $\hat{y}(x') \neq y$.~\citet{42503} used the Limited Memory Broyden-Fletcher-Goldfarb-Shanno (L-BFGS) algorithm~\citep{Liu1989} to transform this difficult optimization problem into a box-constrained formulation where the goal is to find $x'$ that minimizes

\begin{equation} \label{equations:lbfgs}
c||r||_2 + \mathcal{L}(x',t) \ \textrm{such that} \ x' \in [0,1],
\end{equation}

\noindent where elements of $x$ are normalized to $[0, 1]$, $\mathcal{L}(x',t)$ is the true loss function of the targeted model (e.g., categorical cross-entropy), and $t$ is the target misclassification label. Since this objective does not guarantee that $x'$ will be adversarial for any specific value of $c > 0$, the above optimization process is iterated for increasingly large values of $c$ via line search until an adversary is found. Optionally, the resulting $c$ value can be further optimized using the bisection method (a.k.a. binary search) between the range of the final line search segment.

This attack was successfully applied to misclassify many image instances on both AlexNet~\citep{Krizhevsky:2012:ICD:2999134.2999257} and QuocNet~\citep{Le:2012:BHF:3042573.3042641}, which were state-of-the-art classification models at the time. Thanks to its $L_2$ Euclidean distance constraint, L-BFGS produces adversaries that are perceptually similar to the original input $x$. Moreover, a key advantage of modeling adversarial examples generation process as a general optimization problem is that it allows for flexibility in folding additional criteria into the objective function. For instance, one may choose to use perceptual similarity metrics other than the $L_2$ distance, depending on requirements of a given application domain. We will see concrete examples of such criteria in subsequent sections.

\subsection{Fast Gradient Sign Method} \label{FGSM} 

The Fast Gradient Sign Method (FGSM)~\citep{43405} is designed to quickly find a perturbation direction for a given input such that the training loss function of the target model will increase, reducing classification confidence and increasing the likelihood of inter-class confusion. While there is no guarantee that increasing the training loss by a given amount will result in misclassification, this is nevertheless a sensible direction to take since the loss value for a misclassified instance is by definition larger than otherwise.

FGSM works by calculating the gradient of the loss function with respect to the input, and creating a small perturbation by multiplying a small chosen constant $\epsilon$ by the sign vector of the gradient:

\begin{equation} \label{equations:fgsm}
x' =  x + \epsilon \cdot \textrm{sign}(\nabla _x \mathcal{L}(x,y)),
\end{equation}

\noindent where $\nabla _x \mathcal{L}(x,y)$ is the first derivative of the loss function with respect to the input $x$. In the case of deep neural networks, this can be calculated through the backpropagation algorithm~\citep{Rumelhart1986}. In practice, the generated adversarial examples must be within the bounds of the input space (e.g., [0, 255] pixel intensities for an 8-bit image), which is enforced by value-clipping.

The authors proposed to bound the input perturbation $r$ under the $L_{\infty}$ supremum metric (i.e., $||r||_{\infty} \leq \epsilon$) to encourage perceptual similarity between $x$ and $x'$. Under this $\infty$-norm constraint, the sign of the gradient vector maximizes the magnitude of the input perturbation, which consequently also amplifies the adversarial change in the model's output. A variant of FGSM that uses the actual gradient vector rather than its sign vector was later introduced as the Fast Gradient Value (FGV) method~\citep{DBLP:journals/corr/RozsaRB16}.

A sample adversarial example and perturbation generated by FGSM can be seen in Fig.~\ref{fig:fgsm_panda}. Note that this attack can be applied to any machine learning model where $\nabla _x \mathcal{L}(x,y)$ can be calculated. Compared to the numerically-optimized L-BFGS attack, FGSM computes gradients analytically and thus finds solutions much faster. On the other hand, FGSM does not explicitly optimize for the adversary $x'$ to have a minimal perceptual difference, instead using a small $\epsilon$ to weakly bound the perturbation $r$. Optionally, once an adversarial example is found at a given $\epsilon$ value, one can use an iterative strategy similar to the L-BFGS attack's line search of $c$ to further enhance perceptual similarity, although the resulting $r$ may still not have minimal perceptual difference since perturbations are only searched along the sign vector of the gradient.

\citet{DBLP:journals/corr/HuangPGDA17} have also shown that FGSM can be used to attack reinforcement learning models~\citep{Sutton:1998:IRL:551283} where the policies are parameterized using neural networks such as the Trust Region Policy Optimization (TRPO)~\citep{pmlr-v37-schulman15}, Asynchronous Advantage Actor-Critic (A3C)~\citep{pmlr-v48-mniha16}, and Deep Q-Network (DQN)~\citep{mnih-dqn-2015}. By applying the FGSM to modify images from various Atari games (e.g., Pong, Chopper Command, Seaquest, Space Invaders)~\citep{Bellemare:2015:ALE:2832747.2832830}, the agent can be fooled into taking sub-optimal actions.


\subsection{Basic Iterative Method} \label{BIM} 

The Basic Iterative Method (BIM)~\citep{DBLP:journals/corr/KurakinGB16, 45816} is one of many extensions of FGSM~\citep{43405} (see Section~\ref{FGSM}), and is sometimes referred to as Iterative FGSM or I-FGSM. BIM applies FGSM multiple times within a supremum-norm bound on the total input perturbation, $||r||_{\infty} \leq \epsilon$. The adversarial examples generated by BIM are defined as

\begin{equation} \label{equations:ifgsm}
x'_{i+1} =  Clip_{\epsilon} \Big\{ x'_i + \alpha \cdot \textrm{sign}(\nabla _x \mathcal{L}(x'_i,y)) \Big\} \ \textrm{for} \ i = 0 \ \textrm{to} \ n, \ \textrm{and} \ x'_0 =  x
\end{equation}

\noindent where $n$ is the total number of iterations and $0 < \alpha < \epsilon$ is the per-iteration step size. The clipping operator $Clip \{\cdot \}$ constrains each input feature (e.g., pixel) at coordinate $(u,v,w)$ to be within an $\epsilon$-neighborhood of the original instance $x$, as well as within the feasible input space (e.g., $[0, 255]$ for 8-bit intensity values):

\begin{equation} \label{equations:clip}
Clip_{\epsilon} \{x'_{i,(u,v,w)} \} =  \textrm{min} \Big\{ 255, x_{(u,v,w)} + \epsilon, \textrm{max} \Big\{ 0,x_{(u,v,w)} - \epsilon,x'_{i,(u,v,w)} \Big\} \Big\}\
\end{equation}

BIM was the first method shown to be effective with printed paper examples (see Section~\ref{PHYSICAL}). Beyond classification models, BIM has also been used to attack semantic segmentation models~\citep{DBLP:journals/corr/FischerKMB17}, such as the FCN~\citep{Shelhamer:2017:FCN:3069214.3069246}.

\subsection{Iterative Least-Likely Class Method} \label{ILLCM} 

The authors of BIM also proposed a targeted variant of the attack called the Iterative Least-Likely Class Method (ILLCM), where the goal is to generate an adversarial example which is misclassified as a specific target class $t$~\cite{DBLP:journals/corr/KurakinGB16,45816}. In fact, ILLCM targets the class with the \emph{least likelihood} of being chosen by the original classifier, i.e., $t = \arg \min f(x)$. The corresponding iterative update is given as:

\begin{equation} \label{equations:illcm}
x'_{i+1} =  Clip_{\epsilon} \Big\{ x'_i - \alpha \cdot \textrm{sign}(\nabla _x \mathcal{L}(x'_i,t)) \Big\} \ \textrm{for} \ i = 0 \ \textrm{to} \ n, \ \textrm{and} \ x'_0 =  x.
\end{equation}

\noindent This update is nearly identical to Equation~\ref{equations:ifgsm}, except that the predicted class in the cross-entropy loss is changed from the true label $y$ to an adversarial target $t \neq y$, and the sign of the gradient update is reversed. Thus, whilst the non-targeted BIM and FGSM attacks \emph{increase} the original classifier's training loss, effectively \enquote{undoing training} and encouraging inter-class confusion, the targeted ILLCM \emph{reduces} classification loss of an \emph{adversarial training} pair $(x, t)$ to misguide the model into having excessive confidence towards the target class $t$. 

Why target the least likely class, $t = \arg \min f(x)$? Doing so maximizes misclassification robustness, preventing the model finding trivial adversarial examples for which the true classes are very similar. This is particularly relevant when working with large datasets such as ImageNet~\cite{imagenet_cvpr09,ILSVRC15}, which contain many similar-looking classes. For instance, ImageNet includes examples of both Siberian Huskies and Alaskan Malamutes (both wolflike dogs). Non-targeted attacks risk finding trivial adversarial examples which cause confusion between the two, which is both easy to achieve and relatively benign. In contrast, ILLCM aims to maximize the negative impact of an attack through dramatic misclassification (e.g., dog $\rightarrow$ airplane), while minimizing input perturbations.

\subsection{R+FGSM} \label{R+FGSM} 

The randomized single-step attack, or R+FGSM~\citep{tramèr2018ensemble}, adds a small random perturbation to the input before applying the adversarial perturbation generated by FGSM. This helps avoid the defensive strategy of gradient masking~\citep{DBLP:journals/corr/PapernotMGJCS16, 8406613,tramèr2018ensemble}, which is illustrated in Fig.~\ref{fig:gradient_masking}. 

\citet{tramèr2018ensemble} discussed how adversarially training~\citep{43405} (see Section~\ref{ADV_TRAINING}) a model on FGSM~\citep{43405} adversaries can effectively lead to gradient masking.~\citet{tramèr2018ensemble} showed that there are many orthogonal adversarial directions, and that local loss gradients do not necessarily translate to the direction where the global loss of a model will be maximum. This led to some complacency in the machine learning community, in the mistaken belief that adversarially trained models are robust to unseen adversarial examples.

\def\fs{0.6}
\begin{figure}[ht]
   \begin{minipage}{0.5\linewidth}
   \centering
   \subfloat[Loss surface of an adversarially trained model]{\label{gradient_masking:a}\includegraphics[scale=\fs]{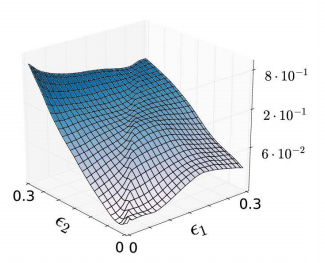}}
   \end{minipage}%
   \begin{minipage}{0.5\linewidth}
   \centering
   \subfloat[Same loss surface zoomed in for small $\epsilon$]{\label{gradient_masking:b}\includegraphics[scale=\fs]{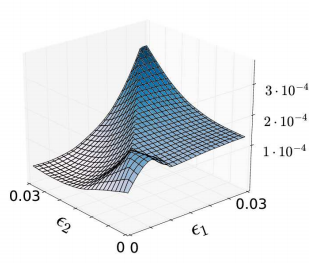}}
   \end{minipage}
   \caption{Illustration of a specific instance of gradient masking for an adversarially trained model (taken from~\citet{tramèr2018ensemble}). Starting with values for model parameters $\epsilon_1$ and $\epsilon_2$ near the center of Fig.~\ref{gradient_masking:b}, the gradient locally points towards smaller values for both parameters, while we see that on a wider scale the loss can be maximized (within an adversarial attack process) for larger values of $\epsilon_1$ and $\epsilon_2$.}
   \label{fig:gradient_masking}
\end{figure}

In order to escape from this particular case of gradient masking, R+FGSM first modifies the input into $x^\dagger$ by pushing it towards a random direction (which is sampled as the sign of an unit multi-variate Gaussian), and then calculates the derivative of the loss with respect to $x^\dagger$:

\begin{equation} \label{equations:rfgsm}
x' =  x^\dagger + (\epsilon-\alpha)\cdot \textrm{sign}(\nabla _{x^\dagger} \mathcal{L}(x^\dagger,y)), \textrm{where} \ x^\dagger = x + \alpha \cdot \textrm{sign} (\mathcal{N}(0,1)).
\end{equation}

\noindent Here, $\alpha$ is a positive constant hyperparameter such that $0<\alpha<\epsilon$. While the adversarial gradient now contributes less perturbation magnitude than in FGSM (i.e., $(\epsilon - \alpha)$ instead of $\epsilon$), the random pre-perturbation increases the chance of finding an attack direction that escapes gradient masking.

\citet{tramèr2018ensemble} found that R+FGSM has a higher attack success rate compared to FGSM on adversarially trained Inception ResNet v2~\citep{45169} and Inception v3 models~\citep{44903}. Furthermore R+FGSM was also found to be stronger compared to 2-steps BIM~\citep{DBLP:journals/corr/KurakinGB16, 45816} (see Section~\ref{BIM}) on adversarially trained Inception v3 model, which suggests that random sampling helps provides better directions than using the local gradients. This strengthens the suspicion of gradient masking as random guesses should not produce better adversarial directions than using the actual gradients.

\subsection{Adversarial Manipulation of Deep Representations} \label{AMDR} 

The approaches discussed so far optimize for properties of the output softmax or logits. \citet{DBLP:journals/corr/SabourCFF15} proposed an altered formulation of the L-BFGS attack~\citep{42503} that instead generates examples which resemble the target class in terms of their hidden layer activations. Adversarial Manipulation of Deep Representations (AMDR) thus optimizes for similarity between a perturbed source image and a target image with different class label in the \textit{intermediary} layers of a network, rather than the \textit{output} layers.

Formally, AMDR requires a target neural network classifier $f$, a source image $x$, and a target image of a different class, $x_t$, where $f(x_t) \neq f(x)$. Given the above, the goal is to find an adversarial example $x'$ that looks like $x$, yet whose internal representations resembles those of of $x_t$. The resulting optimization objective is thus:

\begin{equation} \label{equations:feature_optimization}
\arg \min_{x'} ||f_l(x') - f_l(x_t)||_2^2 ~~~ \textrm{such that} \ ||x' - x||_{\infty}<\epsilon,
\end{equation}

\noindent where $f_l(\cdot)$ denotes the output of $f$ at the $l$-th layer, i.e., the internal representations of the input, and $\epsilon$ denotes the bound for the $L_{\infty}$ norm. Contrary to the L-BFGS attack, this method does not require a target label, but rather needs a target image $x_t$ and a chosen feature layer $l$.

As shown by~\citet{DBLP:journals/corr/SabourCFF15}, the AMDR attack successfully finds adversaries when evaluated against CaffeNet~\citep{Jia:2014:CCA:2647868.2654889}, AlexNet~\citep{Krizhevsky:2012:ICD:2999134.2999257}, GoogleNet~\citep{43022}, and a VGG variant~\citep{Chatfield14} on ImageNet~\citep{imagenet_cvpr09,ILSVRC15} and Places205~\citep{Zhou:2014:LDF:2968826.2968881} datasets. The authors also qualitatively evaluated the internal representations by inverting them back into an image using the technique proposed by~\cite{mahendran15understanding} and found that the inverted images resembled the target images.

\subsection{DeepFool} \label{DEEPFOOL} 

The DeepFool algorithm~\cite{DBLP:journals/corr/Moosavi-Dezfooli15} estimates the distance of an input instance $x$ to the closest decision boundary of a multi-class classifier. This result can be used both as a measure of the robustness of the model to attacks, and as a minimal adversarial perturbation direction. As motivation, the authors note that for a binary linear classifier, the distance to the decision boundary (which is simply a line) can be analytically computed using the point-to-line distance formula. This readily generalizes to a multi-class linear classifier, where the desired measure can be computed as the distance to the \emph{nearest} of the decision boundary lines among the classes $c$.

Generalizing further to non-linear multi-class neural networks, DeepFool iteratively perturbs the input $x$ by linearizing the model's per-class decision boundaries around the current set-point $x'_i$ (starting from $x'_0 = x$), identifies the class $j$ with the closest linearized decision boundary, and moves $x'_i$ to this estimated boundary point. As shown in Algorithm~\ref{alg:deepfool}, this process is repeated till $f(x'_i)$ becomes misclassified.\,\footnote{In the original formulation~\cite{DBLP:journals/corr/Moosavi-Dezfooli15}, the iterative algorithm terminates when $\hat{y}(x'_i) \neq \hat{y}(x)$, while our variant terminates on $\hat{y}(x'_i) \neq y$. These are identical when the classifier's prediction $\hat{y}(x)$ of a given input $x$ correctly reflects the ground truth label $y$. Otherwise, $x$ does not need to be perturbed if the original model already misclassifies it, and perturbing it to the nearest decision boundary might actually \emph{correct} the misclassification.} Recall that $f(x)_{(c)}$ denotes the $c$-th element of the softmax prediction vector $f(x)$, while $\odot$ denotes element-wise multiplication.

\begin{algorithm}[h]
   \caption{DeepFool Algorithm for Multi-Class Classifier~\citep{DBLP:journals/corr/Moosavi-Dezfooli15}}
   \label{alg:deepfool}
\begin{algorithmic}
   \STATE {\bfseries Input:} input $x$, ground truth label $y$, number of classes $N$, classifier $f$, desired norm $p$, let $q = \frac{p}{p-1}$
   \STATE {\bfseries Output:} adversarial perturbation $r$
   \STATE {\bfseries Initialize:} $x'_0 \leftarrow x$, $r \leftarrow \{0,0,...\}$, $i \leftarrow 0$
   \WHILE{$\hat{y}(x'_i) == \hat{y}(x)$}
   \FOR{$c=0$ \TO $N$}
   \IF{$c \neq y$}
   \STATE $w'_c \leftarrow \nabla_{x'_i} f(x_i')_{(c)} - \nabla_{x'_i} f(x_i')_{(y)}$
   \STATE $f'_c \leftarrow f(x'_i)_{(c)} - f(x'_i)_{(y)}$
   \ENDIF
   \ENDFOR
   \STATE $j \leftarrow \arg \min_{c \neq y} \frac{|f'_c|}{||w'_c||_q}$
   \STATE $r_i \leftarrow \frac{|f'_j|}{||w'_j||_q^q} \cdot \textrm{sign} (w'_j) \odot |w'_h|^{q-1}$
   \STATE $x'_{i+1} \leftarrow x'_i + r_i$
   \STATE $i \leftarrow i + 1$
   \ENDWHILE
   \STATE {\bfseries Return:} $r = \sum_i r_i$
\end{algorithmic}
\end{algorithm}

The size of the resulting perturbation can be interpreted as a measure of the model's robustness to adversarial attacks. DeepFool can compute this measure using a variety of different $L_p$ distance metrics including $p=2$ Euclidean norm and $p=\infty$ supremum norm. In practice, once an adversarial perturbation $r$ is found, the adversarial example is nudged further beyond the decision boundary to guarantee misclassification, e.g., $x' = x + 1.02r$.

DeepFool has been successful in attacking various models such as LeNet~\citep{726791}, Network in Network~\citep{DBLP:journals/corr/LinCY13}, CaffeNet~\citep{Jia:2014:CCA:2647868.2654889}, and GoogLeNet~\citep{43022}. Furthermore,~\cite{DBLP:journals/corr/Moosavi-Dezfooli15} found that DeepFool generates adversaries that have $5$ times smaller perturbations compared to those resulting from FGSM~\citep{43405} (see Section~\ref{FGSM}) on MNIST and CIFAR10 models, and $10$ times smaller perturbations for ImageNet models. DeepFool was also found to produce adversaries with slightly smaller perturbations compared to the L-BFGS attack~\citep{42503} (see Section~\ref{L-BFGS}) while being much faster (e.g., more than $30$x speedup) to compute.

\subsection{Jacobian-based Saliency Map Attacks} \label{JSMA} 


The notion of the saliency map was originally conceived for visualizing how deep neural networks make predictions~\citep{DBLP:journals/corr/SimonyanVZ13}. The saliency map rates each input feature (e.g., each pixel in an image) by its influence upon the network's class prediction. Jacobian-based Saliency Map Attacks (JSMA)~\cite{DBLP:journals/corr/PapernotMJFCS15} exploit this information by perturbing a small set of input features to cause misclassification. This is in contrast to attacks like the FGSM~\citep{43405} (see Section~\ref{FGSM}) that modify most, if not all, input features. As such, JSMA attacks tend to find sparse perturbations.

Given the predicted softmax probabilities vector $f(x)$ from a neural network classifier, one formulation of the saliency map $S^+$ is

\begin{equation} \label{equations:jsma_increase}
  S^+(x_{(i)},t) = 
  \begin{cases}
  0 & \text{if} \ \nabla_{x_{(i)}}f(x)_{(t)} < 0 \ \text{or} \ \sum_{c \neq t} \nabla_{x_{(i)}}f(x)_{(c)} > 0 \\
  - \nabla_{x_{(i)}}f(x)_{(t)} \cdot \sum_{c \neq t} \nabla_{x_{(i)}}f(x)_{(c)} & \text{otherwise},
  \end{cases}
\end{equation}

\noindent where $x_{(i)}$ denotes the $i$-th element of $x$, and $t$ is a specified label of interest, e.g., the target for visualization or attack. Intuitively, the $S^+$ saliency map uses components of the gradient $\nabla_{x_{(i)}}f(x)$ to quantify the degree to which each input feature $x_{(i)}$ \emph{positively} correlates with a target class of interest $t$, while on average \emph{negatively} correlating with all other classes $c \neq t$. If either condition is violated for a given feature $x_{(i)}$, then $S(x_{(i)}, t)$ is set to zero, effectively ignoring features which are not preferentially associated with the target class. The features with the largest saliency measure can then be increased to amplify the model's predicted confidence for a target class $t$, whilst attenuating confidences for all other classes. 

This process can easily be inverted to provide a negative saliency map, $S^-$, describing which features should be reduced to increase a target class probability. This formulation requires inverting the inequalities for the two low-saliency conditions:

\begin{equation} \label{equations:jsma_decrease}
  S^-(x_{(i)},t) = 
  \begin{cases}
  0 & \text{if} \ \nabla_{x_{(i)}}f(x)_{(t)} > 0 \ \text{or} \ \sum_{c \neq t} \nabla_{x_{(i)}}f(x)_{(c)} < 0 \\
  - \nabla_{x_{(i)}}f(x)_{(t)} \cdot \sum_{c \neq t} \nabla_{x_{(i)}}f(x)_{(c)} & \text{otherwise}.
  \end{cases}
\end{equation}

\begin{algorithm}[h]
   \caption{Jacobian-based Saliency Map Attack by Increasing Pixel Intensities~\citep{DBLP:journals/corr/PapernotMJFCS15}}
   \label{alg:jsma}
\begin{algorithmic}
   \STATE {\bfseries Input:} $n$-dimensional input $x$ normalized to $[0,1]$, target class $t$, classifier $f$, maximum number of iterations $I_{max}$, perturbation step $\theta = +1$
   \STATE {\bfseries Initialize:} $x' \leftarrow x$, $i \leftarrow 0$, search domain $\Gamma \leftarrow \{1,...,n\}$
   \WHILE{$\hat{y}(x') \neq t$ \AND $i < I_{max}$ \AND $|\Gamma| \geq 2$ }
   \STATE Calculate $\nabla_{x'} f(x')$
   \STATE $\gamma \leftarrow 0$
   \FOR{every pixel pair ($p,q$) $\in \Gamma$}
   \STATE $\alpha = \sum_{k = p,q} \nabla_{x'_{(k)}} f(x')_{(t)}$
   \STATE $\beta = \sum_{k = p,q}\sum_{c \neq t} \nabla_{x'_{(k)}} f(x')_{(c)}$
   \IF {$\alpha>0 \ \AND \ \beta<0 \ \AND \ -\alpha \cdot \beta > \gamma$}
   \STATE $p^*,q^* \leftarrow p,q$
   \STATE $\gamma \leftarrow -\alpha \cdot \beta$
   \ENDIF
   \ENDFOR
   \STATE $x'_{(p^*)}, x'_{(q^*)} \leftarrow (x'_{(p^*)} + \theta), (x'_{(q^*)} + \theta)$
   \IF{$x'_{(p^*)}==0 \ \OR \ x'_{(p^*)}==1$}
   \STATE $\text{Remove} \ p^* \ \text{from} \ \Gamma$
   \ENDIF
   \IF{$x'_{(q^*)}==0 \ \OR \ x'_{(q^*)}==1$}
   \STATE $\text{Remove} \ q^* \ \text{from} \ \Gamma$
   \ENDIF
   \STATE $i \leftarrow i + 1$
   \ENDWHILE
   \STATE {\bfseries Return:} $x'$
\end{algorithmic}
\end{algorithm}

\citet{DBLP:journals/corr/PapernotMJFCS15} notes that both saliency measures $S^+$ and $S^-$ are overly strict when applied to individual input features (e.g., single image pixels), since it is likely that the sum of gradient contributions across non-targeted classes will trigger the minimal-saliency criterion. Consequently, as shown in Algorithm~\ref{alg:jsma}, the Jacobian-based Saliency Map Attack alters the saliency measures to search over \emph{pairs} of pixels $(p, q)$ instead. Concretely, given a search domain $\Gamma$ initialized to contain the indices of all input features, the algorithm finds the most salient pixel pair, perturbs both values by $\theta = +1$, and then removes saturated feature indices from the search domain. This process is repeated until either an adversary is found, or in practice following a maximum number of iteration $I_{max}$, e.g.:

\begin{equation} \label{equations:jsma_iter}
I_{max} = \frac{(\#~\textrm{of input features}) \cdot (\textrm{maximum desired percentage of perturbed features})}{(\#~\textrm{of features to be modified per iteration}) \cdot \lceil \theta^{-1} \rceil \cdot 100},
\end{equation}

The formulation in Algorithm~\ref{alg:jsma} finds adversarial examples by \textit{increasing} feature values ($\theta > 0$) based on the $S^+$ saliency measure. An alternative attack variant that \emph{decreases} feature values can be constructed by substituting $S^+$ with $S^-$ and setting $-1 \leq \theta < 0$. Both variants are targeted attacks that increase a classifier's softmax prediction confidence $f(x)_{(t)}$ for a chosen adversarial target class $t$.

The original authors prescribed $\theta=1$ in order to find adversaries in as few iterations as possible. In general though, we can use a smaller feature perturbation step, i.e., $0 < \theta \leq 1$, to produce adversarial examples with fewer saturated features. Additionally, the feature saturation criterion can be also altered to be $\epsilon$-bounded around the initial input values $x_i$ (e.g., using $Clip\{ \cdot \}$ from Equation~\ref{equations:clip}), to further constrain per-pixel perceptual difference.

\citet{DBLP:journals/corr/CarliniW16a} note that the above saliency measures and attack can be alternatively applied to evaluate the gradient of logits $Z(x)$ rather than of the softmax probabilities $f(x)$. While using different saliency measures results in favoring slightly different pixel pairs, both variants successfully find adversarial examples. We designate the original algorithm variants as JSMA+F and JSMA-F, and those using logit-based saliency maps as JSMA+Z and JSMA-Z, where + and - indicate whether input features are increased or decreased.

\citet{DBLP:journals/corr/PapernotMJFCS15} showed that JSMA can successfully fool a model by just modifying a few input features. They found that adversaries can be found by just modifying $4\%$ of the input features in order to fool a targeted MNIST model. However, there are still room for improving the misclassification rate and efficiency by picking which features should be updated in a more optimal way. For example, note that Algorithm~\ref{alg:jsma} needs to test every possible pixel pairs in the search domain before deciding on which pixel pairs should be updated for every iteration, which is computationally expensive to perform.

All JSMA variants above must be given a specific target class $t$. This choice affects the speed and quality of the attack, since misclassification under certain classes are easier to attain than others, such as perturbing a hand-written digit \enquote{$0$} to look like \enquote{$8$}. Instead of \textit{increasing} the prediction probability (or logit) of an adversarial target $t \neq y$, we propose to remove this dependency altogether by instead altering JSMA to \textit{decrease} the model's prediction confidence of the true class label ($t=y$). These non-targeted JSMA variants are realized by swapping the saliency measure employed, i.e., follow $S^-$ when increasing feature values (NT-JSMA+F / NT-JSMA+Z), or $S^+$ when decreasing feature values (NT-JSMA-F / NT-JSMA-Z).

\begin{algorithm}[h]
   \caption{\textcolor{blue}{Maximal} Jacobian-based Saliency Map Attack}
   \label{alg:m-jsma}
\begin{algorithmic}
   \STATE {\bfseries Input:} $n$-dimensional input $x$ normalized to $[0,1]$, \textcolor{blue}{true class label $y$}, classifier $f$, maximum number of iterations $I_{max}$, pixel perturbation step $\theta$ \textcolor{blue}{$\in (0, 1]$}, \textcolor{blue}{maximum perturbation bound $\epsilon \in (0,1]$}
   \STATE {\bfseries Initialize:} $x' \leftarrow x$, $i \leftarrow 0$, $\Gamma = \{1,...,n\}$, \textcolor{blue}{$\eta = \{0,0,...\}^n$}
   \WHILE{\textcolor{blue}{$\hat{y}(x') == y$} \AND $i < I_{max}$ \AND $|\Gamma| \geq 2$ }
   \STATE Calculate $\nabla_{x'} f(x')$
   \STATE $\gamma \leftarrow 0$
   \FOR{every pixel pair ($p,q$) $\in \Gamma$ \textcolor{blue}{and every class $t$}}
   \STATE $\alpha = \sum_{k = p,q} \nabla_{x'_{(k)}} f(x')_{(t)}$
   \STATE $\beta = \sum_{k = p,q}\sum_{c \neq t} \nabla_{x'_{(k)}} f(x')_{(c)}$
   \IF {$-\alpha \cdot \beta > \gamma$}
   \STATE $p^*,q^* \leftarrow p,q$
   \STATE $\gamma \leftarrow -\alpha \cdot \beta$
   \textcolor{blue}{
   \IF {$t == y$}
   \STATE $\theta' \leftarrow - sign(\alpha) \cdot \theta$
   \ELSE
   \STATE $\theta' \leftarrow sign(\alpha) \cdot \theta$
   \ENDIF
   }
   \ENDIF
   \ENDFOR
   \STATE $x'_{(p^*)}, x'_{(q^*)} \leftarrow \textcolor{blue}{Clip_\epsilon\{} (x'_{(p^*)} + \textcolor{blue}{\theta'}) \textcolor{blue}{\}}, \textcolor{blue}{Clip_\epsilon\{} (x'_{(q^*)} + \textcolor{blue}{\theta'}) \textcolor{blue}{\}}$
   \IF{$x'_{(p^*)}==0 \ \OR \ x'_{(p^*)}==1 \textcolor{blue}{ \ \OR \ \eta_{(p^*)} == -\theta'}$ }
   \STATE $\text{Remove} \ p^* \ \text{from} \ \Gamma$
   \ENDIF
   \IF{$x'_{(q^*)}==0 \ \OR \ x'_{(q^*)}==1 \textcolor{blue}{ \ \OR \ \eta_{(q^*)} == -\theta'}$ }
   \STATE $\text{Remove} \ q^* \ \text{from} \ \Gamma$
   \ENDIF
   \STATE \textcolor{blue}{$\eta_{(p^*)}, \eta_{(q^*)} \leftarrow \theta'$}
   \STATE $i \leftarrow i + 1$
   \ENDWHILE
   \STATE {\bfseries Return:} $x'$
\end{algorithmic}
\end{algorithm}

Extending further, we propose a combined attack, termed Maximal Jacobian-based Saliency Map Attack (M-JSMA), that merges both targeted variants and both non-targeted variants together. As shown in Algorithm~\ref{alg:m-jsma}, at each iteration the maximal-salient pixel pair is \textit{chosen over every possible class} $t$, whether adversarial or not. In this way, we find the most influential features across all classes, in the knowledge that changing these is likely to change the eventual classification. Furthermore, instead of enforcing low-saliency conditions via $S^+$ or $S^-$, we identify which measure applies to the most salient pair $(p^*,q^*)$ to \textit{decide on the perturbation direction $\theta'$ accordingly}. A history vector $\eta$ is added to prevent oscillatory perturbations. Similar to NT-JSMA, M-JSMA terminates when the predicted class $\hat{y}(x) = \arg \max_c f(x)_{(c)}$ no longer matches the true class $y$.

\begin{table}[h]
\centering
\caption{Performance comparison of the original JSMA, non-targeted JSMA, and maximal JSMA variants ($|\theta|=1$, $\epsilon=1$): \% of successful attacks, average $L_0$ and $L_2$ perturbation distances, and average entropy $H(f(x))$ of misclassified softmax prediction probabilities.}
\begin{tabular}{l|rrrr|rrrr|rrrr}
\multirow{2}{*}{Attack} & \multicolumn{4}{c|}{MNIST} & \multicolumn{4}{c|}{F-MNIST} & \multicolumn{4}{c}{CIFAR10}\\
{} & \% & $L_0$ & $L_2$ & H & \% & $L_0$ & $L_2$ & H & \% & $L_0$ & $L_2$ & H \\
\hline
JSMA+F & 100 & 34.8 & 4.32 & 0.90 & 99.9 & 93.1 & 6.12 & 1.22 & 100 & 34.7 & 3.01 & 1.27 \\
\hline
JSMA-F & 100 & 32.1 & 3.88 & 0.88 & 99.9 & 82.2 & 4.37 & 1.21 & 100 & 36.9 & 2.13 & 1.23 \\
\hline
NT-JSMA+F & 100 & 17.6 & 3.35 & 0.64 & 100 & 18.8 & 3.27 & 1.03 & 99.9 & 17.5 & 2.36 & 1.16 \\
\hline
NT-JSMA-F & 100 & 19.7 & 3.44 & 0.70 & 99.9 & 33.2 & \textbf{2.99} & \textbf{0.98} & 99.9 & 19.6 & \textbf{1.68} & \textbf{1.12} \\
\hline
M-JSMA\_F & 100 & \textbf{14.9} & \textbf{3.04} & \textbf{0.62} & 99.9 & \textbf{18.7} & 3.42 & 1.02 & 99.9 & \textbf{17.4} & 2.16 & \textbf{1.12} \\
\hline
\end{tabular}
\label{table:jsma_exp1}
\end{table}

Table~\ref{table:jsma_exp1} summarizes attacks carried out on correctly-classified test-set instances in the MNIST~\citep{726791}, Fashion MNIST~\citep{DBLP:journals/corr/abs-1708-07747}, and CIFAR10~\citep{Krizhevsky2009LearningML} datasets, using targeted, Non-Targeted, and Maximal JSMA variants.
For targeted attacks, we consider only adversaries that were misclassified in the \textit{fewest} iterations over target classes.
The JSMA+F results showed that on average only $(34.8~L_0~\text{distance})/(28*28~\text{pixels of an MNIST image}) = 4.4\%$ of pixels needed to be perturbed in order to create adversaries, thus corroborating findings from~\cite{DBLP:journals/corr/PapernotMJFCS15}.
More importantly, as evidenced by lower $L_0$ values, NT-JSMA found adversaries much faster than the \textit{fastest} targeted attacks across all 3 datasets, while M-JSMA was consistently even faster and on average only perturbed $(14.9~L_0~\text{distance})/(28*28~\text{pixels}) = 1.9\%$ of input pixels.
Additionally, the quality of adversaries found by NT-JSMA and M-JSMA were also superior, as indicated by smaller $L_2$ perceptual differences between the adversaries $x'$ and the original inputs $x$, and by lower misclassification uncertainty as reflected by prediction entropy $H\left(f(x)\right) = - \sum_c f(x)_{(c)} \cdot \log f(x)_{(c)}$.
Since M-JSMA considers all possible class targets, and both $S^+$ and $S^-$ metrics and perturbation directions, these results show that it inherits the combined benefits from both the original JSMA and NT-JSMA.

\subsection{Substitute Blackbox Attack} \label{SUBSTITUTE} 
All of the techniques covered so far are whitebox attacks, relying upon access to a model's innards. \citet{DBLP:journals/corr/PapernotMGJCS16} proposed one of the early practical blackbox methods, called the Substitute Blackbox Attack (SBA). The key idea is to train a substitute model to mimic the blackbox model, and use whitebox attack methods on this substitute. This approach leverages the transferability property of adversarial examples. Concretely, the attacker first gathers a synthetic dataset, obtains predictions on the synthetic dataset from the targeted model, and then trains a substitute model to imitate the targeted model's predictions. 

After the substitute model is trained, adversaries can be generated using any whitebox attacks since the details of the substitute model are known (e.g.,~\cite{DBLP:journals/corr/PapernotMGJCS16} used the FGSM~\citep{43405} (see Section~\ref{FGSM}) and JSMA~\citep{DBLP:journals/corr/PapernotMJFCS15} (see Section~\ref{JSMA})). We refer to SBA based on the type of adversarial attacks used when attacking the substitute model. For example, if the attacker uses FGSM to attack the substitute model, we refer this as FGSM-SBA. 

The success of this approach depends on choosing adequately-similar synthetic data samples and a substitute model architecture using high-level knowledge of the target classifier setup. As such, an intimate knowledge of the domain and the targeted model is likely to aid the attacker. Even if the absence of specific expertise, the transferability property suggests that adversaries generated from a well-trained substitute model are likely to fool the targeted model as well.

\citet{DBLP:journals/corr/PapernotMGJCS16} note that in practice the attacker is constrained from making unlimited query to the targeted model. Consequently, the authors introduced the Jacobian-based Dataset Augmentation technique, which generates a limited number of additional samples around a small initial synthetic dataset to efficiently replicate the target model's decision boundaries. Concretely, given an initial sample $x$, one calculates the Jacobian of the predicted class' likelihood assigned by the targeted model $f(x)_{(\hat{y})} = \max_{c} f(x)_{(c)}$ with respect to the inputs. Since the attacker cannot apply analytical backpropagation to the targeted model, this gradient is instead calculated using the substitute model $f'(x)$, which we denote as $\nabla_{x} f'(x)_{(\hat{y})}$. A new sample $x'$ is then synthesized by perturbing $x$ along the sign of the gradient, $x' = \alpha \cdot \text{sign}\left( \nabla_{x} f'(x)_{(\hat{y})} \right)$ by a small step $\alpha$. While this process resembles FGSM, its purpose is instead to create samples that are likely to be classified with high confidence $f'(x')_{(\hat{y})} \approx f(x')_{(\hat{y})}$. \citet{DBLP:journals/corr/PapernotMGJCS16} noted that the resulting augmented dataset better represents the decision boundary of the targeted model, in comparison to randomly sampling more data points that would most likely fall outside the target model's training-set manifold.

\begin{algorithm}[h]
   \caption{Substitute Model Training with Jacobian-based Dataset Augmentation~\citep{DBLP:journals/corr/PapernotMGJCS16}}
   \label{alg:sba}
\begin{algorithmic}
   \STATE {\bfseries Input:} initial training set $X_0$, targeted model $f$, initial substitute model $f'$, maximum number of iterations $I_{max}$, small constant $\alpha$
   \STATE {\bfseries Output:} refined substitute model $f'$
   \FOR{$i=0$ \TO $I_{max}$}
   \STATE $Y_i \leftarrow \{ f(X_i) \} \ \forall \ x \in X_i$
   \STATE Train $f'$ on $(X_i,Y_i)$ input-label pairs
   \STATE $X'_{i} \leftarrow \{x+\alpha \cdot \text{sign}\left( \nabla_{x} f'(x)_{(\hat{y})} \right) \ \forall \ x \in X_i \}$
   \STATE $X_{i+1} \leftarrow X'_{i} \cup X_i$
   \ENDFOR
   \STATE {\bfseries Return:} $f'$
\end{algorithmic}
\end{algorithm}


The entire training procedure for the substitute model is summarized in Algorithm~\ref{alg:sba}. The attacker first creates a small initial training set $X_0$. For example, $X_0$ can be initialized by picking one sample from each possible class of a set that represents the input domain of the targeted model. The substitute model is then trained on the synthetic dataset using labels provided by the targeted model (e.g., by querying the targeted model). New datapoints are then generated by perturbing each sample in the existing dataset along the general direction of variation. Finally, the new inputs are added to the existing dataset, i.e., the size of the synthetic dataset grows per iteration. This process is then repeated several times.

It is interesting to note that the targeted model does not have to be differentiable for the attack to succeed. The differentiability constraint applies only to the substitute model. As long as the substitute model has the capacity to approximate the targeted model, this attack is feasible. \citet{DBLP:journals/corr/PapernotMGJCS16} showed that substitute blackbox attack can be used to attack other machine learning models like logistic regression, Support Vector Machines (SVM)~\citep{Cortes:1995:SN:218919.218929}, k-Nearest Neighbor (kNN)~\citep{Cover:2006:NNP:2263261.2267456}, and non-differentiable models such as decision trees~\citep{DBLP:books/wa/BreimanFOS84}.

The authors evaluated SBA by targeting real world image recognition systems from Amazon, Google, and MetaMind on the MNIST dataset~\citep{726791}, and successfully fooled all targets with high accuracies ($>80\%$). This method also successfully attacked a blackbox deep neural network model that was trained on German Traffic Sign Recognition Benchmarks (GTSRB) dataset~\citep{Stallkamp2012}. Furthermore, SBA was shown to also circumvent defense methods that rely on gradient masking such as adversarial training on FGSM adversaries~\citep{43405} (see Section~\ref{ADV_TRAINING}) and defensive distillation~\citep{journals/corr/PapernotMWJS15} (see Section~\ref{DISTIL}).

\subsection{Hot/Cold Attack} \label{HOT-COLD} 

Building on the idea of altering the input to increase classifier loss, as in FGSM~\citep{43405} (see Section \ref{FGSM}), \citet{DBLP:journals/corr/RozsaRB16} proposed an attack algorithm based upon setting the values of the classification logits $Z'(x)$. We can then use the gradients with respect to the input to push inputs towards producing the desired logits. The logits are modified such that the per-class gradients will point in directions where the output of the network will increase the probability of target (\enquote{hot}) class and decrease the probability of the ground truth (\enquote{cold}) class. The Hot/Cold attack alters a target classifier's logits $Z(x)$ into:

\begin{equation} \label{equations:hot_cold}
  Z'(x) = 
  \begin{cases}
  |Z(x)_{(c)}| & \text{if $c==t$ (target class)} \\
  -Z(x)_{(c)} & \text{if $c==y$ (correct class)} \\
  0 & \text{otherwise},
  \end{cases}
\end{equation}

\noindent where $Z(x)_{(j)}$ denotes the $j$-th element of the logits vector $Z(x)$.

Intuitively, by maximizing these modified logits using gradient ascent on $x$, $Z(x)_{(t)}$ will be increased (assuming that it starts with a positive setpoint value) while $Z(x)_{y}$ will be decreased. Correspondingly, the target model will predict the adversarial class with increased likelihood $f(x)_{(t)}$ and predict the true class with decreased likelihood $f(x)_{(y)}$, since the softmax function is monotonically increasing with respect to its logit inputs. Finally, letting the other elements have zero values isolates the adversarial perturbation search to focus only on target and ground truth classes. 

Once $Z'(x)$ is obtained, and gradient directions are extracted with respect to the input $x$, we then search for the closest adversarial perturbations $x'$ along these directions using line search and bisection search, as in the L-BFGS attack. While the original manuscript~\cite{DBLP:journals/corr/RozsaRB16} had some ambiguities on how to exactly compute gradient(s) from $Z'(x)$, one sensible approach is to consider \textit{two separate} directions based on $\nabla_x Z'(x)_{(t)}$ and $\nabla_x Z'(x)_{(t)}$, perform line search on each, and select the closest adversary. Also, while not explicitly specified by the original authors, since the Hot/Cold Attack finds the \textit{closest} adversary via gradient line search, it correspondingly enforces perceptual similarity between $x'$ and $x$ following the $L_2$ metric.

The authors carried out preliminary assessments of this adversarial attack for the LeNet classification model on the MNIST dataset, and attained decent adversarial rates. Nevertheless, this work's main goal was to use generated adversaries to robustify a classifier model against further attackers via Adversarial Training~\citep{43405} (see Section~\ref{ADV_TRAINING}). The authors also anecdotally noted that many perturbations found using Hot/Cold Attack had structural visual appearances, and argued that such perturbations resemble patterns in natural images more so than the perceptually random-noise perturbations generated by FGSM~\citep{43405}.

\subsection{Carlini \& Wagner Attacks} \label{CW} 

\citet{DBLP:journals/corr/CarliniW16a} introduced a family of attacks for finding adversarial perturbations that minimize diverse similarity metrics: $L_0$, $L_2$, and $L_{\infty}$. The core insight transforms a general constrained optimization strategy similar to the L-BFGS attack~\citep{42503} (see Section~\ref{L-BFGS}) into an empirically-chosen loss function within an \textit{unconstrained} optimization formulation:

\begin{equation} \label{equations:cw_loss}
\mathcal{L}_{CW}(x',t) = \max \left( \max_{i \neq t} \{Z(x')_{(i)}\} - Z(x')_{(t)},\,-\kappa \right),
\end{equation}

\noindent where $Z(x')_{(i)}$ denotes the $i$-th component of the classifier's logits, $t$ denotes the target label, and $\kappa$ represents a parameter that reflects the minimum desired confidence margin for the adversarial example.

Conceptually, this loss function minimizes the distance in logit values between class $t$ and the second most-likely class. If $t$ currently has the highest logit value, then the difference of the logits will be negative, and so the optimization will stop when this logit difference between $t$ and the runner-up class exceeds $\kappa$. On the other hand, if $t$ does not have the highest logit value, then minimizing $\mathcal{L}(x',t)$ brings the gap between the logits of the winning class and the target class closer together, i.e., either reducing the highest class’ prediction confidence and/or increasing the target class’ confidence. 

Furthermore, the $\kappa$ parameter establishes a best-case stopping criterion, in that the logit of the adversarial class is larger than the runner-up class' logit by at least $\kappa$. Thus, $\kappa$ explicitly encodes a minimum desirable degree of robustness for the target adversary. Note that when $\kappa = 0$, the resulting adversarial examples would misclassify the network with weak robustness, as any further slight perturbations may revert to a non-adversarial softmax selection. 

The $L_2$ C\&W attack formulation is given as:

\begin{equation} \label{equations:cw_objective}
\arg \min_{w} \left( ||x'(w) - x||_2^2 + c \cdot \mathcal{L}_{CW}(x'(w), t) \right) ~~~ \textrm{where} \ x'(w) = \frac{1}{2} \left( \tanh(w) + 1 \right),
\end{equation}

\noindent where $w$ is a change of variable such that $x' = \frac{1}{2}(tanh(w)+1)$, which is introduced to bound $x'$ to be within $[0, 1]$. The minimum value of the parameter $c$ is chosen through an outer optimization loop procedure, which is detailed further below.

The $L_0$ C\&W attack is much more complex than the $L_2$ variant, since its associated distance metric is non-differentiable. Instead, an iterative strategy is proposed to successively \emph{eliminate non-significant} input features such that misclassification can be achieved by perturbing as few input values as possible. During initialization, an \emph{allowed set} $S$ is defined to include all input features in $x$. Next, at each iteration, an $L_2$ attack attempt is carried out, under the constraint of perturbing only features within $S$. If the attempt is successful, then the next non-significant feature $i$ is identified and removed from $S$, where $i^* = \arg \min_i g_{(i)} \cdot r_{(i)}$, $g = \nabla_{x'} \mathcal{L}_{CW}(x',t)$, and $r = x' - x$. This iterative procedure is repeated until the $S$-constrained $L_2$ attack fails to find an adversarial example, at which time the latest successful adversarial example is returned. To speed up this iterative algorithm, the authors suggested to \enquote{warm-start} each $L_2$ attack attempt by using the latest adversarial example found in previous iteration, which is pre-modified to satisfy the reduced $S$ set. Intuitively, the selection criterion $g_{(i)} \cdot r_{(i)}$ quantifies how much the loss value is affected by perturbing the $i$-th feature. Thus, eliminating $i^*$ with the minimum criterion score has the least amount of impact on potential misclassification.


Similar to the $L_0$ C\&W attack, the $L_{\infty}$ attack variant also requires an iterative algorithm, since the $L_{\infty}$ metric is not fully differentiable. Its optimization objective is given as:

\begin{equation} \label{equations:cw_objective_infty}
\arg \min_r \left( c \cdot \mathcal{L}_{CW}(x + r, t) + \sum_i \max(0, r_{(i)} - \tau) \right).
\end{equation}

\noindent The parameter $\tau$ is initialized to $1$, and is reduced after every iteration by a factor of $0.9$ if $r_{(i)} < \tau$ for all $i$, until no adversarial example is found. In short, this strategy successively constrains the magnitude of adversarial perturbations to be bounded by successively smaller $\tau$. Again, similar to the $L_0$ attack, warm-start can be used at every iteration to speed up the entire process.

As another practical implementation enhancement for finding robust adversaries, the authors recommend to optimize the scale parameter $c$ empirically rather than fixing it to a constant value. Concretely, in an outer optimization loop, $c$ is first set to a very stringent value (e.g., $c = 10^{-4}$), and then iteratively relaxed via doubling until a first adversary is found.

\citet{DBLP:journals/corr/CarliniW16a} empirically showed their methods to be superior to the state-of-the-art attacks at the time when evaluated on MNIST, CIFAR10, and ImageNet. In their evaluations, the proposed $L_0$ attack was compared against JSMA~\citep{DBLP:journals/corr/PapernotMJFCS15} (see Section~\ref{JSMA}), $L_2$ variant was compared against DeepFool~\citep{DBLP:journals/corr/Moosavi-Dezfooli15} (see Section~\ref{DEEPFOOL}), and $L_{\infty}$ method was compared against FGSM~\citep{43405} (see Section~\ref{FGSM}) and BIM~\citep{DBLP:journals/corr/KurakinGB16,45816} (see Section~\ref{BIM}). The three C\&W attacks consistently outperformed the incumbents in terms of average distortion and attack success rate. Furthermore, they found that JSMA is too expensive to perform when used on ImageNet models since the dimension of ImageNet data are much higher compared to MNIST or CIFAR10, while their $L_0$ method has no difficulty finding adversarial examples on ImageNet. As mentioned previously, the proposed attacks successfully circumvented defensive distillation with 100\% success rate, while keeping the adversarial examples to be similar with the original input under $L_0$, $L_2$, and $L_{\infty}$ metrics. Finally, the authors also showed that their attacks transfer between models, even to those trained using defensive distillation. They also report that increases in $\kappa$ lead to increases in transferability, supporting the notion that $\kappa$ modulates the robustness of examples. 

\subsection{Universal Adversarial Perturbation} \label{UAP} 

\begin{figure}[ht]
\centering
\includegraphics[width=0.35\textwidth]{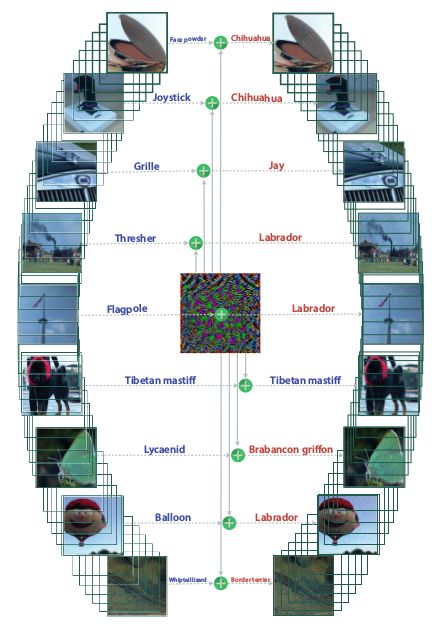}
\caption{Illustration of UAP (taken from~\citet{DBLP:journals/corr/Moosavi-Dezfooli16}).
}
\label{fig:uap}
\end{figure}

All of the attack methods covered thus far search for adversarial perturbations of a \textit{specific} input instance, such as distorting a particular image of a dog to be misclassified as \enquote{cat}. There are no guarantees that such perturbations will remain adversarial when added to a different input instance. In contrast,~\cite{DBLP:journals/corr/Moosavi-Dezfooli16} demonstrated the existence of Universal Adversarial Perturbations (UAP); perturbations that are input-agnostic. They show that a single perturbation can cause misclassification when added to \textit{most} images from a given dataset (see Fig.~\ref{fig:uap} for the illustration of UAP). The authors argued that the high adversarial efficacy of UAP indicates the presence of non-random, exploitable geometric correlations in the target model's decision boundaries.

UAP works by accumulating perturbations calculated over individual inputs. As shown in Algorithm~\ref{alg:uap}, this meta-procedure runs another \textit{sample-specific} attack (e.g., FGSM~\citep{43405}) within a loop to incrementally build a sample-agnostic UAP, $v$. In each iteration, an universally-perturbed input sample $x_i + v$ is further altered by $\Delta v_i$ via a per-sample attack method. Then, the universal perturbation is adjusted to account for $\Delta v_i$, all-the-while satisfying a maximum bound $\epsilon$ on the $p$-norm of the UAP. In this manner, the resulting $v$ can be added to \textit{each} data sample in order to push them towards a nearby decision boundary of the target model. This meta-attack procedure is repeated until the portion of samples that are misclassified, e.g., the \enquote{fooling rate}, exceeds a desired threshold $(1-\delta)$.

\begin{algorithm}[h]
   \caption{Universal Adversarial Perturbation~\citep{DBLP:journals/corr/Moosavi-Dezfooli16}}
   \label{alg:uap}
\begin{algorithmic}
   \STATE {\bfseries Input:} input-label pairs $(x_i,y_i)$ from a dataset $X$ of $|X| = N$ entries, target classifier $f$, maximum $L_p$ norm constraint $\epsilon$, minimum desired fooling rate $(1-\delta)$
   \STATE {\bfseries Output:} universal adversarial perturbation $v$
   \STATE {\bfseries Initialize:} $v \leftarrow \{0,0,...\}$
   \WHILE{fooling rate$_{X} \leq 1 - \delta$}
   \FOR{$i = 1$ \TO $N$}
   \IF{$f(x_i + v) == y_i$}
   \STATE $\Delta v_i \leftarrow \arg \min_r  ||r||_2 \ \textrm{such that} \ f(x_i + v + r) \neq y_i$
   \STATE $v_i \leftarrow v + \Delta v_i$
   \STATE $v \leftarrow \arg \min_{v'} ||v_i - v'||_2 \ \textrm{such that} \ ||v'||_p \leq \epsilon$
   \ENDIF
   \ENDFOR
   \ENDWHILE
   \STATE {\bfseries Return:} $v$
\end{algorithmic}
\end{algorithm}

Algorithm~\ref{alg:uap} is slightly modified from~\citet{DBLP:journals/corr/Moosavi-Dezfooli16}, in that we only update $v$ when an universally-perturbed sample $x_i + v$ is correctly classified (i.e., $f(x_i + v) == y_i$). The original formulation considered all samples that did not change prediction labels $f(x_i + v) == f(x_i)$. These two variants differ only for $x_i$s that are misclassified by the target model, i.e., $f(x_i) \neq y_i$. Our practical stance is that such data samples \textit{do not need to be perturbed further} to be adversarial, and that further perturbations might actually \textit{make them non-adversarial}, i.e., $f(x_i + v) == y_i$.

The authors demonstrated existence of UAPs for various models including CaffeNet~\citep{Jia:2014:CCA:2647868.2654889}, GoogLeNet~\citep{43022}, ResNet-152~\citep{He2016DeepRL}, and several variants of VGG~\citep{Simonyan14c,Chatfield14} with fooling rates as high as $93.7\%$ for the ImageNet validation dataset. UAP was also shown to be highly transferable between these models. This method was subsequently extended to find UAP that fools semantic segmentation models~\citep{metzen2017universal}.

One potential deficiency of Algorithm~\ref{alg:uap} is that it does not guarantee that each updated UAP $v$ will still be adversarial to data points that appear before the update. Indeed, the proposed method may need to iterated over the same dataset multiple times before it can attain a high desired fooling rate. It would be interesting for future research to identify a UAP variant that can update $v$ without weakening its adversarial nature on previously-seen data samples.

Additionally, we observe there is no mandatory need for Algorithm~\ref{alg:uap} to enforce $L_2$ norms, both when computing the per-instance perturbation $\Delta v_i$, and when updating $v$ towards $v_i$. While this choice of norm may be indirectly supported by related bounds for random perturbations in the authors' comparative analyses, it is nevertheless not fully justified. We argue that the UAP meta-attack procedure has two key objectives: finding a robustly-adversarial (i.e., one that confidently mis-classifies the target model) universal perturbation, while being perceptually indistiguishable as quantified by an imposed $L_p$-norm constraint, $||v||_p \leq \epsilon$. When $p \neq 2$, the per-instance attack might find an adversarial perturbation $v + \Delta v_i$ that could arbitrarily violate the latter constraint, and much of its adversarial robustness might be subsequently lost when updating $v$ towards $v_i$.

To address the above concern, we suggest that Algorithm~\ref{alg:uap} could be updated to use a per-instance adversarial attack that optimized a matching $L_p$ norm constaint, and also update $v$ as $\arg \min_{v'} ||v_i - v'||_p$. This might then mitigate the need to re-enforce the $\epsilon$ constraint explicitly. This meta-procedure could be further generalized to consider, on each update step, a multitude of adversaries generated by diverse per-instance attacks. One might then be able to choose an update direction for $v$ such that its adversarial robustness is increased while always preserving the stipulated $L_p$ constraint. We encourage the readers to substantiate this conjecture.

\subsection{Data-Free UAP} \label{DFUAP} 

As an extension of UAP~\citep{DBLP:journals/corr/Moosavi-Dezfooli16} (see Section~\ref{UAP}), \citet{DBLP:journals/corr/MopuriGB17, gduap-mopuri-2018} proposed a new algorithm to generate UAP without requiring access to the training data (i.e., data-free).\footnote{We mainly focus on the variant proposed by~\cite{gduap-mopuri-2018}, as it improves upon~\cite{DBLP:journals/corr/MopuriGB17}.} The Data-Free UAP (DFUAP) method aims to find an universal perturbation $v$ that saturates all activations of a neural network \textit{by itself}, by minimizing the following loss:

\begin{equation} \label{equations:dfuap}
\mathcal{L}_{DF} = -\log \Bigg( \prod_{i=1}^L ||f_{l_i}(v)||_2 \Bigg).
\end{equation}

\noindent Here, $f_{l_i}(v)$ denotes the activations of the model $f$ at the $i$-th layer (among $L$ total layers), given $v$ as the entire input. As is common for adversarial attacks, a perceptual similarity constraint $\epsilon$ is imposed on the magnitude of $v$ under the $\infty$-norm, i.e., $||v||_{\infty} < \epsilon$.

To implement the above optimization, $v$ is initialized randomly, and then updated via gradient descent while constrained by the $\infty$-norm bound. Also, the authors empirically found that minimizing \textit{a small subset} of activations (e.g., only convolutional layers, or only the last layers of convolutional blocks) resulted in similar fooling rates compared to minimizing \textit{all} activations, thus the former objective should be used in practice for efficiency sakes.

Although DFUAP was designed to be completely agnostic of training data, it can be slightly altered to take advantage of prior knowledge on the training data when available. For instance, \citet{gduap-mopuri-2018} suggested how one can leverage information such as the mean of a dataset, dynamic range of the dataset (e.g., [0, 255] for 8-bits RGB images), or samples of the training data. Concretely, in the case where the mean of the training data and dynamic range of the input are available, the loss function can be changed to:

\begin{equation} \label{equations:dfuap_mean_variance}
\mathcal{L}_{d} = -\sum_{d \sim \mathcal{N}(\mu,\sigma)} \log \Bigg( \prod_{i=1}^L ||f_{l_i}(d+v)||_2 \Bigg),
\end{equation}

\noindent where $d$ denotes a random perturbation sampled from Gaussian distribution with mean $\mu$ and variance $\sigma$ such that $d$ lies within the dynamic range of the input. Alternatively, if some training samples $x \in X$ are available, then the loss could be modified to:

\begin{equation} \label{equations:dfuap_data_prior}
\mathcal{L}_{x} = -\sum_{x \sim X} \log \Bigg( \prod_{i=1}^L ||f_{l_i}(x+v)||_2 \Bigg).
\end{equation}

Although in general DFUAP is not as effective at fooling the target model as the UAP generated by~\cite{DBLP:journals/corr/Moosavi-Dezfooli16}, this method can be significantly faster to compute, since one does not have to iterate through the entire dataset. \citet{gduap-mopuri-2018} also showed that this method can be used to attack semantic image segmentation (e.g., FCN~\citep{Shelhamer:2017:FCN:3069214.3069246}) and depth estimation (e.g., Monodepth~\citep{monodepth17}) models. Furthermore,~\cite{DBLP:journals/corr/Moosavi-Dezfooli16} showed that the fooling rate of the data-free UAP method is comparable to the original UAP method when data prior is available. Perhaps more interestingly, they showed that the original UAP method only outperformed the data-free method when the dataset size was sufficiently large (e.g., at least 10,000 images for ImageNet-level models such as the GoogLeNet~\citep{43022} or CaffeNet~\citep{Jia:2014:CCA:2647868.2654889}).

\subsection{VAE Attacks} \label{ADV_GEN} 

Although the majority of work in the field has focused upon fooling classifiers, recent research has extended adversarial attacks to generative models.~\citet{DBLP:journals/corr/KosFS17} focus upon the scenario where a generative model is being used as a sophisticated compression mechanism, allowing a sender and receiver to efficiently exchange information using a shared latent space. Concretely, a sender takes an input image, encodes it with the model, and sends the receiver the latent code. The receiver uses the same model to decode from this latent representation to an output image. In this case, an adversarial example would be one that is poorly reconstructed after encoding and decoding through the model.

\citet{DBLP:journals/corr/KosFS17} describe ways to find adversarial examples for models like the Variational Autoencoder (VAE)~\citep{DBLP:journals/corr/KingmaW13} and VAE-GAN~\citep{Larsen:2016:ABP:3045390.3045555}. The authors proposed three separate strategies: classifier attack, $\mathcal{L}_{VAE}$ attack, and latent attack. These methods apply to models that specifically have an encoder and a decoder component, which respectively, compresses the input $x$ into a latent representation $z$, and reconstructs the input $\hat{x}$ from $z$, i.e., $z = f_{enc}(x)$ and $\hat{x} = f_{dec}(z)$.

The simplest method, the classifier attack, augments the frozen generative model (e.g., VAE or VAE-GAN), by training a new classifier network $y = f_{classifier}( f_{enc}(x) )$, and then applies an existing algorithm like FGSM~\citep{43405} (see Section \ref{FGSM}) to find adversaries $x'$. During the process, this approach also alters the latent encoding $z$ into an adversarial counterpart $z'$ such that the decoder $f_{dec}(z')$ produces poor reconstructions $\hat{x}'$. Nevertheless, there are no general guarantees that success in fooling the classifier will result in fooling the decoder.

The setup of the second attack method, $\mathcal{L}_{VAE}$ attack, involves altering the loss function for training a VAE:

\begin{equation} \label{equations:vae_loss}
\mathcal{L}_{VAE} \left(x, \hat{x} \right) = - KL \left( q(z|x)\,||\,p(z) \right) + E_q \left[ \log p(\hat{x} | z) \right],
\end{equation}
\noindent where $KL(\cdot)$ denotes the Kullback-Leibler divergence~\citep{kullback1951}, $q(z|x)$ denotes an analytical approximation for the underlying conditional latent distribution $p(z|x)$, $p(z)$ is a prior distribution of $z$ that is assumed to be Gaussian, and $E_q \left[ \log p(x|z) \right]$ is the evidence lower bound (ELBO), which in this setup entails the cross-entropy loss between the input $x$ and its reconstruction $\hat{x}$.

When training a VAE normally, the reconstruction target $\hat{x}$ of $\mathcal{L}_{VAE}(\cdot)$ is naturally set to the output of the encoder-decoder networks, i.e., $\hat{x} = f_{dec}(f_{enc}(x))$. However, this loss metric can be altered to quantify the gap between a given input $x$, and the reconstruction of another \textit{adversarial target} input $\hat{x}_t$. Subsequently, an optimization formulation resembling methods such as the L-BFGS attack~\citep{42503} (see Section~\ref{L-BFGS}) can be formulated as:

\begin{equation} \label{equations:adv_gen}
\arg \min_{x'} \left( c||x - x'||_2 + \mathcal{L}_{VAE}(x', \hat{x}_t) \right) \ \textrm{such that} \ x' \in [0,1].
\end{equation}

\noindent Similar to the L-BFGS attack, the scaling parameter $c$ is initially set to a small value, and then incrementally increased via line search till an adversary is found. This outer optimization loop ensures that the resulting adversary strictly enforces the $\mathcal{L}_{VAE}(x', \hat{x}_t)$ loss.

The third strategy, latent attack, differs in the goal of matching $f_{enc}(x')$ to an adversarial target \textit{latent vector} $z_t$, in contrast to the $\mathcal{L}_{VAE}$ attack's aim of matching the reconstructions between $x'$ and an adversarial target \textit{input} $x_t$. To this end, a different loss function is used:

\begin{equation} \label{equations:latent_loss}
\mathcal{L}_{latent}(x',z_t) = ||f_{enc}(x') - z_t||_2,
\end{equation}

\noindent while the adversarial optimization process remains the same:

\begin{equation} \label{equations:adv_gen_2}
\arg \min_{x'} \left( c||x - x'||_2 + \mathcal{L}_{latent}(x', \hat{x}_t) \right) ~~~ \textrm{such that} \ x' \in [0,1].
\end{equation}

\citet{DBLP:journals/corr/KosFS17} showed that the above attack strategies successfully fool VAE and VAE-GAN on various datasets such as MNIST~\citep{726791}, SVHN~\citep{37648}, and CelebA~\citep{DBLP:journals/corr/LiuLWT14}. However, the classifier attack produces lower quality reconstructions than the $\mathcal{L}_{VAE}$ and latent attacks, which may be due to the fact that the classifier itself is easier to fool compared to the generative model. The authors also report that the $\mathcal{L}_{VAE}$ attack was the slowest since this method builds a reconstruction at every iteration, whilst the latent attack was found to be the most effective.

Although these attacks are meant to fool generative reconstruction models rather than discriminative classification models, some adversarial defense techniques use similar generative models to remove adversarial perturbations from an input, such as PixelDefend~\citep{song2018pixeldefend} (see Section~\ref{PIXELDEFEND}) and DefenseGAN~\citep{samangouei2018defensegan} (see Section~\ref{DEFENSEGAN}). Consequently, knowing that generative models can also be attacked may alert one from naively favoring generative models as a defense strategy.

\subsection{Adversarial Transformation Networks} \label{ATN} 

\begin{figure}[ht]
\centering
\subfloat[AAE]{\label{fig:aae} \includegraphics[width=0.6\textwidth]{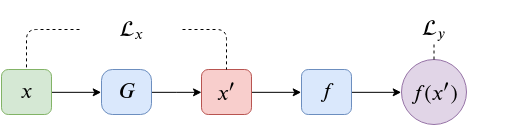}}
\qquad
\subfloat[P-ATN]{\label{fig:patn} \includegraphics[width=0.7\textwidth]{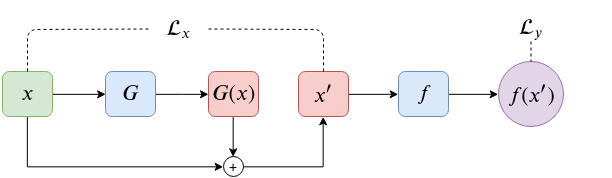}}
\caption{Illustrations of AAE and P-ATN.}
\label{fig:atn}
\end{figure}

Instead of using optimizers like L-BFGS attack directly (see Section~\ref{L-BFGS}), Adversarial Transformation Networks (ATN)~\citep{46527} are neural networks that either transform encode-and-reconstruct non-adversarial inputs into adversarial counterparts, or generates an additive adversarial perturbation given a specific input. Once trained, the network can swiftly generate new adversarial examples. The former variant, the Adversarial Auto-Encoding (AAE) network, is depicted in Fig.~\ref{fig:aae}, which contrasts with the latter Perturbation ATN (P-ATN) model, as seen in Fig.~\ref{fig:patn}.

Given a pre-trained target classifier network $f$, in both variants the generator $G$ is trained to minimize:

\begin{equation} \label{equations:atn_loss}
\mathcal{L}_x(x,x') + \alpha \mathcal{L}_y(f(x'), r(y,t)),
\end{equation}

\noindent where $\mathcal{L}_x$ enforces perceptual similarity between $x$ and $x'$ \footnote{While the original authors used $L_2$ metrics for $\mathcal{L}_x$ and $\mathcal{L}_y$, this is not a mandatory requirement, as others have found adversaries using other similarity metrics, such as $L_2$ loss in feature space~\citep{Gatys2015c,DBLP:journals/corr/JohnsonAL16}.}, while $\mathcal{L}_y$ forces the softmax probabilities of the perturbed input, $f(x')$, to match an adversarial class distribution $r(y,t)$. The scaling hyper-parameter $\alpha$ is either set heuristically or empirically optimized to adequately balance between the perceptual similarity and adversarial misclassification objectives.

One naive instantiation of ATN is to set the adversarial class distribution $r(y,t)$ to a one-hot encoding of $t$, as when training normal classification models. Nevertheless, the authors proposed a more effective \textit{reranking} function, whose $k$-th components are minimally different from the softmax probabilities of the pre-trained classifier $f(x)_{(k)}$:

\begin{equation} \label{equations:reranking function}
  r(y,t)_{(k)} \propto
  \begin{cases}
  \beta \cdot \max(f(x)) & \text{if $k=t$}
  \\
  f(x)_{(k)} & \text{otherwise}
  \end{cases}
\end{equation}

\noindent Since the $\beta>1$ hyper-parameter enhances the desired confidence for the adversarial class, in general $r(y,t)$ needs to be vector-re-normalized. It is important to note that each trained ATN generator can only produce adversarial examples that will be misclassified as a particular class by the targeted model. Multiple ATNs must be trained in order to fool a given classifier into diverse adversarial classes.


~\citet{46527} found that the transferability of adversarial examples generated by ATN is fairly poor, but can be enhanced by training the ATN to fool multiple networks at the same time (e.g., the gradients coming from attacking multiple networks are averaged during backpropagation). ATN exhibits some benefits that other attack methods are lacking: for instance, since ATN takes advantage of the expressiveness of neural networks, the adversarial examples generated by ATN tend to be more diverse. This diversity of adversarial examples can also be leveraged for adversarial training~\citep{43405} (see Section~\ref{ADV_TRAINING}). Similar to preliminary investigations by~\citet{DBLP:journals/corr/abs-1711-04368}, a future interesting research direction is to formulate a min-max game between a classifier and an ATN that can provide some \emph{guarantee} of the classifier's robustness, by learning from diverse sets of adversarial examples generated by the ATN without suffering from catastrophic forgetting~\citep{Kirkpatrick3521} between training iterations.

\subsection{Dense Adversary Generation} \label{DAG} 

Dense Adversary Generation (DAG)~\citep{DBLP:journals/corr/XieWZZXY17} is a targeted attack method for semantic image segmentation and object detection models. DAG generalizes from attacks on classification models in that it aims to misclassify \emph{multiple target outputs} associated to each given input, namely multiple pixel labels for semantic segmentation tasks, or multiple region proposals for object detection tasks. DAG requires both the ground truth class labels as well as specific adversarial labels for every target output of a given model. Using this information, DAG incrementally builds an adversarial input perturbation $r$ that decreases the predicted logits for the true classes while increasing the logits for the adversarial labels among target outputs, thus resulting in as many misclassified outputs as possible.\footnote{While the original formulation for DAG computes input-gradients targeting the logit layer $Z(x,\tau)$ rather than targeting the normalized softmax probabilities $f(x,\tau)$, there is no fundamental limitation to forbid undertaking the latter variant. Nevertheless, as shown by \citet{DBLP:journals/corr/CarliniW16a} and others, in certain setups such as with defensively-distilled models~\citep{journals/corr/PapernotMWJS15}, attack methods that operate on the logit layer tend to be significantly more successful than those targeting the softmax layer.}

Formally, given an image $x$, we expand the definitions of the logit vectors $Z(x,\tau)$, softmax probabilities $f(x,\tau)$, and predicted class labels $\hat{y}_\tau(x) = \arg \max_c f(x,\tau)_{(c)}$ to be specific to each of the target outputs $\tau \in 1:N$ of the target model. As shown in Algorithm~\ref{alg:dag}, the input $x$ is incrementally perturbed by $r_i'$ into $x'_{i+1}$ on each iteration $i$. To compute each such perturbation, first DAG identifies the set of target outputs $\mathcal{T}$ that are still correctly classified by the model given $x'_{i}$. Among each $\tau \in \mathcal{T}$, the $r_i$ vector is then built by accumulating the positive input-gradients of logits for output-specific adversarial classes $\{t_n\}$, i.e., $\nabla_{(x'_i)} Z_{(t_n)}(x'_i, \tau)$, as well as the negative input-gradients of the true-class logits, $\nabla_{(x'_i)} Z_{(y_n)}(x'_i, \tau)$. 
This resulting perturbation $r_i$, when added to $x'_i$, has the effect of amplifying the likelihoods for predicting adversarial classes over true classes among the non-misclassified outputs, thus will likely result in more misclassified outputs overall. For numerical stability, the authors recommend to down-scale $r_i$ by an adversarial gain hyper-parameter $\gamma$, e.g., $\gamma = 0.5$. This iterative procedure is repeated for a fixed number of iterations $I_{max}$, which is empirically set to $I_{max}=200$ for semantic segmentation tasks and $I_{max}=150$ for object detection tasks.

\begin{algorithm}[h]
   \caption{Dense Adversary Generation~\citep{DBLP:journals/corr/XieWZZXY17}}
   \label{alg:dag}
\begin{algorithmic}
   \STATE {\bfseries Input:} input image $x$, model $f$, ground truth output-specific labels $\{y_1,...,y_N\}$, adversarial labels $\{t_1,...,t_N\}$, adversarial gain $\gamma$, maximum number of iterations $I_{max}$
   \STATE {\bfseries Output:} adversarial perturbation $r$
   \STATE {\bfseries Initialize:} $r = \left\{0,0,...\right\}$, $i = 0$, $x'_0 \leftarrow x$, $\mathcal{T} \leftarrow \{1,...,N\}$
   \WHILE{$i < I_{max}$ \AND $\mathcal{T} \neq \varnothing$}
   \STATE $\mathcal{T} = \{ \tau: \hat{y}_\tau(x_i) == y_{\tau}, \tau \in 1:N \}$
   \STATE $r_i \leftarrow \sum_{\tau \in \mathcal{T}} \left[ \nabla_{(x'_i)} Z_{(t_n)}(x'_i, \tau) - \nabla_{(x'_i)} Z_{(y_n)}(x'_i, \tau) \right]$
   \STATE $r'_i \leftarrow \frac{\gamma}{||r_i||_{\infty}} r_i$
   \STATE $r, x'_{i+1} \leftarrow r + r'_i, x'_i + r'_i$
   \STATE $i \leftarrow i + 1$
   \ENDWHILE
   \STATE {\bfseries Return:} $r$
\end{algorithmic}
\end{algorithm}

The authors showed that DAG can be used to attack various Fully Convolutional Networks (FCN)~\citep{Shelhamer:2017:FCN:3069214.3069246} for semantic segmentation tasks, as well as some variants of Faster-RCNN~\citep{NIPS2015_5638} for object detection tasks. Intriguingly, the generated adversarial perturbations for attacking a semantic segmentation model can also be used to fool an object detection model, and vice-versa. This transfer is particularly effective when both models employ the same convolutional backbone (e.g., both FCN and Faster-RCNN models use a VGG16 architecture).

In practice, for object detection models, the authors noted that region proposal target outputs often change drastically when different perturbations are added. Consequently, they heuristically recommend to consider only outputs that have an Intersection over Union (IoU) with the closest ground truth object that is larger than $0.1$ along with a confidence score greater than $0.1$. If there are several target outputs belonging to the same object satisfy both conditions, only the one with the largest IoU is considered. This heuristic assumes that the changes in values among nearby region proposals will transfer to each other accordingly.

\subsection{Zeroth Order Optimization} \label{ZOO} 

\citet{Chen:2017:ZZO:3128572.3140448} introduced a blackbox attack method called the Zeroth Order Optimization (ZOO). Given only the capability to query a target model, ZOO approximates the gradients of the objective function with respect to the input using finite-difference numerical estimates. Unlike other methods that use stochastic gradient descent~\citep{Bottou10large-scalemachine}, ZOO employs stochastic coordinate descent~\citep{Wright:2015:CDA:2783158.2783189} over small batches of input dimensions per iteration, to avoid calculating derivatives with respect to all input features (e.g., all pixels of an image).

Two ZOO variants were introduced: ZOO-Adam and ZOO-Newton. ZOO-Adam uses first-order approximations for calculating the derivatives and then use the Adam optimizer~\citep{DBLP:journals/corr/KingmaB14} to update the input. Given a particular loss function $\mathcal{L}(x)$ to optimize for misclassification, the coordinate-wise Jacobian is estimated as:

\begin{equation} \label{equations:zoo_jacobian}
J_i = \frac{\partial \mathcal{L}(x)}{\partial x_{(i)}} \approx \frac{\mathcal{L}(x + h \cdot e_i) - \mathcal{L}(x - h \cdot e_i)}{2 h},
\end{equation}

\noindent where $x_{(i)}$ is the $i$-th element of the input $x$, $h$ is a small constant (e.g., $h= 10^{-10}$), and $e_i$ is an elementary vector where only the $i$-th element is set to 1 while other entries are 0. The algorithm for ZOO-Adam can be found in Algorithm~\ref{alg:zoo_adam}.

\begin{algorithm}[h]
   \caption{ZOO-Adam~\citep{Chen:2017:ZZO:3128572.3140448}}
   \label{alg:zoo_adam}
\begin{algorithmic}
   \STATE {\bfseries Input:} original input $x \in \mathbb{R}^n$, model $f$, learning rate $\alpha$, Adam hyper-parameters $\beta_1, \beta_2, \epsilon$
   \STATE {\bfseries Output:} adversarial input $x'$
   \STATE {\bfseries Initialize:} first-moment vector estimate for Adam $m \leftarrow \{0,...,0\}^n$, second-moment vector estimate for Adam $v \leftarrow \{0,...,0\}^n$, per-coordinate timestep for Adam $t \leftarrow \{0,...,0\}^n$, $x' \leftarrow x$
   \WHILE{$x'$ is not adversarial}
   \STATE set $e_i$ by randomly choosing coordinate $i \in \{1,...,n\}$
   \STATE $J_i \leftarrow \frac{\mathcal{L}(x + h \cdot e_i) - \mathcal{L}(x - h\cdot e_i)}{2h}$
   \STATE update Adam estimates and compute $\delta$:
   \STATE $t_{(i)} \leftarrow t_{(i)} + 1$
   \STATE $m_{(i)} \leftarrow \beta_1 m_{(i)} + (1 - \beta_1) J_i$
   \STATE $v_{(i)} \leftarrow \beta_2 v_{(i)} + (1 - \beta_2) J_i^2$
   \STATE $\hat{m_{(i)}} \leftarrow \frac{m_{(i)}}{1 - \beta_1^{t_{(i)}}}$
   \STATE $\hat{v_{(i)}} \leftarrow \frac{v_{(i)}}{1 - \beta_2^{t_{(i)}}}$
   \STATE $\delta \leftarrow -\alpha \frac{\hat{m_{(i)}}}{\sqrt[]{\hat{v_{(i)}}} + \epsilon}$
   \STATE $x'_{(i)} \leftarrow x'_{(i)} + \delta$
   \ENDWHILE
   \STATE {\bfseries Return:} $x'$
\end{algorithmic}
\end{algorithm}

On the other hand, ZOO-Newton employs second-order approximations for derivatives, and thus additionally requires estimates for the coordinate-wise Hessian:

\begin{equation} \label{equations:zoo_hessian}
H_i = \frac{\partial^2 \mathcal{L}(x)}{\partial x^2_{i}} \approx \frac{\mathcal{L}(x + h \cdot c_i) - 2\mathcal{L}(x) + \mathcal{L}(x - h \cdot c_i)}{h^2}.
\end{equation}

\noindent Algorithm~\ref{alg:zoo_newton} demonstrates how ZOO-Newton works.

\begin{algorithm}[h]
   \caption{ZOO-Newton~\citep{Chen:2017:ZZO:3128572.3140448}}
   \label{alg:zoo_newton}
\begin{algorithmic}
   \STATE {\bfseries Input:} original input $x \in \mathbb{R}^n$, model $f$, learning rate $\alpha$
   \STATE {\bfseries Output:} adversarial output $x'$
   \WHILE{$x'$ is not adversarial}
   \STATE set $e_i$ by randomly choosing coordinate $i \in \{1,...,n\}$
   \STATE $J_i \leftarrow \frac{\mathcal{L}(x+h \cdot e_i) - \mathcal{L}(x-h \cdot e_i)}{2h}$
   \STATE $H_i \leftarrow \frac{\mathcal{L}(x+h \cdot e_i) - 2\mathcal{L}(x) + \mathcal{L}(x-h \cdot e_i)}{h^2}$
   \IF{$H_i \leq 0$}
   \STATE $\delta \leftarrow -\alpha J_i$
   \ELSE
   \STATE $\delta \leftarrow -\alpha \frac{J_i}{H_i}$
   \ENDIF
   \STATE $x'_i \leftarrow x'_i + \delta$
   \ENDWHILE
   \STATE {\bfseries Return:} $x'$
\end{algorithmic}
\end{algorithm}

In both algorithms, one component of the algorithm remains to be specified: the adversarial loss function $\mathcal{L}(\cdot)$. The authors presented both a targeted variant (with adversarial target class $t$) as well as a non-targeted version (with $y$ as the ground truth label), with $\kappa$ being a minimal confidence gap hyper-parameter:

\begin{equation} \label{equations:zoo_loss_targeted}
\mathcal{L}^{targeted}(x, t) = \max \left\{ \max_{i \neq t} \log (f(x)_{(i)}) - \log (f(x)_{(t)}),\,-\kappa \right\},
\end{equation}

\begin{equation} \label{equations:zoo_loss_nontargeted}
\mathcal{L}^{non-targeted}(x) = \max \left\{ \log (f(x)_{(y)}) - \max_{i \neq y} \log (f(x)_{(i)}),\,-\kappa \right\}.
\end{equation}

While these loss functions are inspired and thus similar to those used in the C\&W attack~\citep{DBLP:journals/corr/CarliniW16a} (see Section \ref{CW}), the authors argued that the use of the $\log$ transform of $f(x)$ was important to reduce the potentially large differences in prediction scores between the best class and its runner-up. Another difference with the loss functions of ZOO and C\&W is that the former targets the softmax layer, while the latter targets the logit layer. While the authors compared ZOO to C\&W attacks, it would nevertheless be interesting to evaluate variants of ZOO where the loss functions are computed with respect to logits $Z(x)_{(i)}$ rather than softmax probabilities $f(x)_{(i)}$.

Several suggestions were made to increase the efficiency of ZOO. First,~\cite{Chen:2017:ZZO:3128572.3140448} suggested to perform dimensionality reduction on the attack space using bilinear interpolation or discrete cosine transformations~\citep{Ahmed:1974:DCT:1309267.1309385}. Although this trick reduces the computational costs, note that it also lowers the chance of finding adversarial examples. The authors thus proposed a hierarchical attack scheme, which gradually increasing the attack space dimensionality every several number of iterations until satisfactory examples are found. Second, when the attack space dimension is large, importance sampling can be used to only update the most influential coordinates, as a further optimization heuristic. This can be realized by first dividing an input into 8 by 8 pixel regions, where each region is assigned a sampling probability. At the beginning of attack process, sampling probabilities for each region are assigned uniformly. During the attack process, after every few iterations, these probabilities are calculated by performing max pooling on the absolute pixel value changes in each region, which are then normalized to $[0,1]$ in order to get the probability values.

Although ZOO algorithms are empirically slower compared to the SBA~\citep{DBLP:journals/corr/PapernotMGJCS16} (see Section~\ref{SUBSTITUTE}), \citet{Chen:2017:ZZO:3128572.3140448} showed that ZOO are more effective in terms of attack success rate on models that rely on defensive distillation~\citep{journals/corr/PapernotMWJS15} (see Section~\ref{DISTIL}), and comparable even to C\&W attacks~\citep{DBLP:journals/corr/CarliniW16a}.

\subsection{One-Pixel Attack} \label{ONEPIX} 

The One-pixel Attack (OPA)~\citep{DBLP:journals/corr/abs-1710-08864} is a \enquote{semi-blackbox} method\footnote{The authors coined the term \enquote{semi-blackbox} to denote the requirements of needing both the predicted classes $\hat{y}(x)$ and the softmax prediction probabilities $f(x)$ from the target model. This semantic designation is questionable, as only some blackbox models in practice provide prediction probabilities.} that uses an Evolutionary Algorithm strategy called Differential Evolution~\citep{Storn1997,5601760} to find adversarial perturbations. OPA aims to fool the target model by modifying as few as one feature of a given image $x$. In practice the authors have shown high misclassification rates for diverse models and datasets by perturbing only 3 or 5 pixels.

Concretely, given an image $x$, an $N$-pixel attack works by evolving a population of candidate modifications over many generations to ultimately home in onto a globally-optimal solution. Each candidate $r$ is represented as a $N$-tuple of coordinates and values to be modified from $x$. For example, a candidate in a $3$-pixel attack on an RGB image consists of a vector of $(3 \ \textrm{pixels}) \times \left((2 \ \textrm{coordinates}) + (3 \ \textrm{intensity values})\right) = 15$ elements. 

A \enquote{fitness function} $F(x \bigoplus r)$ quantifies the optimality of a resultant image after applying ($\bigoplus$) the modification $r$ onto the source image $x$. OPA can be instantiated as either a targeted attack or a non-targeted one, which differ only in their fitness functions. Notably, the former fitness function is the model's softmax prediction probability for a specified adversarial class $t$, i.e., $F(\cdot) = f(x \bigoplus r)_{(t)}$, while the latter fitness function for a non-targeted attack is the \textit{negative} of the softmax probability for the ground-truth class $y$, i.e., $F(\cdot) = -f(x \bigoplus r)_{(y)}$.

As summarized in Algorithm~\ref{alg:onepix}, a population of $P=400$ candidates is initialized randomly, with pixel coordinates sampled uniformly and 8-bit intensity values sampled from Gaussian distributions $\mathcal{N}(\mu=128, \sigma=127)$. This population is evolved for up to $I_{max}=100$ generation. An early-stopping heuristic is also employed when a sufficiently large portion of the population becomes adversarial. During the $i$-th generation, the $p$-th candidate is evolved by choosing the modification with greater fitness between either the existing vector $r^i_p$ or a donor mutation $r'$ combined from 3 other distinct candidates:\footnote{The proposed OPA procedure~\citep{DBLP:journals/corr/abs-1710-08864} is a simplified variant of the standard Differential Evolution algorithm~\citep{5601760} in that there is no crossover mutation between the existing candidate $r^i_p$ and the donor $r'$. Also, \citet{5601760} prescribes the 3 randomly sampled candidates to be distinct from the existing candidate as well, i.e., $p \neq p_1 \neq p_2 \neq p_3$. Nevertheless, we have empirically corroborated that the simplified evolutionary strategy used by OPA can successfully find adversaries, despite possibly suffering from reduced search efficiency.}

\begin{equation} \label{equations:onepix_generation}
r^{i+1}_p = \arg \max \left( F(x \bigoplus r^i_p ), F(x \bigoplus r') \right), \ \textrm{where} \ r' = r^i_{p_1} + \lambda( r^i_{p_2} - r^i_{p_3}); p_1 \neq p_2 \neq p_3 \in \{1,2,...,P\},
\end{equation}

\noindent and $\lambda = 0.5$ is a scaling hyper-parameter.

\begin{algorithm}[h]
   \caption{Targeted Variant of One-Pixel Attack~\citep{DBLP:journals/corr/abs-1710-08864}}
   \label{alg:onepix}
\begin{algorithmic}
   \STATE {\bfseries Input:} input $x$, model $f$, target class label $t$, population size $P$, maximum number of generations $I_{max}$, mutation scale $\lambda$, minimum population ratio of adversarial examples for early-stopping $\rho$
   \STATE {\bfseries Output:} adversarial output $x'$
   \STATE define fitness function $F(x') \leftarrow f(x')_{(t)}$
   \STATE randomly initialize population $\{ r^1_p \}$ for $p \in \{1,2,...,P\}$
   \STATE $i \leftarrow 0$
   \WHILE{$i < I_{max}$}
   \STATE $i \leftarrow i + 1$
   \FOR{$p \in \{1,2,...,P\}$}
   \STATE sample distinct $p_1,p_2,p_3 \in \{1,2,...,P\}$ such that $p_1 \neq p_2 \neq p_3$
   \STATE $r' \leftarrow r^i_{p_1} + \lambda( r^i_{p_2} - r^i_{p_3})$
   \STATE $r^{i+1}_p = \arg \max \left( F(x \bigoplus r^i_p ), F(x \bigoplus r') \right)$
   \ENDFOR
   \IF {ratio of $\{p: \hat{y}(x \bigoplus r^i_p) == t, p \in \{1,2,...,P\} \} \geq \rho$}
   \STATE \textbf{break}
   \ENDIF
   \ENDWHILE
   \STATE {\bfseries Return:} $\arg \max_{x'} \{F(x'): x' = x \bigoplus r^i_p, p \in \{1,2,...,P\}, \hat{y}(x') == t \}$
\end{algorithmic}
\end{algorithm}

As practical considerations, each donor mutation vector $r'$ should be clipped to be within the respective bounds for pixel dimensions and intensity values. Also, all pixel intensity values should be discretized to the same resolution as $x$, e.g., 8-bits.

The authors showed that many images in CIFAR10~\citep{Krizhevsky2009LearningML} and ImageNet~\citep{imagenet_cvpr09,ILSVRC15} can be minimally modified to fool various classification models, such as the All Convolutional Net~\citep{DBLP:journals/corr/SpringenbergDBR14}, Network in Network~\citep{DBLP:journals/corr/LinCY13}, VGG16~\citep{Simonyan14c}, and AlexNet~\citep{Krizhevsky:2012:ICD:2999134.2999257}.


\subsection{Houdini} \label{HOUDINI} 

All optimization-based adversarial attacks operate by maximizing an \textit{attack loss} $\ell^{atk}(x', y)$, such as the distance between the model prediction $\hat{y}(x')$ versus the ground-truth target $y$, while constrained by a perceptual bound $||x'-x||_p \leq \epsilon$ of the input perturbations:

\begin{equation}
x' = \text{arg\,max}_{x'': ||x''-x||_p \leq \epsilon} \ell^{atk}(x'', y) \nonumber
\end{equation}

Typically, the attack loss $\ell^{atk}$ is derived from the \textit{training loss} $\ell^{train}=\mathcal{L}(f(x'),y)$ of a given model $f$. For most tasks, since $\mathcal{L}$ is differentiable, this has the convenient consequence of allowing attacks to use efficient gradient-based optimization strategies. Nevertheless, to generate robust adversaries, one should ideally choose an attack loss $\ell^{atk}$ that maximally damages the \textit{evaluation criterion} $\ell^{eval}$, which is different from the training loss for some family of tasks such as structured prediction. This poses a technical challenge for the attack however, as often the evaluation metric is non-differentiable and even combinatorial. For example, while semantic segmentation models are trained to minimize per-pixel misclassification error, they are evaluated using the mean Intersection-Over-Union (mIOU) ratio across classes. The latter metric is more desirable as it does not exhibit over-confidence for imbalanced class labels~\citep{DBLP:conf/nips/CisseANK17}.

To address the above challenge, \citet{DBLP:conf/nips/CisseANK17} introduced a differentiable surrogate attack loss, called Houdini $\ell_H$, for efficiently generating adversarial examples in structured prediction tasks. This is achieved by combining a non-differentiable evaluation metric $\ell^{eval}$ with a stochastic margin on $\delta(x,y) = f(y) - f\left(\hat{y}(x)\right)$:

\begin{equation} \label{equations:houdini}
\ell_H(x,y) = \mathbb{P}_{\gamma \sim \mathcal{N}(0,1)}\left[ \delta(x,y) <\gamma \right] \cdot \ell^{eval}(x,y) = \frac{1}{\sqrt[]{2\pi}} \int_{ \delta(x,y) }^{\infty} e^{\frac{-v^2}{2}} dv \cdot \ell^{eval}(x,y),
\end{equation}

\noindent where $\gamma$ is a random variable following an unit Gaussian distribution $\mathcal{N}(0,1)$. Importantly, this formulation assumes that the victim model $f$ generates a numeric (e.g., probability) \textit{score map} $f(y')$ over all possible output values $y'$. For classification tasks, $f(y'=c)$ pertains to the softmax probability of a specified output class $c$. More complex score maps also exist for structured prediction tasks such as semantic segmentation, which we will clarify when presenting concrete examples below.

The form of Equation~\ref{equations:houdini} is less important in practice than its derivative. Since the stochastic margin is expressed as an \textit{integral} of an exponential term, its \textit{derivative} shares the integrand's form. Thus, the input-Jacobian of the Houdini loss decomposes into a total-derivative sum of products, each containing a Houdini-specific partial derivative and a standard network backpropagation term:

\begin{equation} \label{equations:houdini_input_jacobian}
\nabla_x \ell_H(x,y) = \sum_{y'} \frac{\partial \ell_H(x,y)}{\partial f\left(y'\right)} \frac{\partial f\left(y'\right)}{\partial x},
\end{equation}

\noindent with the first term given as:

\begin{equation} \label{equations:houdini_gradient_final}
  \frac{\partial \ell_H(x,y)}{\partial f\left(y'\right)} =
  \begin{cases}
  -\frac{1}{\sqrt[]{2\pi}}e^{-\frac{\delta(x,y)^2}{2}} \cdot \ell^{eval}(x,y), & y' = y \\
  \frac{1}{\sqrt[]{2\pi}}e^{-\frac{\delta(x,y)^2}{2}} \cdot \ell^{eval}(x,y), & y' = \hat{y}(x) \\
  0, & \text{otherwise.}
  \end{cases}
\end{equation}

\noindent This input-Jacobian can be used to maximize the attack loss $\ell_H(x,y)$ using non-targeted methods such as the FGSM~\citep{43405} (see Section~\ref{FGSM}). Alternatively, one can apply it to minimize the attack loss $\ell_H(x,t)$ of an adversarial class $t \neq y$, using algorithms such as JSMA~\citep{DBLP:journals/corr/PapernotMJFCS15} (see Section~\ref{JSMA}).

A desirable property of the Houdini loss is that it is consistent with the task evaluation criterion, and thus choosing an input perturbation $x'$ that optimizes $\ell_{H}$ would adversarially impact $\ell^{eval}(x',y)$. Moreover, even when a perturbation $x'$ does not change $\ell^{eval}$ (e.g., due to discretization), the gradients in Equation~\ref{equations:houdini_gradient_final} \textit{smoothly} point towards the direction that reduces the score gap $\delta$ between the predicted output $\hat{y}(x')$ and the true target $y$.

The authors demonstrated attacks using the Houdini loss for three structured prediction setups. Firstly, consider a human pose estimation task, where the goal is to use an input image to predict image-plane positions of human body keypoints, namely neck, wrists, elbows, knees, and ankles. A family of models tackle this problem by generating a sequence of confidence maps for each body part, and predicting the location of each keypoint as maximally-confident coordinates~\cite{DBLP:journals/corr/NewellYD16,cao2017realtime}. Even though these models can be trained using smooth losses such as the Mean Squared Error (MSE), in contrast the standard evaluation metric, namely the Percentage of Correctly-detectable Keypoints (PCKh), rejects all hypotheses that are not within an epsilon neighborhood of ground truth positions. Despite having this non-differentiable $\ell^{eval}$, the Houdini loss can be used, by considering $f(y')$ to be the \textit{product of (normalized) confidences of a set of body part keypoints} $y'$. The authors successfully demonstrated both non-targeted as well as targeted attacks using the Houdini loss with the Fast Gradient Value (FGV) method~\citep{DBLP:journals/corr/RozsaRB16} (see Section~\ref{FGSM}).

The second example focuses on semantic segmentation tasks, which labels each pixel in an input image of a natural scene as one of different semantic classes, such as sky, road, and pedestrian. A number of semantic segmentation models are trained using per-pixel classification cross-entropy loss, while prediction performance is evaluated on across-class mIOU ratios instead. The Houdini loss augments mIOU by comparing the \textit{across-pixel product of class probabilities} between the prediction and ground truth maps. While the Dense Adversarial Generation (DAG) method~\citep{DBLP:journals/corr/XieWZZXY17} (see Section~\ref{DAG}) can attack these models using the per-pixel cross-entropy training loss $\ell^{train}$, \citet{DBLP:conf/nips/CisseANK17} has shown that Houdini-based attacks, both non-targeted and targeted, can find adversaries that achieve equal levels of adversarial damage to mIOU compared to DAG while being more perceptually similar.

The third setup centers around a speech recognition task, where the goal is to predict a text transcript given an input audio spectrogram containing speech segments. Although standard evaluation for speech recognition focuses on Word Error Rate (WER), which is non-differentiable, modern models are trained using a specialized Connectionist Temporal Classification (CTC) loss. Thankfully, since this model is trained to maximize the likelihood of the conditional probabilities of a sequence of text labels $l$ given an input spectrogram $x$, the Houdini attack loss can combine this smooth probability map with the non-smooth WER evaluation criterion. The authors have thus successfully demonstrated non-targeted attacks, targeted attacks, and even blackbox attacks based on this surrogate loss function $\ell_H$.

\subsection{Momentum Iterative Fast Gradient Sign Method} \label{MIFGSM} 

\citet{Dong_2018_CVPR} proposed a family of improvements to BIM and ILLCM~\citep{DBLP:journals/corr/KurakinGB16,45816} (see Sections~\ref{BIM} and~\ref{ILLCM}) called the Momentum Iterative Fast Gradient Sign Method (MI-FGSM) by leveraging momentum during optimization process. Momentum terms have been frequently used to accelerate the convergence of iterative optimization processes~\citep{POLYAK19641}. When applied to optimization-based adversarial attacks such as FGSM, the authors argued that the use of momentum helps to stabilize the update directions for perturbations, and also helps to escape from weak local maxima and thus resulting in an increase in the adversarial example's transferability. This work extends upon a previous observation made by~\cite{45816} that adversarial examples generated by BIM are less transferable compared to FGSM~\citep{43405} (see Section~\ref{FGSM}), i.e., stronger adversaries are typically less transferable.

The algorithm for non-targeted variant of MI-FGSM is shown in Algorithm~\ref{alg:mifgsm}. Note that the $Clip \{\cdot \}$ function is the same as the one defined in Equation~\ref{equations:clip} (see Section~\ref{BIM}). In particular, $g_i$ is a (scaled) exponential moving average term that accumulates the \emph{directions} of gradients rather than their raw values, hence the use of a normalizing denominator $||\nabla_{x'_i} \mathcal{L}(x'_i, y)||_1$. This is used to smoothly perturb the adversarial example over iterations, and to prevent strongly oscillatory updates on $x'_i$.

\begin{algorithm}[h]
   \caption{Non-targeted MI-FGSM}
   \label{alg:mifgsm}
\begin{algorithmic}
   \STATE {\bfseries Input:} original input $x$, ground truth label $y$, targeted model $f$, step size $\alpha$, max. number of iterations $I_{max}$, perturbations size $\epsilon$, decay factor $\mu$
   \STATE {\bfseries Output:} adversarial input $x'$
   \STATE {\bfseries Initialize:} $\alpha=\frac{\epsilon}{I_{max}}$, gradients accumulator $g_0=0$
   \FOR{$i=0$ \TO $I_{max}$}
   \STATE $g_{i+1} \leftarrow \mu \cdot g_i + \frac{\nabla_{x'_i} \mathcal{L}(x'_i,y)}{||\nabla_{x'_i} \mathcal{L}(x'_i, y)||_1}$
   \STATE $x'_{i+1} \leftarrow Clip_{\epsilon} \Big\{x'_i + \alpha \cdot \textrm{sign}(g_{i+1})\Big\}$
   \ENDFOR
   \STATE {\bfseries Return:} $x'_{I_{max}}$
\end{algorithmic}
\end{algorithm}


To further increase transferability and robustness of adversaries, an Ensemble MI-FGSM variant can be used to attack multiple models for the same task (e.g., having the same input and output representations). To achieve consensus, logits from each of the $M$ victim models are combined together, e.g., via $Z^{ens}(x) = \sum_{m=1}^{M}Z^m(x)$. The ensemble loss function $\mathcal{L}^{ens}(x,y)$ can then be computed using the softmax of $Z^{ens}(x)$.



Yet another variant, MI-FGM, can be constructed by varying how the underlying FGSM works. Rather than constraining FGSM's perturbations by an $\infty$-norm bound, the authors propose to instead use an $L_2$ constraint:

\begin{equation} \label{equations:fgm}
x'_{i+1} \leftarrow x'_i + \alpha \cdot \frac{\nabla_{x'_{i}} \mathcal{L}(x'_{i},y)}{||\nabla_{x'_{i}} \mathcal{L}(x'_{i},y)||_2},
\end{equation}

\noindent which they called as the Fast Gradient Method (FGM). Note that this is nearly identical to the Fast Gradient Value (FGV) method proposed by~\citep{DBLP:journals/corr/RozsaRB16}, which applies a step size $\alpha$ to the non-normalized gradient $\nabla_x \mathcal{L}(x,y)$. Similar to the change from FGSM to FGM, the MI-FGM attack is formulated as:

\begin{equation} \label{equations:generalized_mifgm}
x'_{i+1} \leftarrow Clip_{\epsilon} \Big\{x'_i + \alpha \cdot \frac{g_{i+1}}{||g_{i+1}||_2}\Big\}.
\end{equation}

\noindent Finally, also note that all of the above methods can be turned into targeted attacks by changing the target class from $y$ to $t$ and negating the sign of the perturbation, similar to how BIM can be extended to ILLCM.


\citet{Dong_2018_CVPR} showed that MI-FGSM enhances the transferability of the adversarial examples compared to both FGSM and BIM when evaluated on various ImageNet models. Furthermore, the use of momentum is argued to let MI-FGSM to have more stable updates. This is substantiated empirically by showing that the attack success rate of MI-FGSM plateaus as the number of iterations increases, while the attack success rate of BIM \textit{decreases} under the same setup. The authors thus argue that BIM easily \enquote{overfits} the victim model, which causes the adversaries to be less transferable. Additionally, the Ensemble MI-FGSM variant was shown to further increase the transferability rate, and achieved higher attack success rates against models defended using ensemble adversarial training~\citep{tramèr2018ensemble} (see Section~\ref{ENSEMBLE_ADV_TRAINING}). Finally, MI-FGSM attained first-place performances at the NIPS 2017 Competition on Adversarial Attacks and Defenses, under both the targeted and non-targeted attack categories~\citep{DBLP:journals/corr/abs-1804-00097}.

\subsection{AdvGAN} \label{ADVGAN} 

\begin{figure}[h]
\centering
\includegraphics[width=0.7\textwidth]{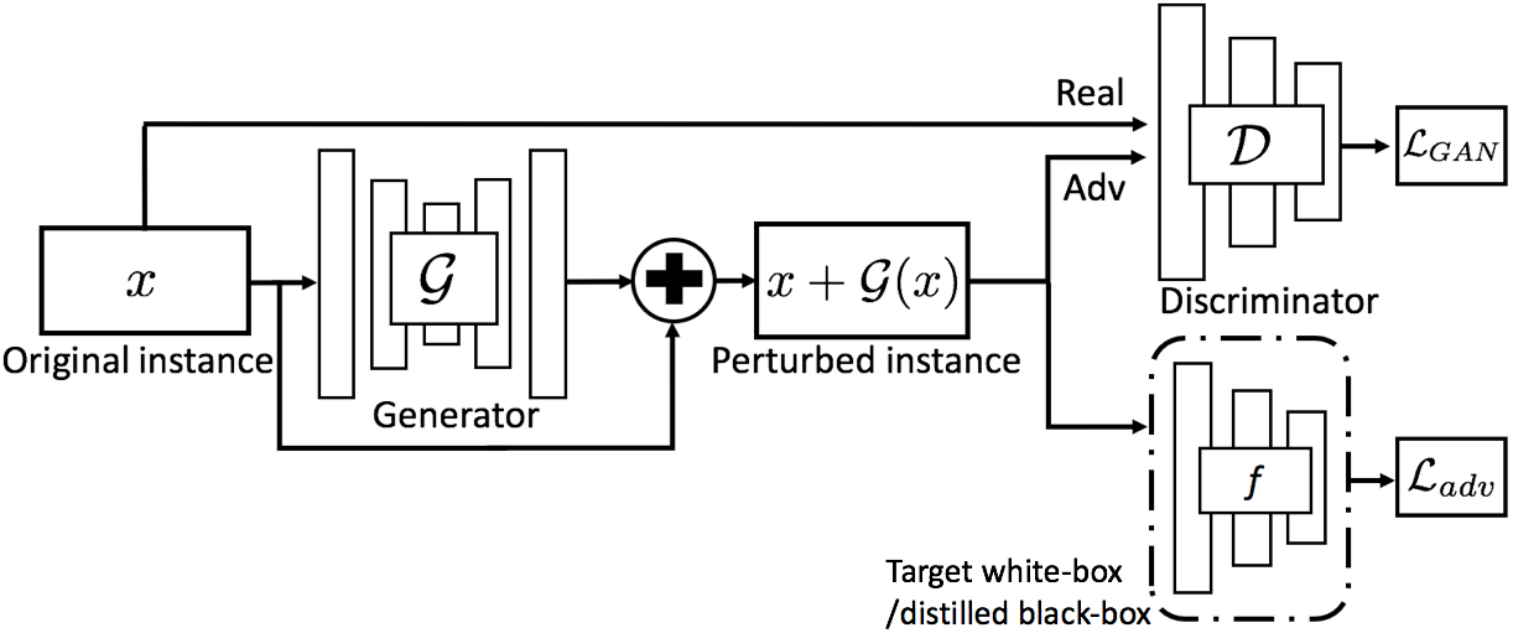}
\caption{Illustration of AdvGAN (taken from~\citet{xiao2018generating}).}
\label{fig:advgan}
\end{figure}

Similar to P-ATN~\citep{46527} (see Section~\ref{ATN}), AdvGAN~\citep{xiao2018generating} aims to generate adversarial perturbations by using a neural network. The generator $G$ in AdvGAN outputs perturbations $r$ which when added to the original instance $x$ will result in adversarial examples $x'$. The difference between AdvGAN and P-ATN is that AdvGAN adds a discriminator network $D$, and trains the generator adversarially as a Generative Adversarial Network (GAN)~\citep{NIPS2014_5423} framework. Fig.~\ref{fig:advgan} illustrates the proposed method. In contrast, whilst P-ATN also employs a generator network, it does not rely on a real/synthetic discriminator for training, and rather uses a custom loss (e.g., $L_2$) for learning perturbations.

The objective function for the generator consists of several terms, including a classifier loss ($\mathcal{L}_{adv}$), an anti-discriminator GAN loss ($\mathcal{L}_{GAN}$), and a hinge loss ($\mathcal{L}_{hinge}$):

\begin{equation} \label{equations:advgan_total_loss}
\mathcal{L} = \mathcal{L}_{adv} + \alpha \mathcal{L}_{GAN} + \beta \mathcal{L}_{hinge},
\end{equation}

\noindent where

\begin{equation} \label{equations:advgan_adv_loss}
\mathcal{L}_{adv} = \mathbb{E}_{x}[\mathcal{L}_f(x+G(x),t)],
\end{equation}

\begin{equation} \label{equations:advgan_gan_loss}
\mathcal{L}_{GAN} = \mathbb{E}_{x}[\log D(x)] + \mathbb{E}_{x}[\log(1 - D(x + G(x)))],
\end{equation}

\begin{equation} \label{equations:advgan_hinge_loss}
\mathcal{L}_{hinge} =  \mathbb{E}_{x}[\max(0,||G(x)||_2 - c)].
\end{equation}

\noindent Here, $\alpha$ and $\beta$ are constants that control the weighting of the loss terms, $\mathcal{L}_f$ is the classification loss function used by $f$ (e.g., cross-entropy), $t$ is the target misclassification label, and $c$ is a constant that represents the bound for the hinge loss. These terms together contribute to help generate perturbations that are adversarial (classifier loss), perceptually similar to the real data distribution (GAN loss), and bounded in magnitude (hinge loss).

AdvGAN was shown to achieve greater performance over FGSM~\citep{43405} and C\&W~\citep{DBLP:journals/corr/CarliniW16a} methods when challenged with models defended via adversarial training~\citep{43405} (see Section~\ref{ADV_TRAINING}), ensemble adversarial training~\citep{tramèr2018ensemble} (see Section~\ref{ENSEMBLE_ADV_TRAINING}), and PGD adversarial training~\citep{madry2018towards} (see Section~\ref{PGD_ADV}). Also, as a preliminary assessment, the authors evaluated AdvGAN with a modified loss function, $\mathcal{L} = \mathcal{L}_{adv} + ||G(x)||_2$, which is more similar to that of P-ATN, and found that AdvGAN's original loss function attained superior results when tested on the MNIST dataset. Finally, the authors performed a human perceptual study following the procedure described in~\cite{zhang2016colorful,Isola_2017_CVPR} and concluded that the adversarial examples generated by AdvGAN are perceptually-indistinguishable from real images.

One potential drawback of the presented experiments was that only FGSM adversaries were included in the adversarial training setups. Thus, while AdvGAN was shown to be capable of finding adversaries that circumvent defense methods that stymie FGSM and C\&W methods, these may not generalize to broader setups comparing against other baseline adversarial attack and defense methods.

\subsection{Boundary Attack} \label{BOUNDARY} 

Boundary attack~\citep{brendel2018decisionbased} is a blackbox method for finding an adversarial example resembling a source image $x$.
It works by iteratively perturbing another \enquote{seed} image $x'_0$, belonging to a different class than $x$, towards and along the decision boundaries between the corresponding class of $x$ and adjacent classes.
This method does not utilize the victim model's input Jacobian $\nabla_{x}f$, and optimizes instead using a rejection sampling formulation that merely queries the model on various perturbed images, $f(x'_i), i \in \{0,1,2,...\}$.

This attack can either be targeted or non-targeted, with the sole difference being how the \enquote{adversarial} condition is defined.
For example, for the non-targeted attack variant, one can for example start from a randomly-sampled image whose class prediction is different from that of $x$ (i.e., adversarial).
The boundary attack process perturbs this adversary first towards and then along the decision boundary bordering the ground truth class of the source image, until ultimately the perturbed adversary minimizes the perceptual difference $d(x,x'_i)$ (e.g., $L_2$ metric) with respect to $x$.

\begin{algorithm}[ht]
   \caption{Targeted Variant of Boundary Attack~\citep{brendel2018decisionbased}}
   \label{alg:boundary_attack}
\begin{algorithmic}
   \STATE {\bfseries Input:} ($D$-dimensional) source input $x$, seed input $x'_0$ with adversarial class, victim model $f$, initial hyperparameters $\epsilon_0$ and $\delta_0$, number of stochastic samples $K$, max number of iterations $I_{max}$
   \STATE {\bfseries Initialize:} $i \leftarrow 1$, $\epsilon \leftarrow \epsilon_0, \delta \leftarrow \delta_0$
   \WHILE{$i \leq I_{max}$}
   \STATE $isAdversarial = False$
   \WHILE{not $isAdversarial$}
   \STATE $O, A \leftarrow \varnothing, \varnothing$ 
   \FOR{$k=1$ to $K$}
   \STATE $\eta \sim \mathcal{N}(0,1)^D$ 
   \STATE $x^{\delta}_{i,k} \leftarrow x'_{i-1} + \frac{\delta}{||\eta||_2} \cdot \eta$ 
   \STATE $x^{\perp}_{i,k} \leftarrow x + \frac{||x_{i-1}'-x||_2}{||x^{\delta}_{i,k} - x||_2} (x^{\delta}_{i,k} - x)$ 
   \STATE $x^{\perp}_{i,k} \leftarrow \min(\max(x^{\perp}_{i,k}, 0), 1)$ 
   \STATE $x^{\epsilon}_{i,k} \leftarrow x^{\perp}_{i,k} - \epsilon (x^{\perp}_{i,k} - x)$ 
   \STATE $x^{\epsilon}_{i,k} \leftarrow \min(\max(x^{\epsilon}_{i,k}, 0), 1)$ 
   \STATE $O \leftarrow O \cup x^{\perp}_{i,k}$
   \STATE $A \leftarrow A \cup x^{\epsilon}_{i,k}$
   \IF {$f(x^{\epsilon}_{i,k}) == f(x'_0 )$} 
   \STATE $isAdversarial = 1$
   \ENDIF
   \ENDFOR
   \IF {$isAdversarial == 1$}
   \STATE $x'_i \leftarrow \arg\min_{x' \in A, f(x') == f(x'_0)} ||x' - x||_2$ 
   \ELSE
   \STATE $x'_i \leftarrow x'_{i-1}$
   \ENDIF
   \STATE adaptively update $\delta, \epsilon$ using $O, A$
   \ENDWHILE
   \STATE $i = i + 1$
   \ENDWHILE
   \STATE {\bfseries Return:} $x'_i$
\end{algorithmic}
\end{algorithm}



\begin{figure}[h]
\centering
\includegraphics[width=0.7\textwidth]{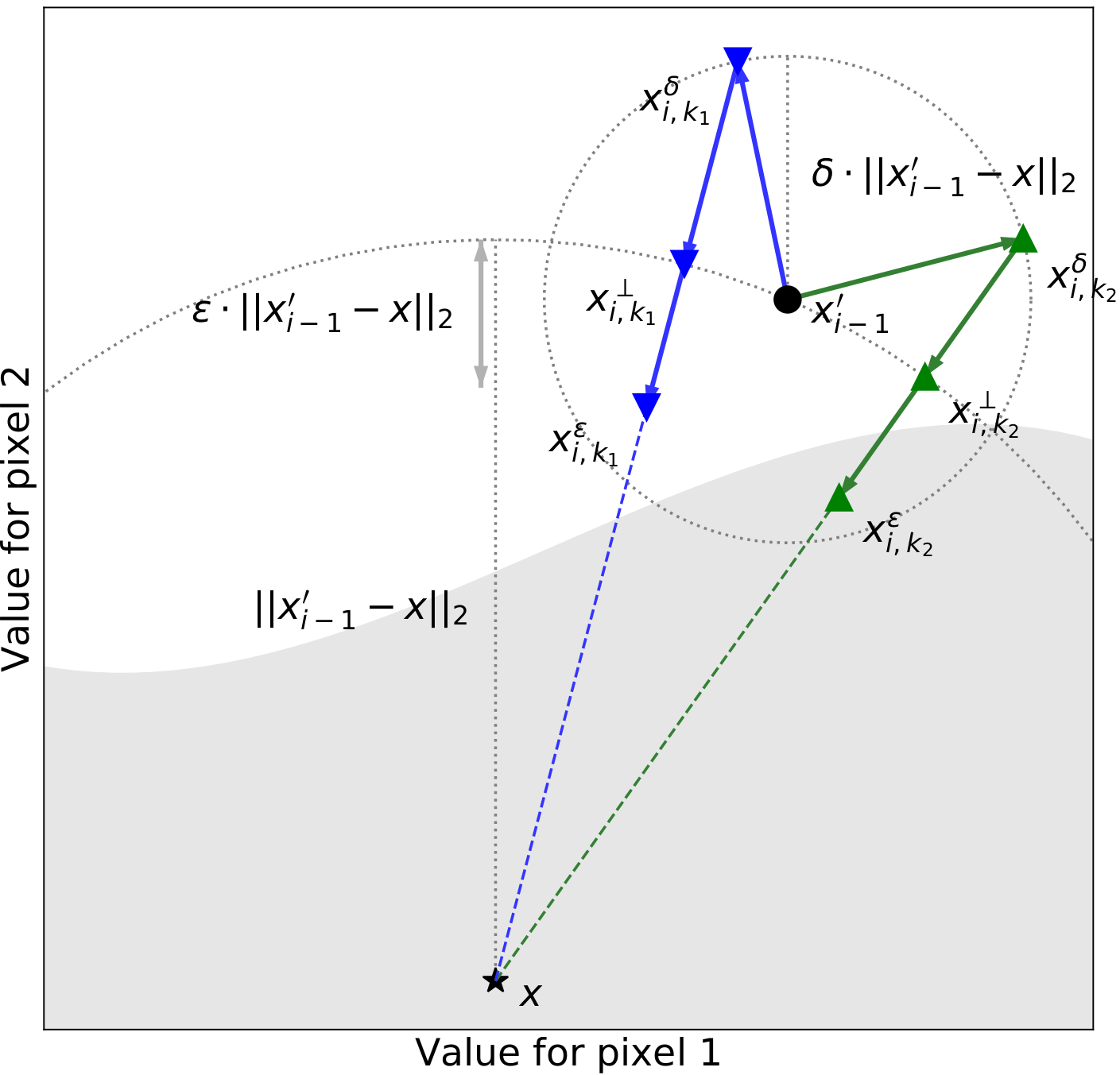}
\caption{One step of the boundary attack process.}
\label{fig:boundary_attack_intuition}
\end{figure}

Each iteration $i$ of the boundary attack process in Algorithm~\ref{alg:boundary_attack} is depicted geometrically in Figure~\ref{fig:boundary_attack_intuition}.
In essence, each iteration samples $K$ perturbed inputs from $x_{i-1}'$, and is importantly driven by two parameters $\delta,\epsilon$.
The $k$-th sample begins by updating $x_{i-1}'$ towards a randomly-sampled direction, and by a distance of $\delta$, to obtain $x^{\delta}_{i,k}$.
Then, $x^{\delta}_{i,k}$ is projected to be at the same distance of $x_{i-1}'$ away from the source image $x$, resulting in $x^{\perp}_{i,k}$.
Note that the update from  $x_{i-1}'$ to $x^{\perp}_{i,k}$ points along a direction that is orthogonal to the gap between the current candidate $x_{i-1}'$ and the source image $x$.
Next, the candidate is moved towards $x$ by a relative distance of $\epsilon$, yielding $x^{\epsilon}_{i,k}$.
In practice, both $x^{\perp}_{i,k}$ and $x^{\epsilon}_{i,k}$ are capped to be within the range of valid inputs.
Finally, the closest $x^{\epsilon}_{i,k}$ to $x$ that satisfies the adversarial condition is set as the latest candidate adversarial input $x_{i}'$; if none of the samples are adversarial, then this iteration is repeated again with $K$ new samples.

This boundary attack efficiently samples new perturbations based on the current local landscape of decision boundaries. This is achieved by adaptively updating the two parameters $\delta,\epsilon$ during each attack iteration. In particular, $\delta$ controls the spatial range to sample orthogonal perturbations $x^{\perp}_{i,k} \in O$, i.e., perturbations that are equidistant to $x$ just like $x'_{i-1}$. Ideally, $\delta$ should be sufficiently large so that the adversarial/non-adversarial decision boundary exhibits a locally linear relationship with respect to the sampling spatial range. This linear relationship can be verified by checking if roughly $50\%$ of the samples in $O$ are adversarial. If this percentage is much lower than $50\%$, then $\delta$ is decreased by $\delta \leftarrow \delta * \delta_{dec}$; otherwise $\delta$ is increased by $\delta \leftarrow \delta * \delta_{inc}$.

Similarly, $\epsilon$ controls the relative distance to move the samples $x^{\perp}_{i,k}$ orthogonally towards $x$. An efficient $\epsilon$ setting should ensure that roughly $50\%$ of the shrunken samples $x^{\epsilon}_{i,k} \in A$ remain adversarial. If the percentage of adversarial shrunken samples is too small, that means that $\epsilon$ took too large of a step, and thus should be decreased by $\epsilon \leftarrow \epsilon * \epsilon_{dec}$. On the other hand, if this percentage is too large, that means that $\epsilon$ can be safely increased, i.e., $\epsilon \leftarrow \epsilon * \epsilon_{inc}$. Theoretically, $\epsilon$ should converge to $0$ when the attack finds a locally optimal adversary.

Due to the brute-force nature of rejection sampling, this adversarial attack may require many iterations (e.g., hundreds of thousands of updates) to find high-quality adversarial examples.
Nevertheless, this blackbox method generally applies to a wide range of real world attack situations, as it merely queries the victim model without needing to know its internals.
The simplicity of the Boundary Attack raises concerns on the security of existing machine learning models since it shows how attackers do not necessarily need a complicated strategy to find adversarial examples.

\citet{brendel2018decisionbased} demonstrated how boundary attack is comparable even to whitebox attacks such as FGSM~\citep{43405} (see Section~\ref{FGSM}), DeepFool~\citep{DBLP:journals/corr/Moosavi-Dezfooli15} (see Section~\ref{DEEPFOOL}), and C\&W attacks~\citep{DBLP:journals/corr/CarliniW16a} (see Section~\ref{CW}).
They also showed that boundary attack can be used to circumvent defensive distillation~\citep{journals/corr/PapernotMWJS15} (see Section \ref{DISTIL}) and gradient masking defenses since boundary attack does not rely on gradient information.

\subsection{Natural Adversarial Attack} \label{NAA-GAN} 

In the computer vision domain, an adversarial example is designed to look like a source image when seen by a human.
However, in the context of natural language, adversaries must further remain coherent, both in terms of grammar (syntax) and in terms of content (semantics).
The need for an adversarial example to be \emph{natural}, i.e., lying within the manifold of natural inputs (be it images, or sentences), is a shortcoming of early adversarial attacks that simply perturb inputs in their raw representation (namely pixel intensities, or characters).

\citet{zhao2018generating} developed a generalized blackbox technique to generate such \emph{natural} adversarial examples, by searching within a pre-learned latent representation space. This way, any perturbed latent point should be able to project into an input in the raw space that adheres to the complex statistics of natural instances, for instance an image of a natural object or a naturally-occurring scene, or a naturally coherent sentence.
We refer to this method as Natural Adversarial Attack (NAA).

Prior to being able to generate adversarial examples, this method first learns a statistic model of natural input instances using the Wasserstein GAN framework~\citep{pmlr-v70-arjovsky17a}. The generator network $G$ in such a formulation maps a latent representation $z$ onto an instance $x$ within the distribution of natural inputs (e.g., images or sentences), i.e., $G:z \mapsto x$. For NAA, one learns an additional inverter network $I$ that maps from $x$ back to $z$, by optimizing

\begin{equation} \label{equations:natural_gan_inverter}
\min_{\gamma} \mathbb{E}_{x \sim p_{real}(x)}||G(I(x)) - x|| + \lambda \cdot \mathbb{E}_{z \sim p_z(z)} ||I(G(z)) - z||,
\end{equation}

\noindent where the hyperparameter $\lambda$ controls the relative weights of two terms, and $\gamma$ denotes the trainable parameters of the inverter.
This loss function is similar to the cycle consistency loss introduced in CycleGAN~\citep{CycleGAN2017}.
Still, NAA differs by first training $G$ (adversarially along with a discriminator network $D$) until convergence, and then separately optimizing the inverter.

After having trained a $G$ and $I$ on an input dataset, a natural adversarial example $x' = G(z')$ can be generated by searching (through various means discussed below) for a mis-classified latent instance $z'$:

\begin{equation} \label{equations:natural_gan_generation}
z' = \arg \min_{\tilde{z}} ||\tilde{z} - I(x)||, \ \textrm{such that } f(G(\tilde{z})) \neq f(x),
\end{equation}

\noindent where $f$ is the victim model.


\begin{figure}[h]
\centering
\includegraphics[width=0.5\textwidth]{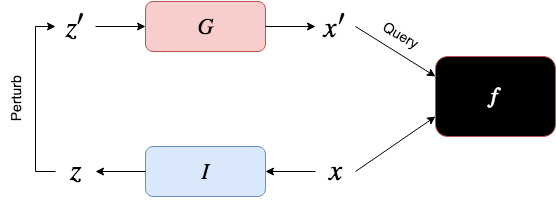}
\caption{Illustration of NAA (adapted from~\citet{zhao2018generating}).}
\label{fig:natural}
\end{figure}

Two methods were proposed for finding $z'$: iterative stochastic search and hybrid shrinking search, as shown in Algorithms~\ref{alg:iterative_stochastic_search} and~\ref{alg:hybrid_shrinking_search}.
The common approach is to perturb the latent representation of the source input $z=I(x)$ by uniformly sampling a vector $\epsilon_i$ within a specific range of magnitudes, and testing whether $G(z+\epsilon_i)$ is adversarial.
For iterative stochastic search, the range of magnitude is initially small, and iteratively grown if no adversaries are found.
More efficiently, hybrid shrinking search first finds an adversarial example within a broad radius around the source input, and then iteratively refines this candidate by searching for smaller-magnitude adversaries.
Empirically, the latter approach is shown to be $4\times$ faster than iterative stochastic search, while producing similar adversarial performance.

\begin{algorithm}[!h]
   \caption{Iterative Stochastic Search~\cite{zhao2018generating}}
   \label{alg:iterative_stochastic_search}
\begin{algorithmic}
   \STATE {\bfseries Input:} source input $x$, victim model $f$, trained generator $G$, trained inverter $I$, search radius increment $\Delta r$, number of perturbations per iteration $N$
   \STATE {\bfseries Output:} adversarial image $x'$
   \STATE {\bfseries Initialize:} $z=I(x)$, $r=0$
   \WHILE {no adversarial examples found}
   \STATE $S \leftarrow \varnothing$
   \FOR{$i = 0$ \TO $N$}
   \STATE $\epsilon_i \sim U\left[ (r, r + \Delta r] \right]$
   \STATE $z'_i \leftarrow z + \epsilon_i$
   \STATE $x'_i \leftarrow G(z'_i)$
   \IF{$f(x'_i) \neq f(x)$}
   \STATE $S \leftarrow S \cup z'_i$ 
   \ENDIF
   \ENDFOR
   \IF{$S == \varnothing$}
   \STATE $r \leftarrow r + \Delta r$
   \ELSE
   \STATE {\bfseries Return:} $\arg \min_{z' \in S}||z' - z||$
   \ENDIF
   \ENDWHILE
\end{algorithmic}
\end{algorithm}

\begin{algorithm}[!h]
   \caption{Hybrid Shrinking Search~\citep{zhao2018generating}}
   \label{alg:hybrid_shrinking_search}
\begin{algorithmic}
   \STATE {\bfseries Input:} source input $x$, victim model $f$, trained generator $G$, trained inverter $I$, maximum search radius $r$, search radius increment $\Delta r$, number of perturbations per iteration for initial search $N$, number of perturbations for refinement $K$, shrink factor $l$
   \STATE {\bfseries Output:} adversarial image $x'$
   \STATE {\bfseries Initialize:} $z=I(x)$, $l = 0$
   \WHILE {$r - l \geq \Delta r$}
   \STATE $S \leftarrow \varnothing$
   \FOR{$i = 0$ \TO $N$}
   \STATE $\epsilon_i \sim U\left[ (l, r] \right]$
   \STATE $z'_i \leftarrow z + \epsilon_i$
   \STATE $x'_i \leftarrow G(z'_i)$
   \IF{$f(x'_i) \neq f(x)$}
   \STATE $S \leftarrow S \cup z'_i$ 
   \ENDIF
   \ENDFOR
   \IF{$S == \varnothing$}
   \STATE $l \leftarrow \frac{l+r}{2}$
   \ELSE
   \STATE $z' \leftarrow \arg \min_{z' \in S}||z' - z||$
   \STATE $l \leftarrow 0$
   \STATE $r \leftarrow ||z' - z||$
   \ENDIF
   \ENDWHILE
   \WHILE{$k < K$ \AND $r > 0$}
   \STATE $S \leftarrow \varnothing$
   \STATE $l \leftarrow \max(0, r - \Delta r)$
   \FOR{$i = 0$ \TO $N$}
   \STATE $\epsilon_i \sim U\left[ (l, r] \right]$
   \STATE $z'_i \leftarrow z + \epsilon_i$
   \STATE $x'_i \leftarrow G(z'_i)$
   \IF{$f(x'_i) \neq f(x)$}
   \STATE $S \leftarrow S \cup z'_i$ 
   \ENDIF
   \ENDFOR
   \IF{$S == \varnothing$}
   \STATE $k \leftarrow k + 1, r \leftarrow r - \Delta r$
   \ELSE
   \STATE $z' \leftarrow \arg \min_{z' \in S}||z' - z||$
   \STATE $k \leftarrow 0$
   \STATE $r \leftarrow ||z' - z||$
   \ENDIF
   \ENDWHILE
   \STATE {\bfseries Return:} $z'$
\end{algorithmic}
\end{algorithm}

For a handwritten digit classification task, the authors demonstrated that natural adversarial examples perturbed pixels around relevant image features, instead of injecting arbitrarily-located noise. Furthermore, NAA successfully generated adversarial texts with minimal grammatical errors on both textual entailment and machine translation tasks where their models were trained on the Stanford Natural Language Inference (SNLI) dataset~\citep{DBLP:journals/corr/BowmanAPM15}. 

\subsection{Spatially Transformed Adversarial Attack} \label{STA} 

Perceptual similarity is an important attribute for adversarial examples in many cases, as ideally they should be indistinguishable from natural images when viewed by humans.
However, many adversarial attacks rely on $L_2$ or $L_{\infty}$ distances in pixel intensities to quantify perceptual similarity, which may not be an ideal metric~\citep{4775883}.
Spatially Transformed Adversarial Attack (stAdv)~\citep{xiao2018spatially} instead optimizes for \emph{geometrical} similarity between a source and adversarial image pair.

To do this, the authors leverage the representation of a flow field $f(\Delta u, \Delta v) \in \mathbb{R}^{H \times W \times 2}$ to represent a spatially-transformed version of a given image.
Each element in the flow field represent a change in horizontal and vertical coordinates away from a designated sampling location, and so the transformed image can be constructed by sampling each pixel at these changed coordinates.
Sampling is done using a differentiable implementation of bilinear interpolation~\citep{NIPS2015_5854,zhou2016view};
first, let $\textrm{UV}(u,v) \in \mathbb{R}^{H \times W \times 2}$ contain spatial coordinates of each pixel in the input $x$:

\begin{equation} \label{equations:stadv_UV}
\textrm{UV}' = \begin{bmatrix} 
    (u_{1}=1,v_{1}=1) & \dots & (u_{1}=1,v_{W}=W) \\
    \vdots & \ddots & \vdots \\
    (u_{H}=H,v_{1}=1) & \dots & (u_{H}=H,v_{W}=W) 
    \end{bmatrix}.
\end{equation}

\noindent Then, by defining $\textrm{UV}' = \textrm{UV} - f$, we can sample the $i$-th element of the modified image $x'$ around the 4-pixel neighborhood $
\mathcal{N}$ of its (real-valued) spatially-transformed coordinate $(u_i,v_i) = (u_i',v_i') + (\Delta u_i, \Delta v_i)$:

\begin{equation} \label{equations:stadv_bilinear}
x'_i = \sum_{n \in \mathcal{N}(u,v)} x_{n} (1 - |u_i - u_n|) (1 - |v_i - v_n|).
\end{equation}

\noindent Note that all operations in the equations above are differentiable, as well as finding floored/ceiling-ed neighboring coordinates of a given spatially-transformed coordinate.
Furthermore, actual pixel values of each neighbor, $x_n$, can be pre-computed beforehand, which makes this entire process differentiable.


stAdv searches for adversaries using this flow field representation by minimizing:

\begin{equation} \label{equations:stadv_objective}
\mathcal{L}_{flow}(f) + \alpha \cdot \mathcal{L}_{adv}(x',t)
\end{equation}

\noindent where $\alpha$ is the weight factor, $\mathcal{L}_{adv}(x',t)$ is the classification loss (e.g., the C\&W loss function as in Equation~\ref{equations:cw_loss} in Section~\ref{CW}), and $\mathcal{L}_{flow}(\cdot)$ is a Total Variation loss~\citep{RUDIN1992259} that encourages small and smooth perturbations.
In their experiments, the authors used L-BFGS~\citep{Liu1989} as the optimizer.




\begin{figure}[h]
\centering
\includegraphics[width=0.6\textwidth]{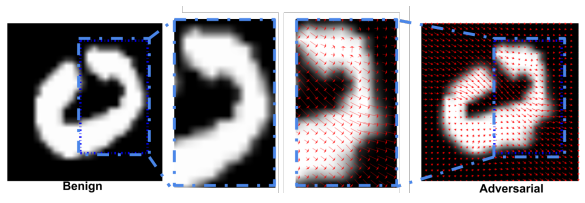}
\caption{Illustration of spatially transformed adversarial example that is misclassified as the digit ``2'' where the red arrows show how the pixels are moved from benign to adversarial image (taken from~\citet{xiao2018spatially}).}
\label{fig:spatial_adversarial}
\end{figure}

Adversarial examples generated by stAdv were shown to be perceptually indistinguishable from the original example by a human, following the procedure in~\cite{zhang2016colorful,Isola_2017_CVPR}.
The authors also evaluated stAdv against a variety of methods that rely upon $L_p$ norm as the similarity metric, such as FGSM~\citep{43405} and C\&W~\citep{DBLP:journals/corr/CarliniW16a}, evaluating against defenses including adversarial training~\citep{43405} (see Section~\ref{ADV_TRAINING}), ensemble adversarial training~\citep{tramèr2018ensemble} (see Section~\ref{ENSEMBLE_ADV_TRAINING}), and PGD adversarial training~\citep{madry2018towards} (see Section~\ref{PGD_ADV}).
In this experiment, stAdv was found to be more effective than $L_p$ norm attacks by a large margin.
Nevertheless, since the authors only considered baseline adversaries generated using FGSM, it remains to be seen if stAdv can produce stronger adversaries than other attacks such as C\&W.
Still, we conclude that the use of different perceptual metric may produce different types of adversaries, which can be used to attack defenses that are designed to be robust against $L_p$ adversaries.

\subsection{Expectation Over Transformation} \label{EOT} 

Early adversarial examples typically have lessened adversarial effects when the source input is transformed, e.g., when a source image is rotated or translated.
In order to create adversaries that are robust to such transformations,~\citet{pmlr-v80-athalye18b} proposed a method called the Expectation Over Transformation (EOT).
Such a robustness attribute is especially important for physical-realizable adversarial examples, e.g., images that are printed or displayed in the real-world and imaged under from different view angles (see Section~\ref{PHYSICAL} for more details).

EOT is a meta-optimization process that can be coupled with various optimization-based attacks.
The key addition is to pre-process each input instance $x$ by one or more parameterized transforms $t(x)$, and to optimize the expectation of the objective function over the probability distributions $T$ of the transformation parameters:

\begin{equation} \label{equations:eot}
\arg \min_{x'} \mathbb{E}_{t \sim T} \bigg[ \mathcal{L}_{adv}(t(x')) + \mathcal{L}_{perc}(x, x') \bigg],
\end{equation}

\noindent where $\mathcal{L}_{adv}$ represents an adversarial loss (e.g., to minimize the distance between model predictions and adversarial labels, or to maximize the model's training error), and $\mathcal{L}_{perc}$ entails a perceptual similarity loss (comparing either the raw source versus perturbed inputs, or their transformed representations).
In the reported experiments, perceptual similarity was measured as Euclidean distance in the CIE Lab space, which is designed to correspond more closely to human perceptual distance compared to RGB (while still remaining differentiable)~\citep{doi:10.1111/j.1478-4408.1976.tb03301.x}.

The authors demonstrated robust adversarial misclassification of generated 2D examples under transformations in the form of $t(x) = Ax + b$ such as rescaling, rotation, and additive Gaussian noise.
They also realized robust physical adversaries using 3D printing, and by optimizing the surface texture of a 3D object under pose-based translations.
Beyond this initial work, EOT opened up a new research direction towards adversarial examples that are physically realizable.
Subsequent work from~\citet{DBLP:journals/corr/abs-1712-09665} used EOT to generate printable adversarial patch~\citep{DBLP:journals/corr/abs-1712-09665} (see Section~\ref{PATCH}).
EOT has also been extended to generate adversarial examples that fool Faster R-CNN~\citep{NIPS2015_5638} by performing EOT over region proposals~\citep{DBLP:journals/corr/abs-1804-05810}.

\subsection{Backward Pass Differentiable Approximation} \label{BPDA} 

Some defense strategies to adversarial attacks rely on mechanisms that obfuscate a model's gradients, for instance by using non-differentiable pre-processing functions or non-differentiable network layers.
Backward Pass Differentiable Approximation (BPDA)~\citep{obfuscated-gradients} is a technique for modifying an adversarial attack to bypass such defenses, by exploiting a differentiable approximation for the defended model to obtain meaningful adversarial gradient estimates.

A simplistic special case of BPDA applies ideas from the straight-through estimator~\citep{DBLP:journals/corr/BengioLC13}: assume that an adversarial defense pre-processes each input $x$ using a non-differentiable transformation $g(x)$, and that $g(x) \approx x$ within some local neighborhood (e.g., $g(\cdot)$ is a smoothing function).
Then, during the adversarial attack, one can simply approximate its gradient by the identity function.
Examples of defense methods that can be circumvented by this technique include bit-depth reduction, JPEG compression, total variance minimization, and quilting-based defense~\citep{guo2018countering} (see Section~\ref{TRANSFORM-DEFENSE}).

More generally, BPDA works by finding a differentiable approximation for a non-differentiable pre-processing transformation $g(\cdot)$ or a non-differentiable network layer, possibly via an engineered guess.
As long as the two functions are similar, adversarial perturbations can still be found by using inexact derivatives from the approximated form during back-propagation, albeit likely needing more attack iterations.
Still, the authors noted that forward inference should still use the original non-differentiable transformation $g(\cdot)$ to obtain true loss values, otherwise the attack may become significantly less effective.

As a concrete example, the authors considered the defense method of thermometer encoding~\citep{buckman2018thermometer} (see Section~\ref{THERMOMETER}), which obfuscates input gradients by discretizing interval-bounded input values $x_{i,j,c} \in [0,1]$ (such as pixel intensities) via thermometer encoding:

\begin{equation} \label{equations:thermometer_encoding_bpda}
  \tau(x_{i,j,c})_k =
  \begin{cases}
  1 & \text{if $x_{i,j,c} \geq \frac{k}{l}$} \\
  0 & \text{otherwise}
  \end{cases},
\end{equation}

\noindent where $k \in \{1,...,l\}$ is the index of the binarized, and $l$ denotes the discretization level (e.g., if $l = 10$, then $\tau(0.66) = `1111110000'$, indexed increasingly from left to right).
This encoding function can be approximated as:

\begin{equation} \label{equations:thermometer_encoding_approximation}
  \hat{\tau}(x_{i,j,c})_k = \min(\max(x_{i,j,c} - \frac{k}{l}, 0), 1),
\end{equation}


\noindent since $\tau(x_{i,j,c})_k = \textrm{ceil}(\hat{\tau}(x_{i,j,c})_k)$.
By using $\hat{\tau}$ during back-propagation,~\citet{obfuscated-gradients} successfully applied an optimization-based adversarial attack to confuse a thermometer-encoded network.

\subsection{Simultaneous Perturbation Stochastic Approximation Attack} \label{SPSA} 

Simultaneous Perturbation Stochastic Approximation (SPSA)~\citep{119632} is an alternative numerical gradient estimator that is more efficient than the Finite Difference (FD) method.
Given a function $f(x)$ where $x \in \mathbb{R}^D$, recall that FD estimates the gradient using $2D$ elementary-vector queries:

\begin{equation}
    \left( \frac{\partial f}{\partial x} \right)_i \approx \left( \frac{f(x+\Delta e_i) - f(x-\Delta e_i)}{2\Delta} \right)_i,
\end{equation}

\noindent where the elementary vector $e_i=[0,...,1,...,0]$ only has a non-zero element at the $i$-th index, and $\Delta$ is a small constant (e.g., $\Delta = 10^{-9}$).
In contrast, SPSA estimates the gradient using $2$ queries to $f(\cdot)$, regardless of $D$, by simultaneously perturbing all dimensions using a vector $v \sim \{-1,1\}^D$ sampled from the Rademacher distribution (i.e., whose elements are either +1 or -1):

\begin{equation}
    \frac{\partial f}{\partial x} \approx \frac{f(x+\Delta v) - f(x-\Delta v)}{2\Delta} \cdot v.
\end{equation}

\citet{pmlr-v80-uesato18a} showed that SPSA can equally be used in a blackbox attack, similar to how the Zeroth-Order Optimization Attack (ZOO-Adam)~\citep{Chen:2017:ZZO:3128572.3140448} (see Section~\ref{ZOO}) uses FD to numerically estimate gradients of a victim model.
As seen in Algorithm~\ref{alg:spsa}, the SPSA Attack makes $N$ numerical estimates of the gradient of an adversarial loss function $\mathcal{L}$ with respect to the input $x$.
It then perturbs the input along the negative averaged gradient direction, and projects the updated input onto the epsilon-ball $N_{\epsilon}(x_0)$ around the source input $x_0$ to enforce perceptual similarity.
In their experiments, the authors initialized $x_0$ similar to the strategy used in R+FGSM~\citep{tramèr2018ensemble} (see Section~\ref{R+FGSM}), by sampling random perturbed input within $[x_0 - \epsilon, x_0 + \epsilon]$, to help avoid gradient masking defenses.


\begin{algorithm}[h]
   \caption{SPSA~\citep{pmlr-v80-uesato18a}}
   \label{alg:spsa}
\begin{algorithmic}
   \STATE {\bfseries Input:} input $x \in \mathbb{R}^D$, perturbation size $\Delta$, step size $\alpha$, gradient estimate batch size $N$, number of iterations $I$
   \STATE {\bfseries Initialize:} $x_0' \leftarrow x$
   \FOR{$i = 0$ \TO $I-1$}
   \FOR{$n = 1$ \TO $N$}
   \STATE $v_n \sim \{-1,1\}^D$
   \STATE $g_n \leftarrow \frac{\mathcal{L}(x_i' + \Delta v_n) - \mathcal{L}(x_i' - \Delta v_n)}{2\Delta} \cdot v_n$
   \ENDFOR
   \STATE $x_{i+1} \leftarrow x_i' - \frac{\alpha}{N} \sum_{n=1}^N g_n$
   \STATE $x_{i+1}' \leftarrow \arg\min_{x \in N_{\epsilon} (x_0)} ||x_{i+1} - x||$
   \ENDFOR
   \STATE {\bfseries Return:} $x_I$
\end{algorithmic}
\end{algorithm}

SPSA was shown to be effective against PixelDefend~\citep{song2018pixeldefend} (see Section~\ref{PIXELDEFEND}), HGD~\citep{DBLP:journals/corr/abs-1712-02976} (see Section~\ref{HGD}), and Random Resizing and Padding (RRP)~\citep{xie2018mitigating} (see Section~\ref{RESIZE-PAD}).
The authors also found SPSA to be stronger than ZOO, where ZOO was implemented without the feature reduction and importance sampling methods.



\subsection{Decoupled Direction and Norm Attack} \label{DDN} 

Constrained optimization-based attacks such as L-BFGS~\citep{42503} (see Section~\ref{L-BFGS}) and C\&W~\citep{DBLP:journals/corr/CarliniW16a} (see Section~\ref{CW}) often optimize for both the misclassification and perceptual similarity conditions at the same time, possibly using expensive line search to balance between these two objectives.
In contrast, \citet{DBLP:journals/corr/abs-1811-09600} proposed the Decoupled Direction and Norm (DDN) attack method, which decouples the perturbation norm penalty and the direction which causes misclassification.

As seen in Algorithm~\ref{alg:ddn}, each attack iteration begins by making an update onto a cumulative perturbation vector $\delta_{i+1}$, by descending by $\alpha$ following the gradient of a misclassification-based loss function $\mathcal{L}$.
However, to obtain the next perturbed input $x_{i+1}'$, only the direction of $\delta_{i+1}$ is used, while the magnitude of the perturbation is set to an independent variable $\epsilon_i$.
Then, depending on whether $x_{i+1}'$ is adversarial, the adaptive norm $\epsilon_i$ is either increased or decreased accordingly.
After $I$ total iterations, the adversarial example that is most perceptually similar to the source input is returned.

\begin{algorithm}[ht]
   \caption{DDN~\citep{DBLP:journals/corr/abs-1811-09600}}
   \label{alg:ddn}
\begin{algorithmic}
   \STATE {\bfseries Input:} input $x$, victim model $f$, true label $y$, target label $t$ (set $t$ to $y$ for non-targeted attack), perturbation size $\alpha$, adaptive norm factor $\gamma \in [0,1)$, number of iterations $I$
   \STATE {\bfseries Initialize:} $x_0' \leftarrow x$, $\epsilon_{0} \leftarrow (1 + \gamma)$, $\delta_{0} \leftarrow 0$, $X' \leftarrow \varnothing$
   \IF {$t == y$}
    \STATE $\sigma \leftarrow +1$
   \ELSE
   \STATE $\sigma \leftarrow -1$
   \ENDIF
   \FOR{$i = 0$ \TO $I-1$}
   \STATE $g \leftarrow \nabla_{x'_{i}} \mathcal{L}(x'_{i},t)$
   \STATE $g \leftarrow \alpha \frac{g}{||g||_2}$
   \STATE $\delta_{i+1} \Leftarrow \delta_{i} + \sigma g$
   \STATE $x'_{i+1} \leftarrow x + \epsilon_{i}\frac{\delta_{i+1}}{||\delta_{i+1}||_2}$
   \STATE $x'_{i+1} \leftarrow \min(\max(x'_{i+1},0),1)$
   \IF {$f(x'_{i+1})$ is adversarial}
   \STATE $X' \leftarrow X' \cup x'_{i+1}$ 
   \STATE $\epsilon_{i+1} \leftarrow (1-\gamma)\epsilon_{i}$
   \ELSE
   \STATE $\epsilon_{i+1} \leftarrow (1+\gamma)\epsilon_{i}$
   \ENDIF
   \ENDFOR
   \STATE {\bfseries Return:} $\arg \min_{x' \in X'} ||x-x'||_2$
\end{algorithmic}
\end{algorithm}

The authors showed that DDN converged faster (e.g., can be as fast as $100$ iterations) and produced adversarial examples with higher perceptual similarity compared to the C\&W~\citep{DBLP:journals/corr/CarliniW16a} $L_2$ attack (see Section~\ref{CW}) when tested on MNIST, CIFAR-10, and ImageNet datasets.
Additionally, DDN was shown to be able to converge as fast as DeepFool~\citep{DBLP:journals/corr/Moosavi-Dezfooli15} (see Section~\ref{DEEPFOOL}), while producing higher attack success rates and lower perturbation norms.
In the Adversarial Vision Challenge at the 32nd Neural Information Processing Systems (NeurIPS'18), DDN was ranked first in the non-targeted attack category, and ranked third in targeted attack category.

Furthermore, the authors showed that when DDN was used to perform adversarial training~\citep{43405} (see Section~\ref{ADV_TRAINING}), the resulting model robustness surpassed PGD adversarial training~\citep{madry2018towards} (see Section~\ref{PGD_ADV}) for all $\epsilon$ values when evaluated on MNIST and CIFAR10 datasets.
Adversarial training with DDN was ranked third in the NeurIPS 2018 Adversarial Vision Challenge when defending against the blackbox attack scenario.

\subsection{CAMOU} \label{CAMOU} 

\citet{zhang2018camou} proposed a method called CAMOU which involves the use of photo-realistic simulation tool (i.e., Unreal engine) to generate 3D adversarial examples in the simulation. CAMOU is a method to generate adversarial textures which when applied to an object, causes detection models like the Mask R-CNN~\citep{he2017maskrcnn} and YOLOv3~\citep{DBLP:journals/corr/abs-1804-02767} to misdetect the object. In other words, the adversarial texture may act like a \textit{camouflage}, making the object invisible to a detector.

\begin{figure}[h]
\centering
\includegraphics[width=0.6\textwidth]{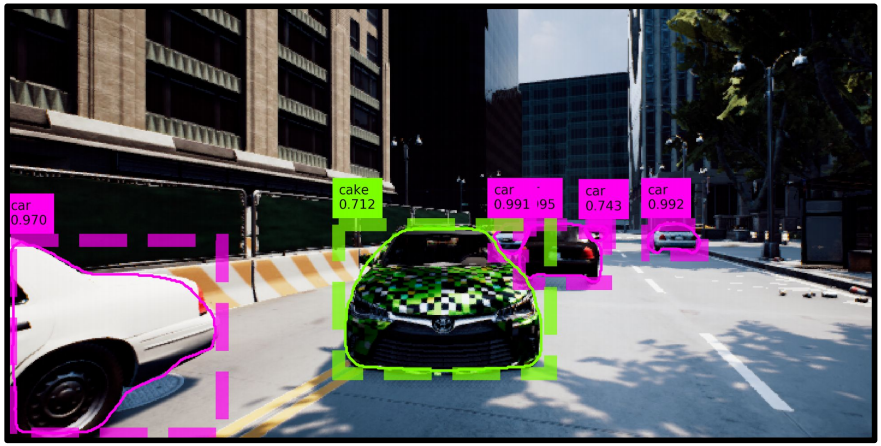}
\caption{The car with camouflage pattern is misdetected as a cake (taken from~\citet{zhang2018camou}).}
\label{fig:camou}
\end{figure}

In this work,~\citet{zhang2018camou} apply the texture on a 3D vehicle model. Given information of the 3D vehicle model and background scene, CAMOU aims to find a camouflage texture $c$ that fools a detector model.~\citet{zhang2018camou} performed their attack in the blackbox scenario where they were constrained from calculating the gradients directly from the target detector model. Concretely, they use a substitute network~\citep{DBLP:journals/corr/PapernotMGJCS16} (see Section~\ref{SUBSTITUTE}) to approximate the output of the target detector, which also helps in bypassing the complex rendering process. EOT~\citep{pmlr-v80-athalye18b} (see Section~\ref{EOT}) is also used to ensure that the adversarial examples are robust to various transformations.

CAMOU works as follows. First, a substitute network $f'$ is created to approximate the detection score of the original model $f$. The substitute network takes the camouflage texture $c$, background scene $b$, and foreground (i.e., the vehicle model) $v$ as inputs, and outputs a scalar value that represents the detection score. Note that the ground truth for the detection score can be obtained by querying the target detection model. Once $f'$ is trained, the adversarial camouflage texture $c'$ can be found using EOT by differentiating through $f'$. Here, the transformations are done by moving the camera to new locations in the simulator, which can simulate various complex 3D transformations. In their implementation,~\citet{zhang2018camou} alternatively trained $f'$ and solved for $c'$. In every iteration, the adversarial camouflage texture $c'$ is added to a set of camouflage pattern used to train the substitute network. The camouflage is also evaluated over various transformations on the target detection model in order to get the best adversarial camouflage texture. These processes are then repeated for specified number of iterations. This method is presented in Algorithm~\ref{alg:camou}.

\begin{algorithm}[h]
   \caption{CAMOU~\citep{zhang2018camou}}
   \label{alg:camou}
\begin{algorithmic}
   \STATE {\bfseries Input:} background scene $b$, foreground/vehicle model $v$, target detection model $f$, substitute detection model $f'$ parameterized by $\theta$, a set of transformation $T$, a set of camouflage texture $C$, number of iteration $I$
   \STATE {\bfseries Output:} adversarial camouflage texture $c'$
   \STATE {\bfseries Initialize:} lowest score achieved by adversary $s^* \leftarrow \infty$
   \FOR{$i = 0$ \TO $I$}
   \STATE $\theta \leftarrow \arg\min_{\theta}\frac{1}{|C||T|}\sum_{c \in C}\sum_{t \in T} \big[ \mathcal{L}\big(f(c,t(b),t(v)),f'(c,t(b),t(v))\big) + \alpha ||\theta||_2\big]$
   \STATE $c' \leftarrow \arg\min_{c} \frac{1}{|T|}\sum_{t \in T}\mathcal{L}(0,f'(c,t(b),t(v)))$
   \STATE $C \leftarrow C \cup c'$
   \STATE $s' \leftarrow \frac{1}{|T|} \sum_{t \in T} f(c',t(b),t(v))$
   \IF {$s' < s^*$}
   \STATE $s^* \leftarrow s'$
   \STATE $c^* \leftarrow c'$
   \ENDIF
   \ENDFOR
   \STATE {\bfseries Return:} $c^*$
\end{algorithmic}
\end{algorithm}

CAMOU successfully fooled Mask R-CNN~\citep{he2017maskrcnn} and YOLOv3~\citep{DBLP:journals/corr/abs-1804-02767} trained on COCO~\citep{10.1007/978-3-319-10602-1_48} dataset in the simulation.~\citet{zhang2018camou} also showed how the adversarial texture can be transfered to fool different detection models to a certain extent. Although~\citet{zhang2018camou} did not print the texture and apply it to a vehicle in the real world, the method offers an important direction to generate adversarial examples in the simulation. In the future, the adversarial examples will be easier to fabricate in the real world as the quality of images produced by the simulation software increases, and as our knowledge on domain adaptation increases.

\pagebreak

\section{Towards Real World Adversarial Attacks} \label{PHYSICAL}

This section discusses the danger of adversarial examples in real world scenarios. Most work on adversarial attacks only test adversarial examples in the virtual domain, raising the question of whether or not these adversarial examples are still effective in the 3D real world environment. Some works presented in this section focus on physically realizable adversarial examples, and others show that adversarial examples need not be physically realizable to be dangerous in real world. We refer physically realizable adversaries as adversarial examples that are still adversarial when viewed in the 3D world from various viewpoints.

\subsection{Printed Adversarial Examples} 

\citet{DBLP:journals/corr/KurakinGB16} studied whether adversarial examples generated using FGSM~\citep{43405} (see Section~\ref{FGSM}), BIM~\citep{45816,DBLP:journals/corr/KurakinGB16} (see Section~\ref{BIM}), and ILLCM~\citep{45816,DBLP:journals/corr/KurakinGB16} (see Section~\ref{ILLCM}) can be printed and remain adversarial in the physical world. This was evaluated by capturing pictures of printed adversarial examples using a cellphone camera, and classified these images with Inception-v3~\citep{44903} model.

In their experiments, they found adversaries generated by FGSM to be the most robust in the physical world, while ILLCM is the least robust. They hypothesized that this is due to the fact that BIM and ILLCM attempt to find more subtle perturbations, which are easier to be destroyed in the real world environment. Further,~\citet{DBLP:journals/corr/KurakinGB16} demonstrated the transferability of the printed adversarial examples by fooling an open source mobile application that performs image classification. Note that the adversarial examples were generated to fool Inception-v3 model, and not the model deployed by the mobile application.

\begin{figure}[h]
\centering
\includegraphics[width=0.6\textwidth]{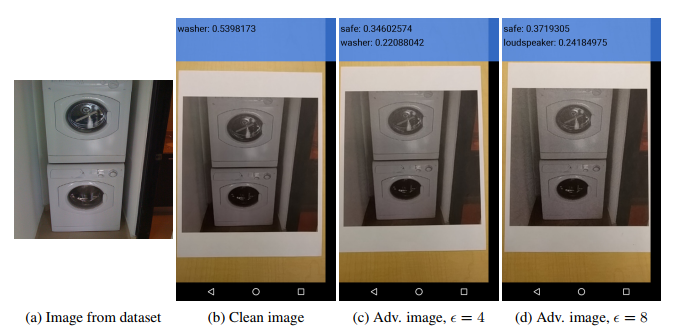}
\caption{Printed adversarial example still fools an image classifier (taken from~\citet{DBLP:journals/corr/KurakinGB16}).}
\label{fig:printed_adversarial}
\end{figure}

The printed adversarial examples generated by this method seem to only work in some limited settings such as when the printed image is located right in front of the camera with small variations in lighting, distance, and camera angle, but they have not been evaluated on more diverse and extreme sets of 2D and 3D transformations. Still, both this work and the adversarial eyeglasses~\citep{Sharif:2016:ACR:2976749.2978392} (see Section~\ref{EYEGLASS}) ignited the discussion of adversarial examples in the physical world.

\subsection{Adversarial Eyeglasses} \label{EYEGLASS} 

\citet{Sharif:2016:ACR:2976749.2978392} demonstrated how a printable adversarial eyeglass can fool face recognition systems. As in the L-BFGS attack~\citep{42503} (see Section~\ref{L-BFGS}),~\citet{Sharif:2016:ACR:2976749.2978392} model this problem as an optimization problem that can be solved using gradient descent. In order to generate an adversary that looks like an eyeglass, a masking matrix $M_x$ is used so that only pixels inside the mask are allowed to change during the optimization process.~\citet{Sharif:2016:ACR:2976749.2978392} noticed the challenges in fabricating the eyeglass as a printer may produce slightly different color when printing. An additional loss term called the Non Printability Score (NPS) was then introduced as part of the optimization process to ensure that the generated adversary is made of colors that are printable by a particular printer. NPS is defined as

\begin{equation} \label{equations:nps}
NPS(\hat{p}) = \prod_{p \in P} |\hat{p} - p|,
\end{equation}

\noindent where $P \subset [0,1]^3$ represents the set of printable RGB triplets, and $\hat{p}$ represents a pixel in the adversarial image. In other words, NPS goes to 0 if $\hat{p}$ is in $P$. In order to get $P$,~\citet{Sharif:2016:ACR:2976749.2978392} took a picture of a printed color palette that contains one fifth of the RGB color space, and assumed the RGB triplets received from the camera to be printable by that particular printer. 
The optimization objective then becomes to find perturbation $r$ that optimizes

\begin{equation} \label{equations:eyeglass}
\arg\min_r \Big( (\sum_{x \in X} \mathcal{L}(f(x + M_x \cdot r), t)) + \lambda_1 \cdot TV(M_x \cdot r) + \lambda_2 \cdot NPS(M_x \cdot r) \Big),
\end{equation}

\noindent where $t$ denotes the target misclassification class, and $\lambda_1$ and $\lambda_2$ are the weighting factors for the objectives.

\begin{figure}[h]
\centering
\includegraphics[width=0.7\textwidth]{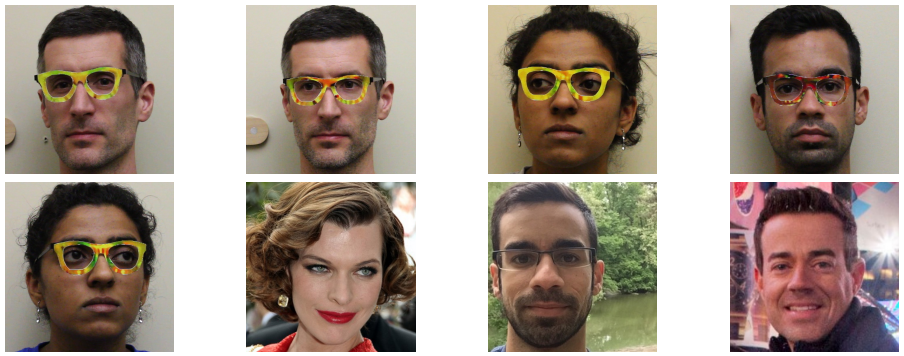}
\caption{Example of the printed adversarial eyeglass (taken from~\citet{Sharif:2016:ACR:2976749.2978392}). The top row shows images of persons that are misclassified as the persons shown in the bottom row. For example, the person in the top row of second column is misclassified as Milla Jovovich.}
\label{fig:printed_eyeglass}
\end{figure}

\citet{Sharif:2016:ACR:2976749.2978392} showed this method can generate adversarial eyeglasses when tested on various face recognition models, including the model proposed by~\cite{Parkhi15} which was the state of the art face recognition model on the Labeled Faces in the Wild dataset (LFW)~\citep{LFWTech,LFWTechUpdate}. They further showed that the eyeglasses could be printed, fooling various models based on the architecture of~\cite{Parkhi15} trained on a dataset which contains pictures of 5 persons from the authors' lab, as well as 5 celebrities from the PubFig dataset~\citep{5459250} (see Fig.~\ref{fig:printed_eyeglass}).

\citet{Sharif17AGNs} extended this work to create adversarial eyeglasses independent of the attacker's face. In order to do this,~\citet{Sharif17AGNs} leverage the GAN~\citep{NIPS2014_5423} framework to train a generator network that generates adversarial eyeglass, where the job of the discriminator is to discriminate between fake and real eyeglasses. The objective of the generator is to minimize a similar loss function in Equation~\ref{equations:eyeglass}). We refer readers to the paper for implementation details.

\subsection{Adversarial Stop Signs} 

\citet{DBLP:journals/corr/EvtimovEFKLPRS17} proposed an attack called the Robust Physical Perturbation (RP$_2$), showing that adversaries can be crafted to be robust to change in angle, distance, and lighting. RP$_2$ was specifically tested on various road signs. Similar to the strategy used to generate adversarial eyeglasses~\citep{Sharif:2016:ACR:2976749.2978392} (see Section~\ref{EYEGLASS}), a masking matrix $M_x$ is used to constrain the area of perturbations. In order to ensure that the adversarial examples generated are robust to change in various conditions and transformations,~\citet{DBLP:journals/corr/EvtimovEFKLPRS17} collected pictures of the same road signs images taken under various conditions, and used them as part of adversarial examples generation process. Formally, the RP$_2$ objective is defined by

\begin{equation} \label{equations:rp2}
\arg\min_r \Big( (\sum_{x \in X} \mathcal{L}(f(x + M_x \cdot r), t)) + \lambda_1 \cdot ||M_x \cdot r||_p + \lambda_2 \cdot NPS(M_x \cdot r) \Big),
\end{equation}

\noindent where $X$ is the set of stop sign images taken in various conditions.

RP$_2$ was used to generate two different types of attacks: poster-printing and sticker attacks. Poster-printing attack is a scenario where an attacker prints a road sign of an actual size on paper which is overlaid on top of a real road sign. Sticker attack is a scenario where an attacker prints some stickers containing the adversarial perturbations and stick them on top of a real road sign. While the masking matrix used in the poster-printing attack allows perturbations to be generated anywhere on the road sign, the mask used in the sticker attack only allows the perturbations to be generated on certain parts of the road sign.

\begin{figure}[h]
\centering
\includegraphics[width=0.2\textwidth]{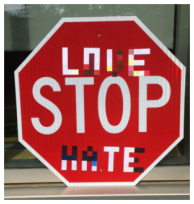}
\caption{An example of sticker attack where it was misclassified as ``Speed Limit 45'' sign when viewed from different viewpoints (taken from~\citet{DBLP:journals/corr/EvtimovEFKLPRS17}).}
\label{fig:printed_stop_sign}
\end{figure}

\citet{DBLP:journals/corr/EvtimovEFKLPRS17} successfully showed how the generated road signs can be physically realizable and fooled the victim model in a controlled environment. For example,~\citet{DBLP:journals/corr/EvtimovEFKLPRS17} demonstrated a sticker attack on a real stop sign that was misclassified as a ``Speed Limit 45'' sign in a stationary lab setting (see Fig.~\ref{fig:printed_stop_sign}). They also conducted drive-by tests where a camera was placed on top of a moving vehicle to confirm that these examples are adversarial even when viewed from different viewpoints. However, one major limitation of RP$_2$ remains: collecting real pictures of an object in the real world under various conditions can be expensive to perform and may not be scalable.

Several works have shown that these adversarial road signs are currently not dangerous to object detection models~\citep{DBLP:journals/corr/LuSFF17,DBLP:journals/corr/abs-1710-03337} which are more commonly deployed in autonomous vehicles than classification models. However, this does not warrant complacency. Object detection models are also vulnerable to attack: see the sections on DAG~\citep{DBLP:journals/corr/XieWZZXY17} (Section~\ref{DAG}) and attacks on Faster R-CNN~\citep{NIPS2015_5638} using EOT~\citep{pmlr-v80-athalye18b} (see Section~\ref{EOT}) over region proposals~\citep{DBLP:journals/corr/abs-1804-05810}. 

\subsection{Adversarial 3D Printed Turtle} \label{TURTLE} 

\begin{figure}[!h]
\centering
\includegraphics[width=0.6\textwidth]{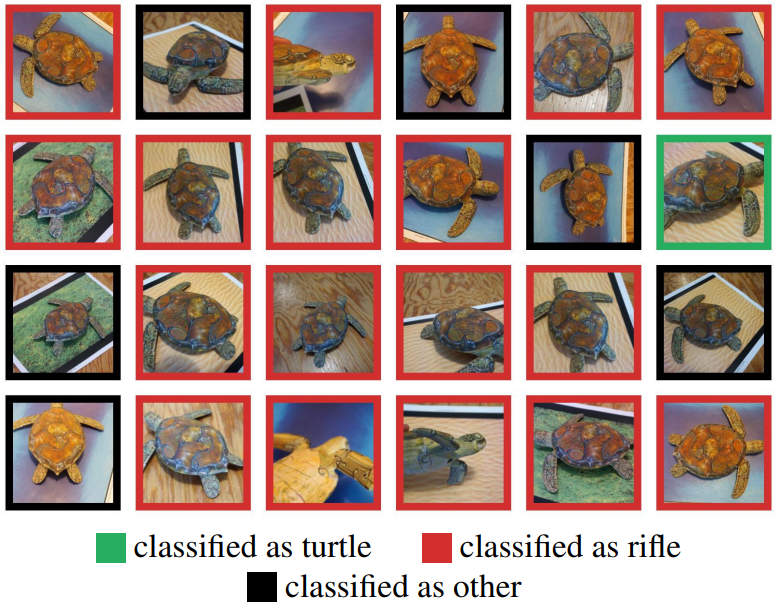}
\caption{The 3D printed adversarial turtle is misclassified as a rifle in most of the cases even when viewed from different angles (taken from~\citet{pmlr-v80-athalye18b}).}
\label{fig:adversarial_turtle}
\end{figure}

\citet{pmlr-v80-athalye18b} demonstrated the effectiveness of EOT (see Section~\ref{EOT}) in generating adversarial examples not only in the virtual image gomain, but also in generating 3D adversarial objects that remain adversarial in the physical world, where the 3D adversarial object is robust to rotation, translation, lighting change, and viewpoint change to reasonable extent. In practice,~\citet{pmlr-v80-athalye18b} modeled the problem using a texture that can be mapped to a 3D shape, and by considering the 3D rendering process as the transformation function. The texture is then updated through an optimization process in order to find an adversarial texture. They showed how they were able to 3D print an adversarial turtle that is misclassified as a rifle, as shown in Fig.~\ref{fig:adversarial_turtle}.

To generate more robust adversaries, instead of only considering similar transformations used in the 2D case (e.g., translation, rotation, Gaussian noise, etc.),~\citet{pmlr-v80-athalye18b} consider more transformations that better reflect the physical world. These include additive/multiplicative lighting and additive/multiplicative per-channel color inaccuracies. They pointed out that modelling color inaccuracies is necessary to account for printing errors.

This perhaps was the first work that showed how adversarial textures can be projected onto some complex 3D shapes, such as the 3D printed turtle, and remain adversarial when viewed from different angles in the real world. We believe this form of attack as one of the most realistic possible attack scenarios in the real world that can dangerously fool intelligent embodied agents such as autonomous cars or mobile robots. Thus, further development of both attack methods that can generate 3D adversarial examples of any shapes in the real world, as well as defense methods that can defend against these adversaries are important research directions.



\subsection{Adversarial Patch} \label{PATCH} 

While many attacks on image classifier focus on modifying an image such that the adversarial image look similar to its original instance,~\citet{DBLP:journals/corr/abs-1712-09665} demonstrated that they were able to generate an adversarial patch that is input-agnostic (i.e., universal, just like UAP~\citep{DBLP:journals/corr/Moosavi-Dezfooli16} (see Section~\ref{UAP})), visually visible, but arguably inconspicuous. The adversarial patch does not look like a real object, but will cause a classifier to misclassify an input whenever the patch is present in the image.

\begin{figure}[h]
\centering
\includegraphics[width=0.6\textwidth]{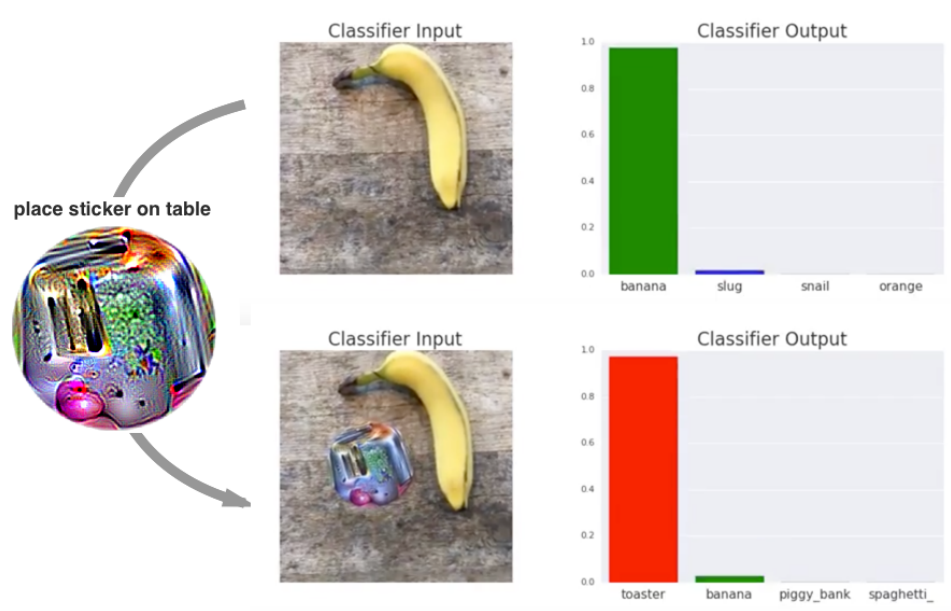}
\caption{The image of a banana is misclassified as a toaster when the adversarial patch is present (taken from~\citet{DBLP:journals/corr/abs-1712-09665}).}
\label{fig:adversarial_patch}
\end{figure}

\citet{DBLP:journals/corr/abs-1712-09665} generated the adversarial patch using a variant of EOT~\citep{pmlr-v80-athalye18b} (see Section~\ref{EOT}), where the objective is to find the patch $p$:

\begin{equation} \label{equations:eot_patch}
\arg\max_p \mathbb{E}_{x \sim X,t \sim T, l \sim L} \Big[ \mathcal{L}(A(p,x,t,l),y) \Big].
\end{equation}

\noindent Here, $A(p,x,t,l)$ denotes a patch application operator that transforms patch $p$ at location $l$ on image $x$ with transformation $t$. In other words, one may see the attack process as the combination of EOT with the method used to generate adversarial eyeglass~\citet{Sharif:2016:ACR:2976749.2978392}. Concretely, the patch application operator can be defined as:

\begin{equation} \label{equations:patch_application_op}
A(p,x,t,l) = t(m(l) \cdot p) + [t(1-m(l)) \cdot x],
\end{equation}

\noindent where $m(l)$ is a binary mask centered at location $l$.

The adversarial patch was shown to be robust in fooling various ImageNet classifiers in whitebox and blackbox settings. Furthermore,~\citet{DBLP:journals/corr/abs-1712-09665} demonstrated that the adversarial patch can be printed and still fool the classifier in the real world. Figure~\ref{fig:adversarial_patch} illustrates how the adversarial patch fools the target model into thinking that the image contains a toaster as opposed to a banana.

\subsection{ShapeShifter} \label{SHAPESHIFTER} 

\begin{figure}[h]
\centering
\includegraphics[width=0.2\textwidth]{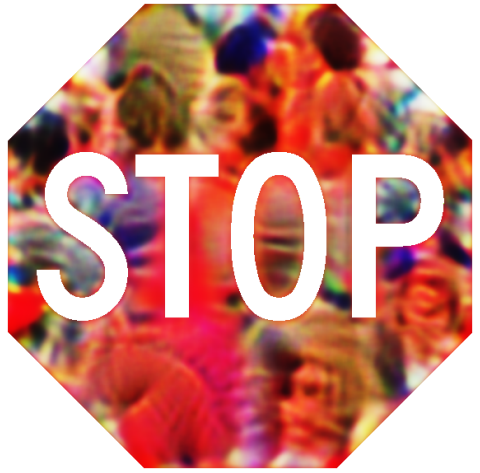}
\caption{An example of a stop sign misdetected as a person with high confidence (taken from~\citet{DBLP:journals/corr/abs-1804-05810}).}
\label{fig:shapeshifter}
\end{figure}

\citet{DBLP:journals/corr/abs-1804-05810} extended EOT~\citep{pmlr-v80-athalye18b} (see Section~\ref{EOT}) and C\&W attacks~\citep{DBLP:journals/corr/CarliniW16a} (see Section~\ref{CW}) to propose ShapeShifter: a method to create physical adversarial objects that fool Faster R-CNN~\citep{NIPS2015_5638}. For example, the perturbed stop sign shown in Fig.~\ref{fig:shapeshifter} is able to fool a Faster R-CNN model that was trained on the MS-COCO dataset~\citep{10.1007/978-3-319-10602-1_48} to mistakenly detect it as a person with high confidence.

Before discussing how the attack works, it is important to realize that Faster R-CNN is made of several main components: (1) region proposal network (RPN)~\citep{DBLP:journals/corr/abs-1804-05810} to produce numerous hypotheses where the objects might be, (2) a classification head to classify the region proposals, and (3) a regression head to refine the bounding box coordinates that represent where the objects are in the image. Since object detection models usually remove some of the region proposals using non-maximum suppression which is a non-differentiable operation,~\citet{DBLP:journals/corr/abs-1804-05810} only consider the region proposals that survive the pruning after running a single forward pass through the RPN. To further simplify the problem, they also only considered the top-$N$ detected objects based on the level of model's confidence.

ShapeShifter aims to attack the classification head for each of the region proposals. Given a classifier $f$, an input image $x$, a set of transformations $T$, and a target class $y'$, the optimization objective is

\begin{equation} \label{equations:shapeshifter}
\arg\min_{x'} \mathbb{E}_{t \sim T} \Big[ \frac{1}{N} \sum_{i=1}^N \mathcal{L}(f(r_i),y') \Big] + c \cdot ||\textrm{tanh}(x') - x||_2^2,
\end{equation}

\noindent where $r_i$ is the $i$-th masked region proposal. Since~\citet{DBLP:journals/corr/abs-1804-05810} wanted to constraint the location of the perturbations, a masking operator was used. It is not clear how the masking operator was implemented, but it may be similar to the masking operator used to generate adversarial eyeglass~\citep{Sharif:2016:ACR:2976749.2978392} (see Section~\ref{EYEGLASS}), except that it has to account for all the region proposals.

\citet{DBLP:journals/corr/abs-1804-05810} showed that the generated adversaries can reasonably fool the Faster R-CNN model in the real world when viewed from different distances and view angles. The adversaries were tested both in a controlled indoor environment and by conducting drive-by tests. They also identified some limitations such as the weakness of the perturbations when viewed from a high viewing angle and lack of transferability to other detection models. However, this work emphasizes once again how adversarial examples can be physically realizable to fool object detection models that are more complex than classification models.

\subsection{Non-physical Adversaries in the Real World} \label{NON_PHYSICAL} 


In some cases, adversarial attacks do not need to be physically realizable to be dangerous. For example, camera model identification is the task of identifying the specific camera that was used to take an image of interest, which can be used to detect copyright infringement and image falsification. Several works have considered deep learning as a means to perform camera model identification~\citep{7786852,tuama:hal-01388975,Bondi:2017:2470-1173:67}.~\citet{8014964} showed that adversarial attack techniques such as the FGSM~\citep{43405} (see Section~\ref{FGSM}) and JSMA~\citep{DBLP:journals/corr/PapernotMJFCS15} (see Section~\ref{JSMA}) are sufficient to fool the models that were trained to perform camera model identification. This type of attack does not only exist in the vision domain, but also in the digital audio domain~\citep{saadatpanah2019adversarial}, where the attack successfully fooled real world music copyright detectors.


Another area where adversarial examples do not have to be physically realizable is in the field of medical imaging. Deep learning systems have shown promising performance when it comes to medical image analysis~\citep{Shen2017}. 
Deep learning has been used for diagnosis of breast cancer~\citep{Bejnordi2017DiagnosticAO}, pneumonia~\citep{Rajpurkar2017CheXNetRP}, skin cancer~\citep{Esteva2017,doi:10.1093/annonc/mdy166}, and has demonstrated super-human performance in some of the cases.~\citet{DBLP:journals/corr/abs-1804-05296} showed that existing adversarial attack methods can be applied on medical images, and discussed various real world scenarios where adversarial examples might negatively impact the healthcare economy when machine learning based systems are used. A bad dermatologist could commit fraud by inserting adversarial perturbations on dermoscopy images in order to create more profit by suggesting to perform unnecessary procedures on patients, which will negatively affect both the patients and insurance companies. A bad pharmaceutical company can apply adversarial perturbations to alter the images used for new drug trials in order to get positive results. A bad insurer can use adversarial examples to alter the healthcare policy by manipulating the rate of surgeries required, which may be based on recommendations from machine learning based systems. All of these are possible scenarios on how adversarial examples can negatively impact the healthcare industry that are important to be addressed.



\pagebreak

\section{Defenses against Adversarial Attacks} \label{DEFENSE} 

\begin{figure}[H]
\centering
\includegraphics[width=1.15\textwidth, angle=90]{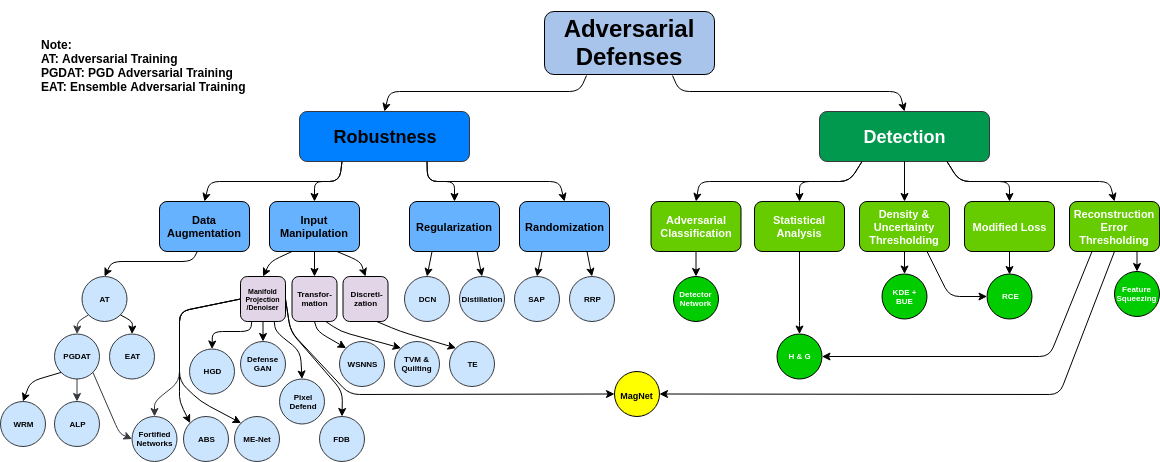}
\caption{Ontology of adversarial defenses covered in this paper.}
\label{fig:defense_ontology}
\end{figure}

This section covers various defense strategies against adversarial attacks. Figure~\ref{fig:defense_ontology} and Table~\ref{table:defense_taxonomy} illustrates the ontology and taxonomy of adversarial defense techniques discussed in this paper, respectively.


\begin{sidewaystable}[!ph]
\centering
\caption{Taxonomy of adversarial defenses covered in this paper.}
\begin{tabular*}{ 1.0 \textwidth}{lll}
\hline
\textbf{Adversarial Defense(s)} & \textbf{Goal} & \textbf{Remarks} \\
\hline
Adversarial Training~\citep{43405} & R & Training on adversarial examples \\
\hline
Ensemble Adversarial Training~\citep{tramèr2018ensemble} & R & More robust to blackbox attacks compared to standard adversarial training \\
\hline
DCN~\citep{DBLP:journals/corr/GuR14} & R & Early defense against adversarial attacks with gradient regularization \\
\hline
Defensive Distillation~\citep{journals/corr/PapernotMWJS15} & R & Circumventable by C\&W~\citep{DBLP:journals/corr/CarliniW16a}, SBA~\citep{DBLP:journals/corr/PapernotMGJCS16}, and variant of JSMA~\citep{DBLP:journals/corr/CarliniW16} \\
\hline
MagNet~\citep{Meng:2017:MTD:3133956.3134057} & R, D & Combination of R \& D, circumventable by C\&W~\citep{DBLP:journals/corr/CarliniW16a} and SBA~\citep{DBLP:journals/corr/PapernotMGJCS16} \\
\hline
Random Resizing \& Padding~\citep{xie2018mitigating} & R & Circumventable by EOT variant (Expectation Over Randomness)~\citep{obfuscated-gradients} \\
\hline
SAP~\citep{s.2018stochastic} & R & Circumventable by EOT variant (Expectation Over Randomness)~\citep{obfuscated-gradients} \\
\hline
TVM \& Quilting~\citep{guo2018countering} & R & Circumventable by combination of BPDA~\citep{obfuscated-gradients} and EOT~\citep{pmlr-v80-athalye18b} \\
\hline
TE~\citep{buckman2018thermometer} & R & Circumventable by BPDA~\citep{obfuscated-gradients} \\
\hline
PixelDefend~\citep{song2018pixeldefend} & R, D & Circumventable by BPDA~\citep{obfuscated-gradients} and SPSA~\citep{pmlr-v80-uesato18a} \\
\hline
Defense-GAN~\citep{samangouei2018defensegan} & R & Circumventable by BPDA~\citep{obfuscated-gradients} \\
\hline
PGD Adversarial Training~\citep{madry2018towards} & R & Training only on PGD adversaries \\
\hline
WRM~\citep{sinha2018certifiable} & R & Adversarial training with robustness certificate \\
\hline
HGD~\citep{DBLP:journals/corr/abs-1712-02976} & R & Circumventable by SPSA~\citep{pmlr-v80-uesato18a} \\
\hline
ALP~\citep{DBLP:journals/corr/abs-1803-06373} & R & Circumventable by PGD~\citep{madry2018towards} with many iterations~\citep{DBLP:journals/corr/abs-1807-10272} \\
\hline
FN~\citep{lamb+al-2018-fortified} & R & Denoising on hidden representations using autoencoders \\
\hline
FDB~\citep{DBLP:journals/corr/abs-1812-03411} & R & Denoising on hidden representations using differentiable denoising operation \\
\hline
ABS~\citep{Schott2018a} & R & Model distribution of the inputs for each class using VAE~\citep{DBLP:journals/corr/KingmaW13} \\
\hline
WSNNS~\citep{DBLP:journals/corr/abs-1903-01612} & R & Replace input with its nearest neighbor from a large database of images \\
\hline
ME-Net~\citep{pmlr-v97-yang19e} & R & Defense using matrix estimation algorithms \\
\hline
H\&G's Methods~\citep{DBLP:journals/corr/HendrycksG16b, DBLP:journals/corr/HendrycksG16c} & D & Circumventable by modified C\&W~\citep{DBLP:journals/corr/CarliniW16a} \\
\hline
Detector Networks~\citep{metzen2017detecting,DBLP:journals/corr/GongWK17,DBLP:journals/corr/GrosseMP0M17,DBLP:journals/corr/HosseiniCKZP17} & D & Circumventable by C\&W~\citep{DBLP:journals/corr/CarliniW16a} and SBA~\citep{DBLP:journals/corr/PapernotMGJCS16} \\
\hline
KDE \& BUE~\citep{DBLP:journals/corr/FeinmanCSG17} & D & Circumventable by modified C\&W~\citep{DBLP:journals/corr/CarliniW16a} \\
\hline
Feature Squeezing~\citep{DBLP:journals/corr/XuEQ17} & D & Detection by comparing the predictions between preprocessed and original inputs \\
\hline
RCE~\citep{DBLP:journals/corr/PangDZ17} & D & Defense using reverse crossentropy loss \\
\hline
\end{tabular*}
\footnotetext{R: Robustness}
\footnotetext{D: Detection}
\label{table:defense_taxonomy}
\end{sidewaystable}

\subsection{Increasing Robustness of Machine Learning Models} \label{ROBUSTNESS} 

A number of approaches have been proposed which seek to make models more robust to adversarial attacks. These techniques do not rely upon the detection of adversarial examples (see Section ~\ref{DETECTION}), instead formulating models which perform equally well with adversarial and normal inputs. At their heart, many of these methods seek to make models less sensitive to irrelevant variation in the input, effectively regularizing models to reduce the attack surface and limit responsiveness to off-manifold perturbations. However, a recurring theme in this section is that many approaches that were thought to be robust have subsequently been shown to rely upon \emph{gradient masking}, and thus to be vulnerable to the suite of attacks which bypass this form of defense. We will highlight where defenses should be re-evaluated in light of attack methods that are specifically designed to circumvent gradient masking, such as BPDA~\citep{obfuscated-gradients} (see Section~\ref{BPDA}) and SPSA~\citep{pmlr-v80-uesato18a} (see Section~\ref{SPSA}).

\subsubsection{Adversarial Training} \label{ADV_TRAINING} 


Adversarial training~\citep{43405} is a defense method in which a model is retrained by injecting adversarial examples into training set at every training iteration. In practice, one can generate adversarial examples using a method of choice such as the FGSM~\citep{43405} (see Section~\ref{FGSM}) or BIM~\citep{DBLP:journals/corr/KurakinGB16,45816} (see Section~\ref{BIM}). Formally, the new training objective becomes

\begin{equation} \label{equations:adversarial_training}
\alpha \mathcal{L}(x,y) + (1-\alpha) \mathcal{L}(x', y),
\end{equation}

\noindent where $\mathcal{L}(x,y)$ denotes the classification loss function (e.g., cross-entropy), and $x'$ denotes the adversarial counterpart of clean example $x$. Here, $\alpha$ is a constant that controls the weighting of the loss terms between normal and adversarial inputs where $\alpha$ is usually set to be 0.5. In this paper, we refer to adversarial training based on the type of adversarial examples the defended model is trained on. For example, if the defended model is trained on FGSM adversaries, we refer this as FGSM adversarial training, or in short, FGSM-AT.

Although adversarial training is relatively simple to implement and performs fairly well, adversarially trained models may still be vulnerable to other types of adversarial examples the model was not trained on. For example, adversarial training on adversarial examples generated using single-step gradient-based attacks (e.g., FGSM) shows robustness against the same type of attacks even when evaluated in large scale~\cite{45816}, but training against single-step methods produces models that are less robust against iterative gradient-based attacks (e.g., BIM).

Furthermore, adversarial training on single-step adversaries can suffer from \emph{label leakage}~\cite{45816}.~\citet{45816} observed that models trained on adversarial examples generated by a specific single step attack method have higher accuracy on adversarial examples generated by the same method compared to clean examples. This suggests that the adversaries generated by single step methods are too simple and can be easily overfitted by the model. This problem perhaps reflects the limited options for generating adversarial examples at the time of publication. Future research might consider adversarial training on various types of adversaries rather than exclusively on those generated by single-step methods. Furthermore, it would be useful to evaluate robustness against attack techniques that were not used during the training process, assessing the generalization of the defense afforded by this method. Along those lines, it might also be useful to treat the problem as a meta-learning task: given a set of attacks, learn parameters that facilitate swift training of defense to new attacks with as few examples as possible.

\citet{tramèr2018ensemble} found that the robustness provided by adversarial training on single-step adversaries was mainly caused by gradient masking (see Section~\ref{R+FGSM}). As a result, the model can be circumvented by other types of attacks like the R+FGSM~\citep{tramèr2018ensemble} (see Section~\ref{R+FGSM}) or the SBA~\citep{DBLP:journals/corr/PapernotMGJCS16} (see Section~\ref{SUBSTITUTE}). One proposed solution is to include examples that have been generated to fool \emph{other} models. This is further discussed in Section~\ref{ENSEMBLE_ADV_TRAINING}. An alternative approach is to train on adversarial examples generated by the Projected Gradient Descent (PGD) method~\citep{madry2018towards} (see Section~\ref{PGD_ADV}).

\subsubsection{Deep Contractive Network} \label{DCN} 

\citet{DBLP:journals/corr/GuR14} proposed a defense for deep neural networks using a regularization term that is adopted from the Contractive Autoencoders~\citep{Rifai:2011:CAE:3104482.3104587} called the contractive penalty, hence the name Deep Contractive Network (DCN). The contractive penalty encourages the model to be invariant to small changes in directions which are irrelevant in the input space. Given $N$ datapoints, DCN with $H$ number of layers is trained to minimize

\begin{equation} \label{equations:dcn}
\sum_{i=1}^N \Big( \mathcal{L}(x_i,y_i) + \sum_{j=1}^{H+1}\lambda_j||\frac{\partial h_j^i}{\partial h_{j-1}^i}||_2 \Big),
\end{equation}

\noindent where $\lambda$ is a constant that control the importance of the layer-wise penalty term and $\mathcal{L}(x,y)$ is the classification loss (e.g., cross-entropy). Here, $h_j^i$ represents the activation at the hidden layer $j$ given the $i$-th input $x_i$. Intuitively, this method establishes a tension: between the classification loss, which requires the network to capture useful information in the input, and the contractive loss, which encourages it to throw away redundant information. As a result, the model gains some invariance to the small pixel-wise perturbations typically introduced by adversarial attacks.


\citet{DBLP:journals/corr/GuR14} evaluated this strategy against the L-BFGS attack and showed that while the attack still successfully found adversarial examples that fool DCN, the average adversarial perturbations needed is larger compared to when attacking standard model. This perhaps is one of the earliest defense methods proposed since the findings of adversarial examples for neural networks in~\citet{42503}. 

\subsubsection{Defensive Distillation} \label{DISTIL} 

The term \emph{distillation} was previously introduced by~\citet{44873} as a method to compress knowledge from an ensemble of deep neural networks into a single neural network. Similarly, defensive distillation~\citep{journals/corr/PapernotMWJS15} is a defense method that works by training two networks, where one network (the ``student'') is trained to approximate the other (the ``teacher''). 

\begin{figure}[h]
\centering
\includegraphics[width=0.8\textwidth]{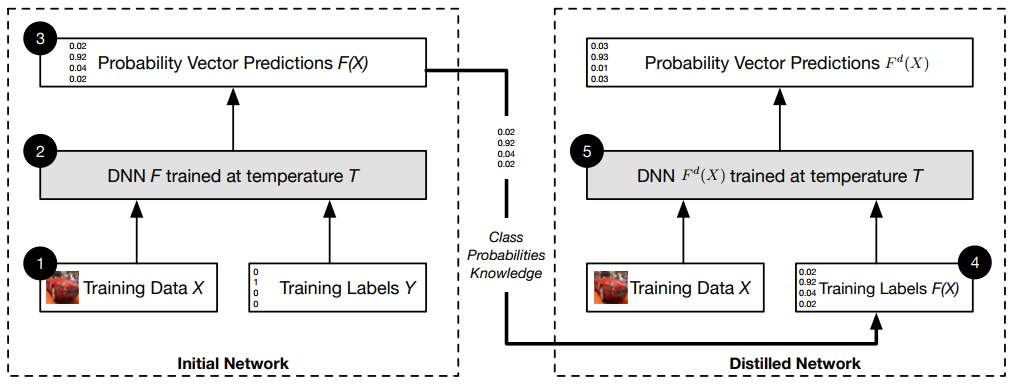}
\caption{Illustration of defensive distillation (taken from~\citet{journals/corr/PapernotMWJS15}).}
\label{fig:defensive_distillation}
\end{figure}

Defensive distillation works as follows. First, the ``teacher'' network is trained with a modified softmax function at the output layer, including a temperature constant $T$, where $T\geqslant1$, that is

\begin{equation} \label{equations:softmax_temp}
softmax(Z(x)) = \frac{e^{Z(x)_{(i)}/T}}{\sum_{c=0}^{C} e^{Z(x)_{(c)}/T}},
\end{equation}

\noindent where $Z(x)$ is the logit vector, and $C$ is the number of classes. Higher temperatures lead to a flatter softmax, thus introducing more noise into the decision process. After training the first network, the second network is trained using the predictions \emph{from the first network} as the training labels for the training data. The hope is that this prevents the distilled network from overfitting. Figure~\ref{fig:defensive_distillation} illustrates the mechanics of defensive distillation. During test time, the temperature constant of the distilled model is set back to 1.

\citet{journals/corr/PapernotMWJS15} showed how increasing the temperature of the softmax function causes the probability vector to converge to $\frac{1}{C}$ as $T \rightarrow \infty$. In other words, the predictions from both the teacher and student models are trained to be smooth. They also showed that the Jacobian of the model with respect to input features is inversely proportional to the temperature: the higher the temperature, the lower the dependence upon input features. They argued that when the temperature is brought back to 1 during test time, it will only cause the probability vector to be more discrete but will not change the Jacobian of the model, arriving at a model with reduced sensitivity to small changes in the input (a similar logic to the Deep Contractive Network in Section~\ref{DCN}).

Although defensive distillation was shown to be robust against adversarial examples generated using the JSMA-F~\citep{DBLP:journals/corr/PapernotMJFCS15} (see Section~\ref{JSMA}), this defense was found to be vulnerable to a variant of JSMA-Z~\citep{DBLP:journals/corr/CarliniW16}, C\&W attacks~\citep{DBLP:journals/corr/CarliniW16a} (see Section~\ref{CW}), as well as SBA~\citep{DBLP:journals/corr/PapernotMGJCS16} (see Section~\ref{SUBSTITUTE}).~\citet{DBLP:journals/corr/CarliniW16a,DBLP:journals/corr/CarliniW16} evaluated this defense and pointed out how defensively distilled model produces logits with higher values compared to standard models. They observed that if the values of the correct logit are large, the softmax probability of the correct class will be closer to $1.0$ while those of the non-maximum elements become smaller. They also found that the values of non-maximum element are even smaller than the precision of 32-bit floating values, and thus are rounded to 0. In this case, the gradients of these units with respect to the input will be 0, which may cause attacks like JSMA-F to fail in some cases. This is an interesting twist, given that the intention of the temperature parameter was originally to produce a smoother output distribution.~\citet{DBLP:journals/corr/CarliniW16} showed that this defense can be circumvented by JSMA-Z variant where the logits are divided by the temperature constant $T$ before being used to calculate the gradients.

\subsubsection{MagNet} \label{MAGNET} 

MagNet~\citep{Meng:2017:MTD:3133956.3134057} is a multi-pronged approach to robustness, which requires a pair of autoencoders: a detector ($D$) and a reformer networks ($R$). Both are trained to reconstruct clean examples. However, the detector $D$ is used to detect if an input is adversarial or not, while the reformer $R$ is used to remove adversarial perturbations from adversarial inputs, moving examples back towards the manifold of clean data.

Since $D$ has been trained to reconstruct clean examples, the detector network outputs images that are close to the distribution of the clean samples, even when the inputs to the detector network are adversarial. If the input was already a clean example, then the output of the detector network looks very similar; if the input was in fact adversarial, then the reconstruction loss between input and output will be high. One can then perform adversarial detection by setting a threshold value on the input reconstruction error.

Since naively measuring the reconstruction error only works if adversarial perturbations are large,~\cite{Meng:2017:MTD:3133956.3134057} leverage the target classifier network $f$ in their detection algorithm. They do this by measuring the Jensen-Shannon Divergence ($D_{JS}$)~\citep{1365067} between $f$'s output when fed the original image and $f$'s output when fed the detector-preprocessed image. The $D_{JS}$ can be understood as symmetric version of the Kullback-Leibler divergence~\citep{kullback1951}. Formally, the $D_{JS}$ is defined as

\begin{equation} \label{equations:jsd}
D_{JS} \bigg( f(x)||f(D(x)) \bigg) = \frac{1}{2} D_{KL}\bigg(f(x)||M\bigg) + \frac{1}{2} D_{KL}\bigg(f(D(x))||M\bigg),
\end{equation}

\noindent where $M = \frac{f(x)+f(D(x))}{2}$, and $D_{KL}$ is the Kullback-Leibler divergence, $D_{KL}(P||Q) = \sum_i P(i) \log \frac{P(i)}{Q(i)}$.


\citet{Meng:2017:MTD:3133956.3134057} argued that although the difference between $x'$ and $D(x')$ may be small (e.g., when $x$ and $x'$ are similar), the difference between $f(x')$ and $f(D(x'))$ will be large due to the misclassification by the target network. Conversely, the difference between $f(x)$ and $f(D(x))$ stays small. Thus, the measure of the $D_{JS}$ gives a signal of whether an input is adversarial or not. To improve numerical stability,~\citet{Meng:2017:MTD:3133956.3134057} set the temperature constant of the softmax function to be larger than one, just as in defensive distillation~\citep{journals/corr/PapernotMWJS15} (see Section~\ref{DISTIL}).

\citet{Meng:2017:MTD:3133956.3134057} realized that the whole MagNet pipeline is differentiable, making it vulnerable in whitebox settings. They thus suggested to use multiple detector and reformer networks, and then to randomly choose two of the candidates as the detector and reformer networks during test time. A regularization term that penalizes similarity between candidates was also proposed. Formally, the training objective is defined as

\begin{equation} \label{equations:magnet_loss}
\sum_{i=1}^n ||x - C_i(x)||_2 - \lambda \sum_{i=1}^n ||C_i(x) - \frac{1}{n} \sum_{j=1}^n C_j(x)||_2,
\end{equation}

\noindent where $n$ is the number of candidates and $C_i$ denotes the $i$-th candidate. In Equation~\ref{equations:magnet_loss}, the first term is the reconstruction loss while the second term is the penalty term which encourages diversity of candidates (as the defense will not be meaningful if the candidates are similar).

\citet{Meng:2017:MTD:3133956.3134057} showed that MagNet is robust against various attacks such as the FGSM~\citep{43405} (see Section~\ref{FGSM}), BIM~\citep{DBLP:journals/corr/KurakinGB16,45816} (see Section~\ref{BIM}), DeepFool~\citep{DBLP:journals/corr/Moosavi-Dezfooli15} (see Section~\ref{DEEPFOOL}), and C\&W attacks~\citep{DBLP:journals/corr/CarliniW16a} (see Section~\ref{CW}). Nevertheless,~\citet{DBLP:journals/corr/abs-1711-08478} found that MagNet is still vulnerable to the SBA~\citep{DBLP:journals/corr/PapernotMGJCS16} (see Section~\ref{SUBSTITUTE}) where the whitebox method used to attack the substitute model was the C\&W attack~\citep{DBLP:journals/corr/CarliniW16a} (see Section~\ref{CW}).

\subsubsection{PGD Adversarial Training} \label{PGD_ADV} 

Building on previous attempts~\citep{DBLP:journals/corr/HuangXSS15,SHAHAM2018195} which defined adversarial robustness from a robust optimization~\citep{BEN:09} perspective,~\citet{madry2018towards} proposed a variant of adversarial training~\citep{43405} (see Section~\ref{ADV_TRAINING}) on worst-case adversaries (see below for an explanation of how ``worst-case'' is defined). In the framework of robust of optimization, the training objective is defined as:

\begin{equation} \label{equations:adversarial_erm}
\arg \min_{\theta} \mathbb{E} \Big[ \max_{||x'-x||_{\infty}} \mathcal{L}(x',y) \Big].
\end{equation}

\noindent In this formulation, the inner maximization aims to find adversarial examples that maximize a loss function, while the outer minimization refers to the training process which aims to find a set of parameters $\theta$ that will minimize the worst-case loss. This is similar to the minimax loss used in game-playing: we are trying to minimize the impact of our adversaries worst moves. Note how this is different than standard adversarial training in the sense that the standard method trains a model on \emph{both} clean and adversarial examples, while in this framework the model is trained only on the adversarial examples. In other words, PGD adversarial training is equivalent to setting $\alpha$ to 0 in Equation~\ref{equations:adversarial_training} in Section~\ref{ADV_TRAINING}.

\citet{madry2018towards} argued that adversarial examples generated using randomly initialized BIM~\citep{DBLP:journals/corr/KurakinGB16,45816} (see Section~\ref{BIM}), which they refer to as Projected Gradient Descent (PGD), are the worst case adversarial examples. They showed that even though PGD was restarted from different starting points (e.g., by adding small random perturbations to the initial input) for several runs, the loss value of a model on these adversaries always plateaus at a similar level. They therefore argued that this suggests the existence of numerous local maxima with similar values, which are similar to the global maximum. This perhaps echoes the argument in~\citep{pmlr-v38-choromanska15} in which the authors argued that the surface loss of neural networks has \emph{many local minima with similar values}. Based on these findings,~\citet{madry2018towards} suggested that robustness against PGD adversaries approximately solves the min-max formulation in Equation~\ref{equations:adversarial_erm}, thus should provide robustness against other first-order adversaries. Concretely, PGD adversarial training was performed by training the network \emph{only on adversarial examples} generated by PGD with random starting points, which one may see as an iterative steps variant of R+FGSM~\citep{tramèr2018ensemble} (see Section~\ref{R+FGSM}).

They found that larger capacity models (e.g., models with more tunable parameters) receive more benefit from this defense approach compared to smaller capacity models. They hypothesized that a model needs to be capable of producing more complicated decision boundaries in order to be robust against adversarial examples, and thus may require more capacity. This argument is supported empirically by the observation that maximum loss decreases as the capacity of the model increases. They also found that increases in capacity helps to reduce effectiveness of transferable adversarial examples. The experimental results show how this defense is \emph{consistently} robust to various types of attacks in both whitebox and blackbox settings, although the defended model may not achieve state-of-the-art performance. This defense was the only defense evaluated by~\citet{obfuscated-gradients} that can live up to its claims.~\citet{pmlr-v80-uesato18a} also showed how this defense does not rely on gradient masking by evaluating this defense against SPSA (see Section~\ref{SPSA}), which is a gradient-free adversarial attack.

As a follow-up work,~\citet{tsipras2018robustness} showed several unexpected benefits to training models with this method. First, they showed how models trained using PGD-AT tend to only utilize features that are strongly correlated with the correct label, effectively reducing the attack surface (thus implicitly achieving the same goal as the Deep Contractive Network, but better, see Section~\ref{DCN}). They argued that standard classifiers utilize too many features that are weakly correlated with the correct label in order to achieve high accuracy, thus sacrificing adversarial robustness. Lastly, they showed how the gradients of a robust model is visually interpretable to a human with no additional steps required, precisely because the robust solution relies upon a small number of important features rather than a large number of weakly relevant ones. Note that other gradient-based explainability methods such as Gradient-weighted Class Activation Mapping (Grad-CAM)~\citep{8237336} needs to perform additional steps to achieve visually interpretable saliency maps. Similar findings were also reported by~\citet{engstrom2019learning}.

Although PGD-AT seems to be promising, the benefits do not come for free. For example, PGD-AT was shown to compromise the model's accuracy on clean examples dramatically~\citep{tsipras2018robustness}. Furthermore, the computational cost of this method is higher than standard adversarial training, since the cost scales with the number of PGD iterations. More recent works also showed how PGD adversarial training is mainly only robust against certain types of adversary used during the training, but not to the others~\citep{Schott2018a,tramer2019adversarial,kang2019transfer}.



\subsubsection{Ensemble Adversarial Training} \label{ENSEMBLE_ADV_TRAINING} 

Ensemble adversarial training~\citep{tramèr2018ensemble} is a variant of adversarial training~\citep{43405} (see Section~\ref{ADV_TRAINING}) in which a model is retrained on adversarial examples generated to attack \emph{various other pre-trained models}. This decoupling of target model and adversarial training examples overcomes the overfitting observed with the vanilla version of adversarial training.~\citet{tramèr2018ensemble} argued that ensemble adversarial training is a good approximation of the inner maximization shown in Equation~\ref{equations:adversarial_erm} due to the transferability of adversarial examples between different models. Moreover, since the adversarial examples generation process is independent from the model that is being trained, they hypothesized ensemble adversarial training to be more robust to future blackbox attacks compared to the standard adversarial training.

In practice,~\citet{tramèr2018ensemble} alternate the source of adversarial examples between the currently trained and other pre-trained models, where the pre-trained models are chosen randomly per batch. This is done in order to ensure diversity of adversarial examples during training.~\citet{tramèr2018ensemble} noted that the gradients of the pre-trained models can be \emph{pre-computed} for the whole training dataset. Since we are not updating the weights of the pre-trained models, the computation needed per training batch in ensemble adversarial training is thus lower compared to standard adversarial training.

\citet{tramèr2018ensemble} demonstrated that this method outperforms regular adversarial training on ILLCM adversaries (see Section~\ref{ILLCM}) in blackbox settings. In their experiments, the models were evaluated against SBA~\citep{DBLP:journals/corr/PapernotMGJCS16} (see Section~\ref{SUBSTITUTE}) that used FGSM~\citep{43405} (see Section~\ref{FGSM}) to attack the substitute model. However, they also found that ensemble adversarial training is slightly less accurate and slower to converge compared to the standard method. They also demonstrated the scalability of ensemble adversarial training by evaluating their defense on ImageNet dataset.

As discussed in Section~\ref{R+FGSM}, performing standard adversarial training causes the defended model to be more robust against 2-step BIM~\citep{DBLP:journals/corr/KurakinGB16,45816} (see Section~\ref{BIM}) than R+FGSM~\citep{tramèr2018ensemble} (see Section~\ref{R+FGSM}), strengthening the suspicion of gradient masking.~\citet{tramèr2018ensemble} showed that this ratio is decreased in the ensemble case. Nevertheless, ensemble adversarial training is still not immune to adversarial examples as it is still circumventable by R+FGSM.

\subsubsection{Random Resizing and Padding as a Defense} \label{RESIZE-PAD} 

\citet{xie2018mitigating} introduced a defense based on Random Resizing and Padding (RRP). The goal of this approach is to remove perturbations through input transformation and introduce randomness during inference so that the gradient of the loss with respect to the input is harder to compute. 
Concretely, RRP is implemented as two additional layers placed right after the input layer of a neural network. The random resizing layer resizes the input image to several numbers of resized images, where the differences between dimensions of input images before and after resizing layer are small. The random padding layer randomly pads the resized images with zeros, returning several padded images. One of these padded images is then randomly selected to be classified by the classifier.

\begin{figure}[h]
\centering
\includegraphics[width=0.7\textwidth]{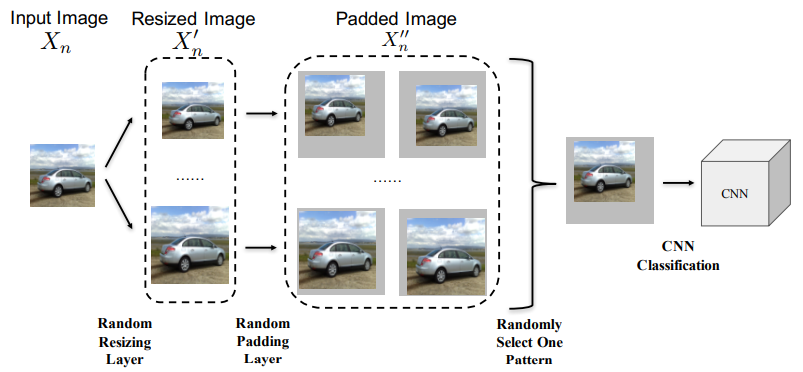}
\caption{Illustration of RRP (taken from~\citet{xie2018mitigating}).}
\label{fig:random_resize_padding}
\end{figure}


They demonstrated that the proposed method requires no fine-tuning of the defended model without sacrificing accuracy on clean examples and can be combined with other defense methods such as adversarial training~\citep{43405} (see Section~\ref{ADV_TRAINING}). This defense method showed good performance against various whitebox attacks such as the FGSM~\citep{43405} (see Section~\ref{FGSM}), BIM~\citep{DBLP:journals/corr/KurakinGB16,45816} (see Section~\ref{BIM}), DeepFool~\citep{DBLP:journals/corr/Moosavi-Dezfooli15} (see Section~\ref{DEEPFOOL}), and C\&W~\citep{DBLP:journals/corr/CarliniW16a} (see Section~\ref{CW}) attacks. Furthermore, this defense was ranked 2nd in the NIPS 2017 adversarial examples defense challenge~\citep{DBLP:journals/corr/abs-1804-00097}.

\citet{xie2018mitigating} argued that the space of random resizing and padding is sufficiently large as to be computationally intractable for an attacker. However, both~\citet{obfuscated-gradients} and~\citet{pmlr-v80-uesato18a} showed that RRP relies on gradient masking~\citep{obfuscated-gradients}, and can be circumvented using EOT~\citep{pmlr-v80-athalye18b} (see Section~\ref{EOT}).

\subsubsection{Stochastic Activation Pruning} \label{SAP} 

\citet{s.2018stochastic} proposed a defense mechanism based on stochastic policy called Stochastic Activation Pruning (SAP) which randomly prunes or drops some activations (i.e., set to 0) during test time. The probability of the nodes $j$ at $i$-th layer $p_j^i$ to be sampled, or survive, is proportional to the magnitude of the activation. This means that the larger an activation is, the smaller the probability that activation will be pruned. The survival probability $p_j^i$ is calculated as

\begin{equation} \label{equations:sap_probability}
p_j^i = \frac{|h_j^i|}{\sum_{k=1}^{K^i}|h_k^i|} ,
\end{equation}


\noindent where $h_j^i$ denotes the activation of the $j$-th node at layer $i$, and $K^i$ denotes the number of activation nodes at the $i$-th layer.

The magnitude of the activations that survive at the $i$-th layer are then scaled up by the inverse of $p_j^i$ given by

\begin{equation} \label{equations:sap_magnitude}
h_j^i \odot \frac{\mathbb{I}(h_j^i)}{1 - (1 - p_j^i)^{r^i}} ,
\end{equation}


\noindent where $\mathbb{I}(h_j^i)$ returns 1 if $h_j^i$ was sampled at least once, and returns 0 otherwise, i.e., the node is pruned. Here, $r^i$ denotes the number of samples to be drawn at $i$-th layer.~\citet{s.2018stochastic} argued that the goal of rescaling the activations is to make the pruned model behaves similarly to the original model when dealing with non-adversarial inputs, while being more robust to adversarial examples since the stochasticity in SAP causes the gradients to be hard to compute by an attacker. The SAP algorithm is described in Algorithm~\ref{alg:sap}.

\begin{algorithm}[h]
   \caption{SAP~\citep{s.2018stochastic}}
   \label{alg:sap}
\begin{algorithmic}
   \STATE {\bfseries Input:} input $x=h^0$, neural network $f$ with $L$ layers, weights of the neural network $W$ at each layer, non-linear activation functions used in each layer $\phi$, number of samples per layer $r^i$, set of activation units not to be pruned $S$
   \STATE {\bfseries Output:} predicted label $\hat{y}(x)$
   \STATE {\bfseries Initialize:} $S \leftarrow \{\}$
   \FOR{each layer $i$}
   \STATE $h^i \leftarrow \phi^i(W^ih^{i-1})$
   \FOR{each activation unit $j$ in layer $i$}
   \STATE $p^i_j \leftarrow \frac{|h_j^i|}{\sum_{k=1}^{K^i}|h_k^i|}$
   \ENDFOR
   \STATE $S \leftarrow \{\}$
   \FOR{number of samples $r^i$}
   \STATE Sample $j \sim h^i$ based on $p^i_j$
   \STATE $S \leftarrow S \cup \{j\}$
   \ENDFOR
   \FOR{each index $j \not\in S$}
   \STATE $h^i_j \leftarrow 0$
   \ENDFOR
   \FOR{each index $j \in S$}
   \STATE $h^i_j \leftarrow \frac{h_j^i}{1 - (1 - p_j^i)^{r^i}}$
   \ENDFOR
   \ENDFOR
   \STATE $\hat{y}(x) \leftarrow \arg\max_c(softmax(h^L))_{(c)}$
   \STATE {\bfseries Return:} $\hat{y}(x)$
\end{algorithmic}
\end{algorithm}


This method is similar to performing dropout~\citep{JMLR:v15:srivastava14a} during test time. The difference is that dropout assigns the probability for the nodes to be pruned uniformly, while SAP assigns them based on the magnitude of the activation and rescales the magnitude of the surviving nodes.~\citet{s.2018stochastic} argued that these differences allow SAP to be added to any pre-trained models without the need of fine-tuning, whilst maintaining accuracy on clean data.

\citet{s.2018stochastic} showed that SAP can defend against FGSM~\citep{43405} (see Section~\ref{FGSM}), where Monte Carlo sampling~\citep{10.2307/2280232} is used to approximate the gradients. SAP was also shown to increase the robustness of Double Deep Q-Learning (DDQN)~\citep{Hasselt:2016:DRL:3016100.3016191} algorithm, which is a reinforcement learning algorithm that uses neural networks to parametrize the action-value function, against adversarial inputs when tested on various Atari games environment~\citep{Bellemare:2015:ALE:2832747.2832830}. Furthermore, they also showed how SAP can be combined with other defense methods such as adversarial training~\citep{43405} (see Section~\ref{ADV_TRAINING}) to further enhance robustness.

Nevertheless, SAP has been shown to rely on gradient masking and vulnerable to an iterative attack which computes the gradients based on the expectation of randomness~\citep{obfuscated-gradients}. Instead of computing $\nabla_x f(x)$ once for each input, the attack computes it based on $k$ predictions and moves in the direction of $\sum_{i=1}^k \nabla_x f(x)$.~\citet{obfuscated-gradients} also suggested performing gradient clipping when attacking SAP to avoid exploding gradients due to the existence of many nodes with zero activations.

\subsubsection{Total Variance Minimization and Quilting} \label{TRANSFORM-DEFENSE} 

\citet{guo2018countering} apply various image transformations to an input before classification. These transformations included image cropping and rescaling, bit-depth reduction~\citep{DBLP:journals/corr/XuEQ17}, JPEG compression~\citep{DBLP:journals/corr/DziugaiteGR16}, Total Variance Minimization (TVM)~\citep{RUDIN1992259}, and image quilting~\citep{Efros:2001:IQT:383259.383296}. They found that TVM and image quilting were the most effective, especially when the model is trained on the transformed images. Both TVM and image quilting introduce randomness, and are non-differentiable operations which makes it harder for an attacker to compute the gradient of the model. These defenses are model-agnostic which means the model does not need to be retrained or fine-tuned, although finetuning was shown to improve the performance of the defended model. These defenses can also be coupled with other defense methods.

TVM as a defense can be implemented by randomly picking pixels from an input and performing iterative optimization to find an image whose colors are consistent with the randomly picked pixels. Concretely, this can be implemented by solving

\begin{equation} \label{equations:tvm_optimization}
\arg\min_{\hat{x}} ||(1 - \tilde{x}) \odot (\hat{x} - x)||_2 + \lambda \cdot TV(\hat{x}),
\end{equation}

\noindent where $\tilde{x}$ is Bernoulli random variables (i.e., either 0 or 1) sampled at each pixel location in the input $x$ (i.e., $\tilde{x}$ has the same dimension as $x$), and $\lambda$ is a weighting constant. Here, the total variance $TV(\hat{x})$ is calculated by

\begin{equation} \label{equations:tvm}
TV(\hat{x}) = \sum_{k=1}^{Ch} \Big[ \sum_{i=2}^{R} ||\hat{x}_{(i,:,k)}-\hat{x}_{(i-1,:,k)}||_p + \sum_{j=2}^{C} ||\hat{x}_{(:,j,k)}-\hat{x}_{(:,j-1,k)}||_p \Big],
\end{equation}

\noindent where $R$, $C$, and $Ch$ denote the number of rows, columns, and channels, respectively. In their work, \citet{guo2018countering} set $p=2$. The intuition behind performing total variance minimization is to encourage spatial smoothness, which may lead to the removal of adversarial perturbations.  

Image quilting involves reconstructing an image using small patches taken from some database. As an adversarial defense method, one needs to ensure that the database only contains non-adversarial patches. Given a patch from an image, one can first find the nearest neighbors of the patch (e.g., using kNN) in the database according to the distance in the pixel space. One of the patches found is then randomly chosen to replace the original patch. The intuition behind image quilting is to construct an image that is free of adversarial perturbations, since quilting only uses clean patches to reconstruct the image. In addition, quilting also adds some randomization components during the process of selecting clean patches. 

\citet{guo2018countering} showed that total variance minimization and quilting outperform other image transformation methods when tested against FGSM~\citep{43405}, BIM~\citep{DBLP:journals/corr/KurakinGB16,45816}, DeepFool~\citep{DBLP:journals/corr/Moosavi-Dezfooli15}, and C\&W~\citep{DBLP:journals/corr/CarliniW16a} attacks when the model is also trained on transformed images. However, there are some tradeoffs of using these defenses. For example, total variance minimization causes slower inference since it needs to solve the optimization for every input, while image quilting degrades the accuracy of the model on non-adversarial inputs. Furthermore, these defense methods were not evaluated in whitebox settings where an attacker can adapt the attack strategy based on the defense used, and were later found to be circumventable using combination of BPDA~\citep{obfuscated-gradients} through the identity function (see Section~\ref{BPDA}) and EOT (see Section~\ref{EOT})~\citep{pmlr-v80-athalye18b}.

\subsubsection{Thermometer Encoding} \label{THERMOMETER} 







Thermometer Encoding~\citep{buckman2018thermometer} (TE) is a defense based upon the hypothesis that the linearity of neural networks render them vulnerable to attack~\citep{43405}. TE quantizes and discretizes the input data, effectively drowning out the effect of small perturbations of the kind typically introduced by adversarial attacks. Formally, the input signal is first quantized according to the desired discretization level $l$. Given the $i$-th element of an input signal $x_i$, the quantization function $q(x_i)$ returns the largest index $k \in \{1,...,l\}$ where $x_i < \frac{k}{l}$. For example, if $l=10$, then $q(0.72) = 8$.

The thermometer vector $\tau$ can then be found by

\begin{equation} \label{equations:thermometer_encoding}
  \tau(q(x_{i}))_k =
  \begin{cases}
  1 & \text{if $k \geq q(x_i)$ }\\
  0 & \text{otherwise}
  \end{cases},
\end{equation}

\noindent where $k \in \{1,...,l\}$ is the index of the thermometer encoded vector, i.e., $\tau(q(x_i)) \in \mathbb{R}^l$. For example, if $l = 10$, then $\tau(q(0.13)) = [0111111111]$. For simplicity, we denote $T_e = \tau(q(x_i))$.

Intuitively, small perturbations that are less than the minimum discretization level will not have an effect on the thermometer-encoded signals (e.g., for $l = 10$, $x_i = 0.72$ or $x_i = 0.75$ will produce same encoded signal, that is $T_e = [0000000111]$). Furthermore, TE works as a defense because the thermometer encoding is non-linear and non-differentiable, which makes it harder for the attacker to compute the gradients.

\citet{buckman2018thermometer} showed that the combination of TE and adversarial training~\citep{43405} (see Section~\ref{ADV_TRAINING}) yields high adversarial robustness, and even surpasses PGD adversarial training~\citep{madry2018towards} (see Section~\ref{PGD_ADV}). In their experiment, $l$ was set to 16, and the evaluations were done on MNIST, CIFAR10, CIFAR100, and SVHN datasets. Nevertheless, TE is considered to rely on gradient masking and circumventable using BPDA~\citep{obfuscated-gradients} (see Section~\ref{BPDA}, where we also provide a concrete example on how to circumvent TE with BPDA as shown by~\citet{obfuscated-gradients}), which approximates thermometer encoding function in a differentiable form, allowing an attacker to approximate the gradients of the model.

\subsubsection{PixelDefend} \label{PIXELDEFEND} 

\citet{song2018pixeldefend} introduced PixelDefend, which aims to ``purify'' an input before passing it to the classifier. They argued that adversarial examples lie in the low probability regions of the training data distribution which lead a model to suffer from covariate shift~\citep{SHIMODAIRA2000227}. PixelDefend uses a generative model trained only on clean data in order to approximate the distribution of the data. The trained generative model should then output data that represents the distribution of the clean training data even when dealing with adversarial inputs. The principle here is strongly reminiscent of MagNet (see Section~\ref{MAGNET}), although in this case a single model plays the role of both detector and reformer.

In PixelDefend, the generative model used was the PixelCNN~\citep{NIPS2016_6527,Salimans2017PixeCNN}. Note that for an 8-bits RGB image $x$, $P_{CNN}(x)$ is in $\mathbb{R}^{i \times j \times c \times dr}$ where $(i,j,c,dr)$ denotes the (row, column, channel, dynamic range) of the input. The dynamic range in this case is 256.



\begin{algorithm}[h]
   \caption{PixelDefend~\citep{song2018pixeldefend}}
   \label{alg:pixeldefend}
\begin{algorithmic}
   \STATE {\bfseries Input:} image $x \in \mathbb{R}^{i \times j \times c}$, pre-trained PixelCNN $P_{CNN}$, defense parameter $\epsilon_{defend}$
   \STATE {\bfseries Output:} purified image $\hat{x}$
   \STATE {\bfseries Initialize:} $\hat{x} \leftarrow x$
   \FOR{each row $i$}
   \FOR{each column $j$}
   \FOR{each channel $c$}
   \STATE range $R \leftarrow [max(x_{i,j,c} - \epsilon_{defend}, 0), min(x_{i,j,c} + \epsilon_{defend}, 255)]$
   \STATE $\hat{x}_{i,j,c} \leftarrow \arg \max_{z \in R} P_{CNN}(\hat{x})_{i,j,c,z}$
   \ENDFOR
   \ENDFOR
   \ENDFOR
   \STATE {\bfseries Return:} $\hat{x}$
\end{algorithmic}
\end{algorithm}

Algorithm~\ref{alg:pixeldefend} describes how PixelDefend works. Note that one needs to set the value of $\epsilon_{defend}$, which controls the maximum change in pixel value during purification process such that $||\hat{x} - x||_{\infty} \leq \epsilon_{defend}$. If $\epsilon_{defend}$ is too large, the purified input may not look like the original input anymore, leading to misclassification. If $\epsilon_{defend}$ is too small, the purified input may not be completely free of adversarial perturbations. In an ideal situation, given that an attacker can generate adversarial examples where $||x' - x||_{\infty} \leq \epsilon_{attack}$, $\epsilon_{defend}$ should be set to be equal to $\epsilon_{attack}$.

In addition,~\citet{song2018pixeldefend} showed that PixelDefend can also be used to detect and reject adversarial examples instead of robustifying the target classifier by performing statistical hypothesis testing using a $p$-value. This is done by comparing the probability density of an input to the training examples under the generative model. 

PixelDefend was shown to be effective against various attacks such as FGSM~\citep{43405} (see Section~\ref{FGSM}), BIM~\citep{DBLP:journals/corr/KurakinGB16,45816} (see Section~\ref{BIM}), DeepFool~\citep{DBLP:journals/corr/Moosavi-Dezfooli15} (see Section~\ref{DEEPFOOL}), and C\&W~\citep{DBLP:journals/corr/CarliniW16a} (see Section~\ref{CW}). Nevertheless, PixelDefend was later found to be vulnerable to BPDA~\citep{obfuscated-gradients} (see Section~\ref{BPDA}) and SPSA~\citep{pmlr-v80-uesato18a} (see Section~\ref{SPSA}).

\subsubsection{Defense-GAN} \label{DEFENSEGAN} 

Similar to PixelDefend~\citep{song2018pixeldefend} (see Section~\ref{PIXELDEFEND}), a concurrent work called Defense-GAN \citep{samangouei2018defensegan} uses generative model trained only on clean data as a way to remove adversarial perturbations. Defense-GAN uses a generative transformation network $G$ that is trained using the GAN framework~\citep{NIPS2014_5423} with Wasserstein loss~\citep{pmlr-v70-arjovsky17a}. Projection of an adversarial example onto the manifold of the trained generator is believed to remove adversarial components from an adversarial input.

\begin{figure}[h]
\centering
\includegraphics[width=0.7\textwidth]{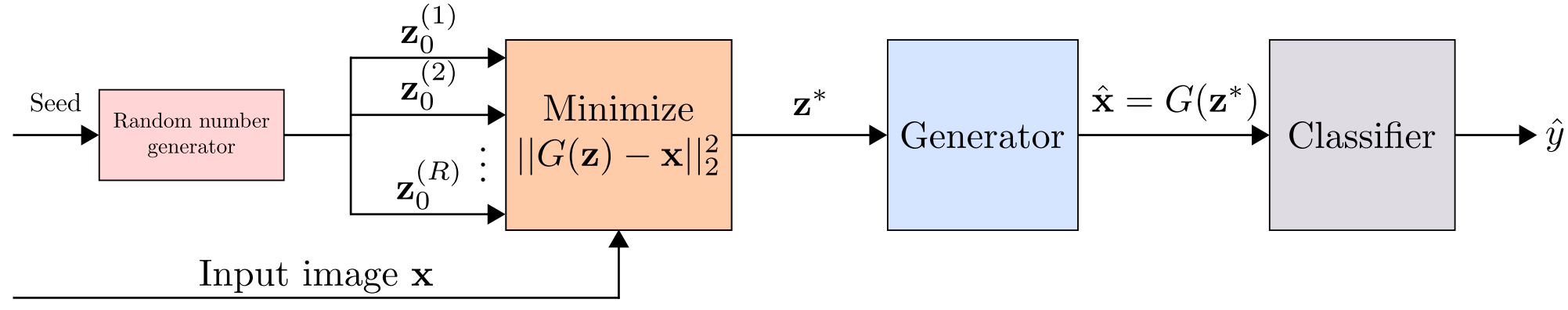}
\caption{Illustration of Defense-GAN (taken from~\citet{samangouei2018defensegan}).}
\label{fig:defense_gan}
\end{figure}

Defense-GAN works as follows. Defense-GAN first initializes $R$ random vectors, i.e., $Z = \{z_0,...,z_R\}$, and tries to find $z^* = \arg \min_{z \in Z^*} ||G(z)-x||^2_2$, where $Z^*$ is found by performing gradient descent for each $z \in Z$ to minimize $||G(z) - x||^2_2$. Once done, $z^*$ is used to generate the input instance by passing it through the generator, which will then be classified. Although the extra gradient descent steps increases computational complexity,~\citet{samangouei2018defensegan} argued that Defense-GAN is harder to be attacked by gradient-based attacks since one has to ``unroll'' the gradient descent iterations.

\citet{samangouei2018defensegan} showed that Defense-GAN can also be used to detect and reject adversarial examples. For example, this can be done by performing threshold-based detection on the value of $||G(z^*) - x||^2_2$. They showed how this method can be used to reliably detect FGSM adversaries on MNIST~\citep{726791} and Fashion MNIST (F-MNIST)~\citep{DBLP:journals/corr/abs-1708-07747} datasets with fairly high AUC score.

Defense-GAN was found to be more stable compared to MagNet~\citep{Meng:2017:MTD:3133956.3134057} and adversarial training~\citep{43405} (see Section~\ref{ADV_TRAINING}) when tested on various models on MNIST and F-MNIST datasets in blackbox setting against FGSM-SBA~\citep{DBLP:journals/corr/PapernotMGJCS16} (see Section~\ref{SUBSTITUTE}) although the robustness of adversarial training is sometimes higher. Furthermore, Defense-GAN showed superior performance compared to MagNet and adversarial training when tested against FGSM~\citep{43405} (see Section~\ref{FGSM}), R+FGSM~\citep{tramèr2018ensemble} (see Section~\ref{R+FGSM}), and C\&W $L_2$~\citep{DBLP:journals/corr/CarliniW16a} (see Section~\ref{CW}) attacks on both MNIST and F-MNIST. Nevertheless,~\citet{obfuscated-gradients} found that Defense-GAN is not as effective on defending CIFAR10 models. Moreover, just as with PixelDefend, Defense-GAN was also found to be vulnerable to BPDA~\citep{obfuscated-gradients} (see Section~\ref{BPDA}).

\subsubsection{WRM: A Distributionally Robust Optimization with Adversarial Training} \label{WRM} 

While adversarial training is relatively easy to implement and reasonably effective, it does not guarantee that an adversarially trained model will be robust against all types of adversaries, nor does it guarantee robustness against some particular type of $\epsilon$-budgeted adversaries. Formal verification techniques of machine learning models that guarantee robustness abound~\citep{DBLP:journals/corr/HuangKWW16, DBLP:journals/corr/abs-1711-00851, DBLP:journals/corr/abs-1801-09344}. However, most of these techniques are in general intractable.~\citet{sinha2018certifiable} proposed a principled adversarial training method called WRM\footnote{\citet{sinha2018certifiable} refer to their method as WRM but we are not entirely sure what WRM stands for.} that balances tractability with provable robustness guarantees, at least to some small $\epsilon$-bounded perturbations under certain similarity metric, that is based on Wasserstein distributional robustness.

\citet{sinha2018certifiable} view adversarial defense as a distributionally robust optimization problem. More explicitly, given training distribution $X$ and another data distribution around it $X'$, if we denote by $\theta$ as the parameters of a neural network model, $(x',y') \in X'$ as an input-label pair, and some loss function $\mathcal{L}(\cdot)$, the goal is to optimize the following objective function:

\begin{equation} \label{equations:robust_opt}
\arg \min_{\theta} \sup_{X'} \mathbb{E}_{X'} [\mathcal{L}(x',y')].
\end{equation}

\citet{sinha2018certifiable} first considered some robustness region under the Wasserstein metric $W_c(\cdot)$ where $W_{c}(X',X) \leq \rho$, and stated that solving Equation~\ref{equations:robust_opt} for some arbitrary $\rho$ is not tractable. They then considered the Lagrangian relaxation of the problem, and the objective for the inner maximization now becomes

\begin{equation} \label{equations:robust_opt_relaxed}
\arg \min_{\theta} \Big\{ \sup_{X'} \mathbb{E}_{X'} [\mathcal{L}(x',y')] - \gamma W_{c}(X',X) = \mathbb{E}_{X} [\sup_{x' \in \mathcal{Z}} \mathcal{L}(x',y') - \gamma \cdot c(x',x)] \Big\},
\end{equation}

\noindent where $x \in X$, $\gamma \geq 0$ is a penalty weight, $\mathcal{Z}$ is the universe of all possible examples, and $c(\cdot)$ is a cost for the attacker to modify $x$ into $x'$. Concretely,~\citet{sinha2018certifiable} considered $c(x,x') = ||x - x'||^2_p$, where $p \geq 1$. They then showed that the worst-case error of a model trained with WRM is bounded by

\begin{equation} \label{equations:wrm_certificate}
\sup_{X':W_{c}(X',X) \leq \rho} \mathbb{E}_{X'} [\mathcal{L}(x',y')] \leq \mathbb{E}_{X_{test}} [ \sup_{X'} \mathcal{L}(x',y) - \gamma \cdot c(x_{test},x') ] + \gamma \cdot \rho,
\end{equation}

\noindent where $X_{test}$ is the empirical distribution of test data, and $\rho = \mathbb{E}_{X_{test}} [ c(x',x_{test}) ]$ can be calculated empirically. Note that $x'$ is obtained by solving the inner maximization in Equation~\ref{equations:robust_opt_relaxed}.

\citet{sinha2018certifiable} showed that for smooth $\mathcal{L}(\cdot)$ and large enough $\gamma$ (or small enough $\rho$ by duality), the inner optimization problem is strongly concave and thus solvable using gradient descent. Another important assumption to note is that the activation function used by the neural networks has to be smooth to ensure convergence, so functions like the ReLU should not be used. In practice,~\citet{sinha2018certifiable} recommend to use either sigmoid or the Exponential Linear Units (ELU)~\citep{DBLP:journals/corr/ClevertUH15}. Keeping all these in mind, WRM was developed and the algorithm is described in Algorithm~\ref{alg:wrm}. It is worth to note that the guarantee of WRM does not hold for $p = \infty$.~\citet{sinha2018certifiable} then proposed a proximal algorithm to solve the inner optimization problem when $p = \infty$. We refer readers to the paper for the proximal algorithm. We also refer readers to the paper for the proofs of all the claims made in developing WRM.


\begin{algorithm}[h]
   \caption{WRM~\citep{sinha2018certifiable}}
   \label{alg:wrm}
\begin{algorithmic}
    \STATE {\bfseries Input:} A non-adversarial dataset $X$, number of iterations $I$, stepsize sequence $\alpha = \{ \alpha_{0},...,\alpha_{I-1} \}$
    \STATE {\bfseries Output:} Adversarially robust $\theta$
    \FOR{$i = 0$ to $I$}
    \STATE $x' \leftarrow \arg\max_{x'} \epsilon \cdot \mathcal{L}(x',y) - \gamma \cdot c(x,x')$, where $(x,y) \sim X$
    \STATE $\theta \leftarrow \theta - \alpha_i \nabla_{\theta} \mathcal{L}(x',y')$
    \ENDFOR
\end{algorithmic}
\end{algorithm}

They empirically showed that the worst-case error of WRM is always below what the certificate of robustness guarantees when tested on MNIST for $p=2$ and $\gamma = 0.3684$. Furthermore, WRM was shown to be more robust compared to various existing adversarial training methods when evaluated against $L_2$ and $L_{\infty}$ adversaries for small $\epsilon$. Moving forward, methods that can provide guarantee for higher robustness and works in more complex datasets are still needed.

\subsubsection{High-Level Representation Guided Denoiser} \label{HGD} 

\citet{DBLP:journals/corr/abs-1712-02976} argued that in order to mitigate adversarial perturbations, one needs to be able to denoise the adversarial examples. They suggested that this can be done by training a denoiser $D$ that learns to generate negative noise maps $d\hat{x}$ given an image as input, i.e., $D(x) = d\hat{x}$. The function of the generated negative noise maps is to remove adversarial noise when added to the adversarial examples $x'$, where the denoised image is calculated by $\hat{x} = x' - d\hat{x}$. In practice, the denoiser that~\citet{DBLP:journals/corr/abs-1712-02976} used follows the architecture of U-Net~\citep{10.1007/978-3-319-24574-4_28}. 

The training objective of the denoiser is to minimize the difference between the high level representations of the inputs and the denoised inputs, hence the name High-Level Representation Guided Denoiser (HGD). Formally, the denoiser is trained to minimize $||f_l(\hat{x}) - f_l(x)||_1$
, where $f_l(x)$ denotes the output of the classification convolutional neural network $f$ at the $l$-th layer.~\citet{DBLP:journals/corr/abs-1712-02976} experimented with two different choices of $l$: the last convolutional layer and the logits layer of $f$. These models were referred as the Feature Guided Denoiser (FGD) and the Logits Guided Denoiser (LGD), respectively. Another variant of HGD was also introduced by minimizing the classification loss of $f$ called the Class Label Guided Denoiser (CGD). These are illustrated in Fig.~\ref{fig:hgd}.


\def\fs{0.32}
\begin{figure}[ht]
   \begin{minipage}{0.5\linewidth}
   \centering
   \subfloat[FGD/LGD]{\label{hgd:a}\includegraphics[scale=\fs]{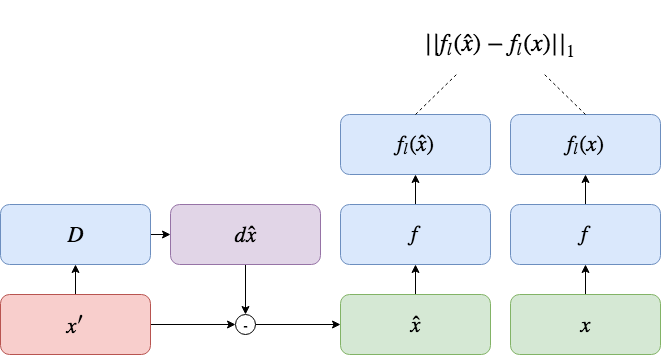}}
   \end{minipage}%
   \begin{minipage}{0.5\linewidth}
   \centering
   \subfloat[CGD]{\label{hgd:b}\includegraphics[scale=\fs]{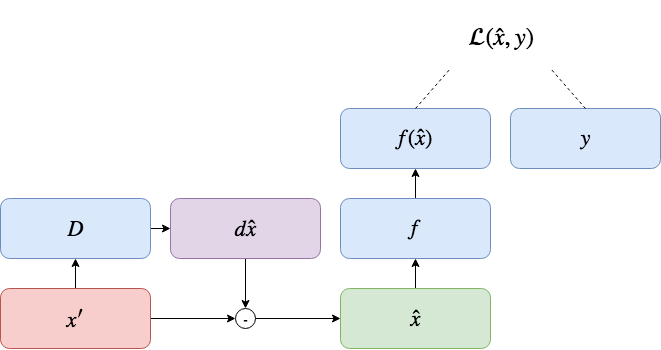}}
   \end{minipage}
   \caption{Illustration of FGD, LGD, and CGD (adapted from~\citet{DBLP:journals/corr/abs-1712-02976}).}
   \label{fig:hgd}
\end{figure}

In their experiments, the adversarial examples used to train the denoiser were generated using FGSM \citep{43405} (see Section~\ref{FGSM}) and BIM~\citep{DBLP:journals/corr/KurakinGB16,45816} (see Section~\ref{BIM}) on various ImageNet models. This resulted in 210,000 training data, consisting of 30K clean inputs randomly sampled from ImageNet training set (30 images per class), and 180K adversarial examples. The denoiser can be trained using only these 210K examples, conspicuously more efficient compared to adversarial training~\citep{43405} (see Section~\ref{ADV_TRAINING}) and ensemble adversarial training method~\citep{tramèr2018ensemble} (see Section~\ref{ENSEMBLE_ADV_TRAINING}). HGD requires only 30 training epochs on the 210K examples, a huge improvement in speed relative to ensemble adversarial training on ImageNet which usually requires more than 150 epochs on much larger size of training examples.

\citet{DBLP:journals/corr/abs-1712-02976} suggested that LGD is the best performer compared to FGD and CGD, giving a good balance between correctly classifying clean and adversarial inputs. This defense method was shown to be transferable, in the sense that denoiser that is trained using a particular classifier can also be used to defend other classifiers. HGD is the winning submission for the NIPS 2017 Defense Against Adversarial Attack Competition~\citep{DBLP:journals/corr/abs-1804-00097}. Nevertheless, just like PixelDefend~\citep{song2018pixeldefend} (see Section~\ref{PIXELDEFEND}), HGD was also found to be vulnerable to SPSA~\citep{pmlr-v80-uesato18a}.


\subsubsection{Adversarial Logit Pairing} \label{ALP} 

\citet{DBLP:journals/corr/abs-1803-06373} proposed a logit pairing defense mechanism that encourages the logits of input pairs (i.e., adversarial and non-adversarial input pairs) to be similar. Two different logit pairing strategies were introduced: Adversarial Logit Pairing (ALP) and Clean Logit Pairing (CLP). The motivation behind ALP is to enforce logits invariance between a \emph{clean input and its adversarial counterpart}, while CLP enforces logits invariance between \emph{any pairs of inputs}. Surprisingly, the latter approach yields significant robustness improvements, despite the fact that examples are frequently drawn from disparate classes.

Formally, ALP introduces a regularization term that penalizes for the differences of logits in minibatches of clean inputs $\{x_1,...,x_{\frac{n}{2}}\}$ and their adversarial counterparts $\{x'_1,...,x'_{\frac{n}{2}}\}$ generated by PGD attack~\citep{madry2018towards} (see Section~\ref{PGD_ADV}), where $n$ denotes the size of the minibatch. Given $Z(x)$ as the logits vector, the training objective becomes


\begin{equation} \label{equations:alp_objective}
\alpha \cdot \frac{2}{n} \sum_{i=1}^{\frac{n}{2}} \mathcal{L}(x_i,y_i) + (1-\alpha) \cdot \frac{2}{n} \sum_{i=1}^{\frac{n}{2}} \mathcal{L}(x'_i, y_i) + \lambda \cdot \frac{2}{n} \sum_{i=1}^{\frac{n}{2}} \mathcal{L}_{LP}(Z(x_i),Z(x'_i)),
\end{equation}

\noindent where the first two terms denote the adversarial training loss, similar to Equation~\ref{equations:adversarial_training}, and $\mathcal{L}_{LP}$ denotes the logit pairing loss (e.g.,~\citet{DBLP:journals/corr/abs-1803-06373} used the $L_2$ loss in their experiments).

\citet{DBLP:journals/corr/abs-1803-06373} showed that the combination of adversarial training on PGD attack with ALP resulted in state of the art robustness against both whitebox (variants of FGSM~\citep{43405} (see Section~\ref{FGSM}) and PGD~\citep{DBLP:journals/corr/KurakinGB16,45816} (see Section~\ref{BIM}) attacks), as well as blackbox attacks (e.g., SBA~\citep{DBLP:journals/corr/PapernotMGJCS16} (see Section~\ref{SUBSTITUTE})) on ImageNet models. Note that scaling adversarial defense methods to large datasets such as ImageNet has been shown to be challenging and computationally expensive~\citep{45816}.

CLP, on the other hand, compares the logits between \emph{random} pairs of clean inputs that \emph{do not have to come from the same class}. The training objective for CLP is defined as:

\begin{equation} \label{equations:clp_objective}
\sum_{i=1}^{n} \mathcal{L}(x_i,y_i) + \lambda \cdot \frac{2}{n} \sum_{i=1}^{\frac{n}{2}} \mathcal{L}_{LP}(Z(x_i),Z(x_{i+\frac{n}{2}})),
\end{equation}

\noindent where the first term denotes the cross-entropy loss for classification task without including any adversarial examples in the training set. CLP was initially introduced as a way to understand the contribution of the logit pairing itself without the advantage of performing adversarial training.

Surprisingly, they found that the use of CLP alone leads to increase in adversarial robustness. They hypothesized that CLP implicitly encourages the norm of the logits to be small, which prevents the model from being overconfident of its predictions. Building upon the unexpected success of CLP, Clean Logit Squeezing (CLS) was introduced as a method that explicitly penalizes the norm of the logits. The training objective of CLS is

\begin{equation} \label{equations:cls_objective}
\sum_{i=1}^{n} \mathcal{L}(x_i,y_i) + \lambda \cdot \frac{1}{n} \sum_{i=1}^{n} ||Z(x_i)||_p,
\end{equation}

\noindent where $p$ denotes the $p$-norm. Directly penalizing the loss with the norm of the logits through CLS was shown to also increase adversarial robustness.


Nevertheless,~\citet{DBLP:journals/corr/abs-1807-10272} showed that ALP can be circumvented by PGD if the attacker performs enough PGD steps. They argued that there are two main differences between ALP and PGD adversarial training which may cause ALP to not have the same level of robustness with PGD adversarial training although ALP is built on top of it. The first difference is that ALP violates the robust optimization formulation by including clean examples during the training process. Note that robust optimization formulation suggests one to \emph{only} optimize on the worst-case scenarios (see Section~\ref{PGD_ADV}). The second difference is that~\citet{DBLP:journals/corr/abs-1803-06373} include targeted adversarial examples during training, which also alters the inner maximization formulation. Finally,~\citet{DBLP:journals/corr/abs-1807-10272} also showed some loss surface visualizations of a model trained with ALP and pointed that the loss surface is ``bumpier'' compared to the loss surface of some baseline models and with small gradients around the starting point, which can be considered as a form of gradient masking. 

\subsubsection{Fortified Networks} \label{FORTIFIED} 

While many denoising-based defense methods such as Defense-GAN~\citep{samangouei2018defensegan} (see Section~\ref{DEFENSEGAN}) and PixelDefend~\citep{song2018pixeldefend} (see Section~\ref{PIXELDEFEND}) operate in the input space, Fortified Networks (FN)~\citep{lamb+al-2018-fortified} is a denoising-based defense that operates in the hidden layers of a neural network.~\citet{lamb+al-2018-fortified} argued that one of the advantages of performing denoising in the hidden layers instead of the input space is to allow the use of simple generative transformation models such as the Denoising Autoencoders (DAE)~\citep{VincentPLarochelleH2008} which have been previously shown to not work well as a defense mechanism when used in the input space~\citep{DBLP:journals/corr/GuR14}. This is motivated by the evidence that hidden representations of a neural network have simpler statistical properties compared to the features in input space~\citep{pmlr-v28-bengio13}.

FN applies one or more DAEs to the hidden layer of a neural network. Formally, we denote decoded representations at the $l$-th layer by

\begin{equation} \label{equations:fortified_dae}
h^l_{decoded} = DAE^l(h^l + \eta),
\end{equation}

\noindent where $\eta \sim \mathcal{N}(\mu,\sigma^2)$ denotes the Gaussian noise, and $DAE^l$ denotes the DAE at the $l$-th layer. The DAEs are trained to minimize reconstruction error, or in other words

\begin{equation} \label{equations:fortified_dae_loss}
\mathcal{L}_{DAE} = ||DAE^l(h^l) - h^l||_2,
\end{equation}

\noindent where the weights for the encoder and decoder of each DAE are tied. In their experiments,~\citet{lamb+al-2018-fortified} combined FN with adversarial training (see Section~\ref{ADV_TRAINING}). For the DAE, the target reconstruction for an adversarial example $x'$ at layer $l$ is the hidden representation of the clean input $x$ at the same layer. Formally,

\begin{equation} \label{equations:fortified_adv_dae_loss}
\mathcal{L}_{AdvDAE} = ||DAE^l(h^l_{adv}) - h^l||_2,
\end{equation}

\noindent where $h^l_{adv}$ is the hidden representation at layer $l$ when the input is an adversarial example. The objective function of FN is then:

\begin{equation} \label{equations:fortified_loss}
\mathcal{L}_{c} + \lambda_1 \mathcal{L}_{DAE} + \lambda_2 \mathcal{L}_{AdvDAE},
\end{equation}

\noindent where $\mathcal{L}_{c}$ denotes the classification loss function used by the network (e.g., cross-entropy) when dealing with both clean and adversarial inputs, while $\lambda_1$ and $\lambda_2$ are the weighting factors for the loss terms.

\citet{lamb+al-2018-fortified} found that the use of multiple DAEs yields higher robustness and shows improvement compared to the PGD adversarial training~\citep{madry2018towards} (see Section~\ref{PGD_ADV}) and Thermometer Encoding~\citep{buckman2018thermometer} (see Section~\ref{THERMOMETER}) against attacks like the FGSM~\citep{43405} (see Section~\ref{FGSM}) and PGD~\citep{madry2018towards} (see Section~\ref{PGD_ADV}) when combined with adversarial training on MNIST and CIFAR10 models. They also showed that FN is more robust compared to DefenseGAN~\citep{samangouei2018defensegan} (see Section~\ref{DEFENSEGAN}) when tested against FGSM and PGD attacks on both MNIST and F-MNIST models. 

\subsubsection{Feature Denoising Block} \label{FEAT_DENOISE} 

\citet{DBLP:journals/corr/abs-1812-03411} observed how adversarial perturbations cause large changes in the feature maps produced by a model compared to the original input. Motivated by this observation, they proposed a method to denoise the feature maps. This in fact is similar to the idea of fortified networks~\citep{lamb+al-2018-fortified} (see Section~\ref{FORTIFIED}).

\begin{figure}[h]
\centering
\includegraphics[width=0.2\textwidth]{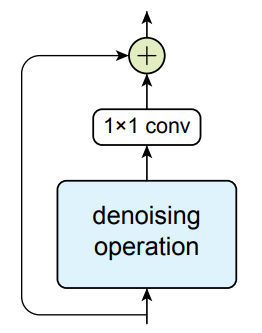}
\caption{Illustration of the denoising block (taken from~\citet{DBLP:journals/corr/abs-1812-03411}).}
\label{fig:fdb}
\end{figure}

In this work, the denoising is done by creating what they call a denoising block that can be placed at any levels in the network. Fig.~\ref{fig:fdb} illustrates the denoising block. For the rest of this paper, we refer to this defense as the Feature Denoising Block, or in short, FDB. This block consists of a differentiable denoising operator, a $1 \times 1$ convolutional layer, and an identity skip connection.~\citet{DBLP:journals/corr/abs-1812-03411} evaluated various denoising operators: non-local means~\citep{Buades:2005:NAI:1068508.1069066}, mean filtering, median filtering, and bilateral filtering~\citep{Tomasi:1998:BFG:938978.939190}. Note that the use of denoising operator may not only remove adversarial perturbations, but also information relevant for classification task. However,~\citet{DBLP:journals/corr/abs-1812-03411} argued that the identity skip connection should help to maintain the useful information, while the $1 \times 1$ convolutional layer will learn to balance between preserving the useful information and removing adversarial perturbations.

\citet{DBLP:journals/corr/abs-1812-03411} found non-local means denoiser as the best performing denoising operator when compared with other operators used in their experiments. The denoising block was also shown to not significantly sacrifice performances on non-adversarial inputs. Finally, they also showed that when combined with PGD adversarial training~\citep{madry2018towards} (see Section~\ref{PGD_ADV}), FDB achieved state of the art adversarial robustness performance in both whitebox and blackbox settings. This method was ranked first in the Competition on Adversarial Attacks and Defenses (CAAD) 2018 where it achieved 50.6\% accuracy when evaluated against 48 different attacks.

\subsubsection{Analysis by Synthesis} \label{ABS} 

The idea of using generative models as a defense mechanism has previously been proposed. Methods like Defense-GAN~\citep{samangouei2018defensegan} (see Section~\ref{DEFENSEGAN}) and PixelDefend~\citep{song2018pixeldefend} (see Section~\ref{PIXELDEFEND}) rely on generative models to learn the distribution of non-adversarial dataset with the goal of being able to project adversarial input to the learned non-adversarial manifold. However, instead of learning the input distribution for the whole dataset,~\citet{Schott2018a} proposed a defense method called Analysis by Synthesis (ABS) that learns the distribution of the input for each class.

ABS uses VAE~\citep{DBLP:journals/corr/KingmaW13} to model the distribution of each class in a dataset, where one VAE is needed for each class. Given a dataset with $C$ classes, during test time, one then needs to perform gradient descent to find the optimal latent variable that maximizes the lower bound of the log-likelihood for each class $y$. Formally, given decoder $p_{\theta}(x|z,y)$, we want to solve for

\begin{equation} \label{equations:abs_gradient_descent}
l^*_y = \max_{z} \log \Big[ p_{\theta}(x|z,y) \Big] - D_{KL} \Big[ \mathcal{N}(z,\sigma_{q}\mathds{1})|| \mathcal{N}(0,\mathds{1})\Big].
\end{equation}

\noindent Concretely,~\citet{Schott2018a} solve this optimization problem using Adam~\citep{DBLP:journals/corr/KingmaB14}. Once $l^*$ for every class are found, the class probability can be calculated by

\begin{equation} \label{equations:abs_inference}
p(y|x) = \frac{e^{\alpha l^*_y(x)} + \eta}{\sum_{c=0}^{C-1} \Big( e^{\alpha l^*_c(x)} + \eta \Big)},
\end{equation}

\noindent where $\alpha$ is a scaling factor, and $\eta$ is an offset term.~\citet{Schott2018a} explained that the reason of adding $\eta$ is to encourage the model to produce uniform distribution over classes in the case of unrecognizable images. In their implementation, $\eta$ was chosen so that the median of the confidence on non-adversarial examples is 0.9. They also introduced an ABS variant called Binary ABS that can be used for grayscale image datasets (e.g., MNIST), which binarized the input image according to a certain threshold value. 

ABS was shown to work fairly well and more robust than the PGD adversarial training~\citep{madry2018towards} (see Section~\ref{PGD_ADV}) when evaluated against $L_{0}$ and $L_{2}$ adversaries on MNIST. However, the robustness of ABS on $L_{\infty}$ adversary seems to be very low. Although they demonstrated the robustness of Binary ABS on $L_{\infty}$ adversaries, Binary ABS cannot be applied to non-grayscale images. It also seems challenging to scale ABS to large datasets that have many classes, since one needs to train multiple VAEs and perform gradient descent during inference for each class. These limitations were properly acknowledged by the authors, and further evaluations of this defense on larger datasets are still needed.

\subsubsection{Web-Scale Nearest-Neighbor Search} \label{WSNNS} 

Assuming adversarial perturbations push an image away from its manifold,~\citet{DBLP:journals/corr/abs-1903-01612} developed a defense with the aim to project an adversarial image back to the true manifold via Web-Scale Nearest-Neighbor Search (WSNNS). The general idea of WSNNS is to find the $K$ nearest neighbors of an input image from a large database of images, and classifying these neighbors instead of the given input image. Note that the images in the database must be selected to match the task at hand, which can be obtained by looking at their labels, hashtags, or any other available indicators. One may see the similarity between WSNNS and quilting-based defense~\citep{guo2018countering} (see Section~\ref{TRANSFORM-DEFENSE}). The difference is that while quilting-based defense operates at a patch-level, WSNSS operates at image-level.

\citet{DBLP:journals/corr/abs-1903-01612} found the $K$ nearest neighbors according to the Euclidean distance in the feature space. In their work, they experimented with various outputs of a certain layer in deep neural networks which may be followed by pooling and dimensionality reduction method (e.g., PCA), as the features. Once the nearest neighbors are identified, all these images are classified where each prediction may be weighted differently. Various weighting strategies were proposed by~\citet{DBLP:journals/corr/abs-1903-01612}. The first strategy is to assign the weights uniformly across predictions (UW strategy). Another strategy is to adjust the weights based on the entropy gap between the prediction and uniform prediction (CBW-E strategy). For CBW-E strategy, given the weights $w$ and $C$ number of classes, $w$ can be computed by

\begin{equation} \label{equations:cbw-e}
w = \Big| \log C + \sum_{c=1}^C s_c \log s_c \Big|,
\end{equation}

\noindent where $s_c$ is the softmax probability of class $c$. Finally, the weights can also be computed based on the difference between the maximum value in a prediction and the next top $M$ values (CBW-D strategy):

\begin{equation} \label{equations:cbw-d}
w = \sum_{m=2}^{M+1} (\hat{s}_1 - \hat{s}_m)^P,
\end{equation}

\noindent where $\hat{s}_m$ is the top $m$-th softmax value. In their experiments, $M$ and $P$ were set to be $20$ and $3$, respectively.

\citet{DBLP:journals/corr/abs-1903-01612} found that CBW-D strategy produces more robust defense compared to UW and CBW-E. Although increasing $K$ generally leads to a more robust defense, they also found that the effect of increasing $K$ saturates at a certain point, thus $K$ can be experimentally adjusted depending on the database used and other parameters. They also discussed the tradeoffs between using features from earlier layers versus later layers of deep networks, where they found that using features from earlier layers generally results in higher robustness while using features from later layers produces higher accuracy but lower robustness. Finally, they showed that increasing number of images in the database leads to higher accuracy with a log-linear relation, and engineering the database to only include images that are related to the task at hand can significantly improve the performance. WSNSS was shown to be effective to defend against PGD adversaries in blackbox and graybox settings, where graybox was defined as the case when the adversarial examples was generated by targeting the same model using the substitute blackbox attack~\citep{DBLP:journals/corr/PapernotMGJCS16} (see Section~\ref{SUBSTITUTE}).

This defense however is not effective to defend against whitebox attacks (i.e., if the attacker has information regarding the defense in place or the database used). Note that the neighbors are determined by the Euclidean distance in the feature space, which is not a robust metric to measure similarity between samples. Hence, the attacker may be able to find a perturbation that will cause an image that belongs to a certain class to be considered close to other images from different class. For example, perhaps AMDR~\citep{DBLP:journals/corr/SabourCFF15} (see Section~\ref{AMDR}) can be used to attack this model. Another possibility is that an attacker may create a substitute database and aims to find an adversarial example that maximizes the sum of the differences between the features of the given image and the features of the images in the substitute database. Another option is for the attacker to maximize the prediction error of both the original input and images in the substitute database. Both variants of these attacks were proposed by~\citet{DBLP:journals/corr/abs-1903-01612} as the nearest-neighbor feature-space attack and nearest-neighbor prediction attack, respectively. Finally, this defense may also exhibit the weaknesses of $K$ nearest neighbors algorithm which it relies heavily on.

\subsubsection{ME-Net} \label{MENET} 

\citet{pmlr-v97-yang19e} proposed a preprocessing-based defense called ME-Net that preprocesses an input with the hope of destroying the structure of adversarial noise. ME-Net works by first dropping pixels in an input image randomly according to some probability $p$, which is hypothesized to destroy the adversarial perturbations. The image is then reconstructed using Matrix Estimation (ME) algorithms, which is a family of methods to recover data of a matrix from its noisy observation.

\begin{figure}[h]
\centering
\includegraphics[width=0.75\textwidth]{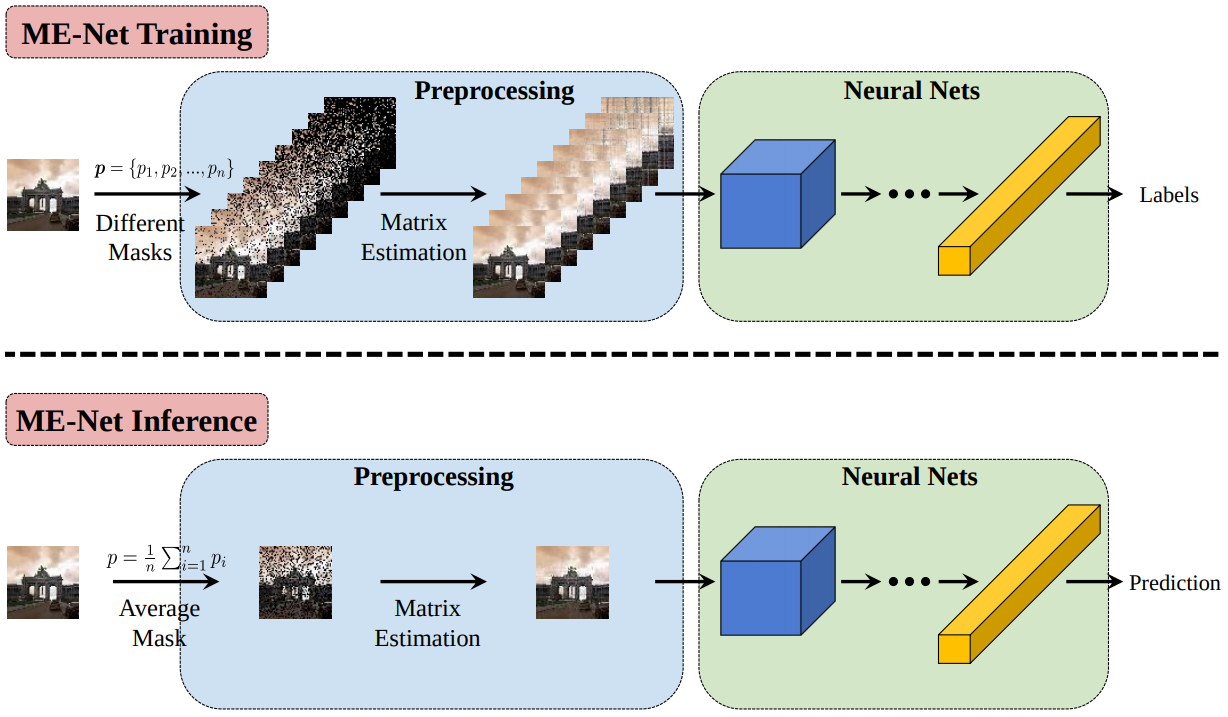}
\caption{Illustration of ME-Net (taken from~\citet{pmlr-v97-yang19e}).}
\label{fig:me_net}
\end{figure}

Fig.~\ref{fig:me_net} illustrates the training and inference mechanics of ME-Net. Given a training dataset of images, each image is masked $n$-times to produce $n$ images according to a set of masking probabilities $\textbf{p} = \{p_1,p_2,...,p_n\}$, where $1 - p_i$ is the probability of each pixel to be set to zero, and $\textbf{p}$ is initialized randomly. Next, each of the masked images is reconstructed using an ME algorithm. All these preprocessed images are then gathered to form a new dataset for the model to be trained on. As for the inference, given an image, the image is first masked based on the average of the masking probabilities used during training (i.e., $p = \frac{1}{n}\sum_{i=1}^n p_i$) before being preprocessed using an ME algorithm. The preprocessed image is then passed to the classifier to be classified.

In their experiments,~\citet{pmlr-v97-yang19e} considered Nuclear Norm Minimization~\citep{Candes:2012:EMC:2184319.2184343}, Soft-Impute~\citep{Mazumder:2010:SRA:1756006.1859931}, and Universal Singular Value Thresholding (USVT)~\citep{chatterjee2015} as the ME algorithms. They also chose to use $10$ masks. When comparing the three ME algorithms, they found that the robustness provided by these algorithms to be comparable. Thus, they suggested to use USVT due to its efficiency compared to the other ME algorithms.

ME-Net was shown to be robust when tested against various $L_{\infty}$ attacks in both blackbox and whitebox settings on CIFAR-10, MNIST, SVHN, and Tiny-ImageNet. In the case of blackbox attack, ME-Net was evaluated against various variants of the substitute blackbox attack~\citep{DBLP:journals/corr/PapernotMGJCS16} (see Section~\ref{SUBSTITUTE}) (CW-SBA~\citep{DBLP:journals/corr/CarliniW16a}, FGSM-SBA~\citep{43405}, PGD-SBA~\citep{madry2018towards}), Boundary Attack~\citep{brendel2018decisionbased} (see Section~\ref{BOUNDARY}), and SPSA~\citep{pmlr-v80-uesato18a} (see Section~\ref{SPSA}), where the model successfully maintained its $\sim90\%$ accuracy on CIFAR10 under all these threats. In whitebox settings, ME-Net successfully defends itself against PGD-based BPDA~\citep{obfuscated-gradients} (see Section~\ref{BPDA}). Note that the evaluation of ME-Net against BPDA is appropriate considering the non-differentiable nature of the preprocessing pipeline. It will be interesting to also evaluate the robustness of the model against the combination of BPDA and expectation over randomness, as this has been shown to successfully circumvent various preprocessing-based defenses. Finally,~\citet{pmlr-v97-yang19e} showed that ME-Net can be combined with adversarial training which further enhances the robustness of the model.

\subsection{Detecting Adversarial Attacks} \label{DETECTION}

Instead of trying to classify adversarial examples correctly, several works have shown progress in the detection of adversarial examples. A sample can then be rejected if it was detected as an adversarial example. In this section, we cover several adversarial detection methods.

\subsubsection{Early Methods: PCA, Softmax, and Reconstruction of Adversarial Images} \label{EARLY_METHODS} 

\citet{DBLP:journals/corr/HendrycksG16b, DBLP:journals/corr/HendrycksG16c} investigated three methods to perform adversarial detection. Broadly speaking, these methods capitalize upon empirical differences between clean and perturbed issues, and are all vulnerable to attacks which specifically aim to circumvent them. Together, we refer to these as H \& G detection methods.

The first method proposed is to analyze the variance of coefficients of PCA whitened input.~\citet{DBLP:journals/corr/HendrycksG16b} found that the later principal components of adversarial images usually have larger variance than those of clean inputs, and thus can be used as an indication to perform threshold based detection. They showed that this method can be used to detect FGSM~\citep{43405} (see Section~\ref{FGSM}) and BIM~\citep{DBLP:journals/corr/KurakinGB16,45816} (see Section~\ref{BIM}) adversaries on MNIST, CIFAR10, and Tiny-ImageNet when the attacker is not aware of the defense in place. 

However,~\citet{DBLP:journals/corr/CarliniW17} found that this method can be bypassed if the attacker is aware of the defense strategy, allowing them to constrain the variance of later principal components during adversarial examples generation process. For example, this can be achieved using C\&W attack~\citep{DBLP:journals/corr/CarliniW16a} (see Section~\ref{CW}) by folding the constraint to the objective function. Furthermore,~\citet{DBLP:journals/corr/CarliniW17} argued that the reason this method works on MNIST may be due to the certain artifacts that exist in the dataset. For example, the pixel values that are not part of the digit have zero values in a clean image, while pixel values that are not part of the digit in an adversarial image usually no longer have zero values due to the modification to the image.~\citet{DBLP:journals/corr/CarliniW17} then evaluated this method on CIFAR10 and found that this method is not as effective as when evaluated on MNIST.

Second,~\citet{DBLP:journals/corr/HendrycksG16b, DBLP:journals/corr/HendrycksG16c} suggested that the softmax distribution between clean and adversarial inputs are different, and thus can be analyzed to perform adversarial detection. This can be done by measuring the Kullback-Leibler divergence~\citep{kullback1951} between uniform distribution and the softmax distribution, then perform threshold-based detection on it. They found that the softmax distribution of normal examples are usually further away from uniform distribution compared to adversarial examples since a model tends to predict an input with a high confidence, at least for adversaries generated by attack methods that do not care about the output distribution as long as the generated adversaries are misclassified.

This method seems to only applicable to specific attacks that stop as soon as an input becomes adversarial such as the JSMA~\citep{DBLP:journals/corr/PapernotMJFCS15} (see Section~\ref{JSMA}), which may be predicted by the model with a low confidence score. Furthermore, evaluation on ImageNet models is needed (and to date, incomplete) since the softmax probability on ImageNet dataset usually is less confident due to the large number of classes. It is also not clear whether or not this strategy will work against adversarial attacks that also target specific confidence score such as the C\&W attack~\citep{DBLP:journals/corr/CarliniW16a} (see Section~\ref{CW}).

Finally,~\citet{DBLP:journals/corr/HendrycksG16b, DBLP:journals/corr/HendrycksG16c} proposed that adversarial examples can be detected by analyzing the input reconstructions obtained by adding an auxiliary decoder~\citep{pmlr-v48-zhangc16} to the classifier model that takes the logits as an input to reconstruct the image. The decoder and classifier are jointly trained only on clean examples. The detection can be done by creating a detector network which takes the input reconstruction, logits, and confidence score as inputs, and outputs a probability of an input being adversarial or not, where the detector network is trained on both clean and adversarial examples.~\citet{DBLP:journals/corr/HendrycksG16b} evaluated this method on MNIST model and showed that this method is able to detect adversarial examples generated by FGSM and BIM. They showed qualitatively how reconstruction of adversarial examples are noisier compared to normal inputs, allowing one to compute the differences as a way to detect adversarial examples.

This defense method however is fully differentiable, and an attacker may still be able to find adversarial examples that fool both the classifier and detector networks in whitebox setting. Although this method needs to be evaluated more extensively on different datasets and attack settings, the use of input reconstruction that is conditioned on the classification seems to be an interesting research direction that can be used to detect both misclassification and adversarial examples.

\citet{DBLP:journals/corr/HendrycksG16b} did admit that these methods are not perfect, and may be circumventable, and in fact have successfully been circumvented. They then suggested that future detection methods should include combination of different detection methods. Indeed, some of the detection methods that we cover in this paper use combination of different strategies to enhance adversarial detection performance.

\subsubsection{Adversary Detector Networks} \label{AUG_DETECTOR} 


\citet{metzen2017detecting} proposed augmenting a pretrained neural network with a binary detector network. The detector network $D$ is a binary classifier network that is trained to differentiate between real and adversarial examples. It takes the output of the classifier network at a certain layer as an input and outputs the probability of an input being adversarial (i.e., $D(f_l(x)) = y_{adv} \ \textrm{or} \ y_{clean}$, where $f_l(x)$ denotes the output of classifier $f$ at the $l$-th layer).

The training of the detector network involves generating adversarial examples to be part of the training set for the detector network. In order to account for future attacks with the assumption that attacker have access to both the classifier and detector networks,~\citet{metzen2017detecting} generated examples which specifically attempted to fool the detector. They generated adversarial examples according to

\begin{equation} \label{equations:metzen}
x'_{i+1} =  Clip_{\epsilon} \Big\{ x'_i + \alpha \Big[ (1 - \sigma) \cdot \textrm{sign}(\nabla _x \mathcal{L}_{classifier}(x'_i,y)) + \sigma \cdot \textrm{sign}(\nabla _x \mathcal{L}_{detector}(x'_i, y_{adv})) \Big] \Big\},
\end{equation}

\noindent where $x'_0 = x$, and $\sigma$ denotes the weight factor that controls whether the attacker's objective is to attack the classifier or the detector, chosen randomly at every iteration. The training procedure used by~\citet{metzen2017detecting} followed that of adversarial training~\citep{43405} (see Section~\ref{ADV_TRAINING}).

This method was found to be effective against FGSM~\citep{43405} (see Section~\ref{FGSM}), DeepFool~\citep{DBLP:journals/corr/Moosavi-Dezfooli15} (see Section~\ref{DEEPFOOL}), and BIM~\citep{DBLP:journals/corr/KurakinGB16,45816} (see Section~\ref{BIM}) when evaluated on CIFAR10 and 10-class Imagenet (i.e., a subset of ImageNet dataset that only contains 10 classes). They also found that placing the detector network at different layer depths gives different results for different types of attacks. Finally,~\citet{metzen2017detecting} noted that fooling a classifier that includes a detector network is harder than an unprotected classifier, since an attacker needs to find adversaries that fool \emph{both} the classifier and detector.

However,~\citet{DBLP:journals/corr/CarliniW17} found that this method has relatively high false positive rate against stronger attacks like the C\&W attacks~\citep{DBLP:journals/corr/CarliniW16a} (see Section~\ref{CW}), and can be circumvented with SBA~\citep{DBLP:journals/corr/PapernotMGJCS16} (see Section~\ref{SUBSTITUTE}). Furthermore, although~\citet{metzen2017detecting} claimed that they did not have to perform extensive tuning to train the detector,~\citet{DBLP:journals/corr/CarliniW17} found the opposite.


\citet{DBLP:journals/corr/GongWK17} proposed a similar method to~\citet{metzen2017detecting} by training a binary classifier network to differentiate between adversarial and clean examples. However, the binary classifier here is a completely separate network from the main classifier. Rather than generating adversarial examples against the detector, they generate adversarial examples for a pretrained classifier, and add these adversarial examples to the original training data to train the binary classifier.

Although the method is relatively simple to implement,~\citet{DBLP:journals/corr/GongWK17} observed several limitations on the generalization of this method. First, they found that the detector network is sensitive to the $\epsilon$ value used to generate FGSM and BIM adversaries, in the sense that detector trained on adversarial examples with $\epsilon_1$ cannot detect adversarial examples with $\epsilon_2$, especially when $\epsilon_2 < \epsilon_1$. They also discovered that training a detector on a single type of adversaries is not guaranteed to generalize to other types of adversaries. For example, a detector trained on FGSM adversaries was found to not be able to detect JSMA adversaries. Furthermore,~\citet{DBLP:journals/corr/CarliniW17} found that this method has high false positive rates when tested on CIFAR10 models, and is vulnerable to C\&W attacks.

Another similar method by~\citet{DBLP:journals/corr/GrosseMP0M17} is a detection method that works by augmenting a classifier network with an additional class node that represents adversarial class. Given a pretrained model, a new model with an extra class node is trained on clean examples and adversarial examples generated for the pretrained model.

\citet{DBLP:journals/corr/GrosseMP0M17} showed how this method can be used to detect adversaries generated by FGSM~\citep{43405} (see Section~\ref{FGSM}) and JSMA~\citep{DBLP:journals/corr/PapernotMJFCS15} (see Section~\ref{JSMA}) on MNIST. They also showed how this method can reliably detect SBA~\citep{DBLP:journals/corr/PapernotMGJCS16} (see Section~\ref{SUBSTITUTE}), particularly FGSM-SBA and JSMA-SBA. Although the method was shown to work on MNIST~\citep{726791},~\citet{DBLP:journals/corr/CarliniW17} showed that this method does not work well on more complex datasets such as CIFAR10 as the false positive rate was too high. Furthermore,~\citet{DBLP:journals/corr/CarliniW17} also found that this method is vulnerable to the C\&W attacks in both whitebox and blackbox settings (i.e., C\&W-SBA).

Concurrent work from~\citet{DBLP:journals/corr/HosseiniCKZP17} proposed a very similar method with slight difference in the detail of the training procedure where the classifier is alternatively trained on clean and adversarial examples (i.e., via adversarial training~\citep{43405}) (see Section~\ref{ADV_TRAINING}). Furthermore, the labels used for training the model are carefully assigned by performing label smoothing~\citep{44903}. Label smoothing sets a probability value to the correct class and distributes the rest uniformly to the other classes.

Having a slightly different goal than~\citet{DBLP:journals/corr/GrosseMP0M17},~\citet{DBLP:journals/corr/HosseiniCKZP17} evaluated their method in the blackbox settings and showed how their method is especially helpful to reduce the transferability rate of adversarial examples. Although the detection method proposed by~\citet{DBLP:journals/corr/HosseiniCKZP17} was not evaluated by~\citet{DBLP:journals/corr/CarliniW17}, the method appears to be similar with the method proposed by~\citet{DBLP:journals/corr/GrosseMP0M17}. Since the evaluation of this method was only done on MNIST and grayscaled GTSRB~\citep{Stallkamp2012}, it is questionable where or not this method will also exhibit high false positive rate when tested on CIFAR10 or other datasets with higher complexity. Finally, this method needs to be evaluated against the C\&W attacks, considering the findings in~\citet{DBLP:journals/corr/CarliniW17} that shows how C\&W attacks can be used to circumvent the detection method proposed in~\citet{DBLP:journals/corr/GrosseMP0M17}.

\subsubsection{Kernel Density and Bayesian Uncertainty Estimates} \label{KERNEL_BAYESIAN} 


Assuming that adversarial examples do not lie inside the non-adversarial data manifold,~\citet{DBLP:journals/corr/FeinmanCSG17} proposed two methods to perform adversarial detection: Kernel Density and Bayesian Uncertainty Estimates. The objective for using Kernel Density Estimates (KDE) is to identify whether a data point is far from a class manifold, while Bayesian Uncertainty Estimates (BUE) can be used to detect data points that are near the low-confidence region in which KDE is not effective.

The density estimate $KDE$ for an input $x$ is calculated by 

\begin{equation} \label{equations:kernel_density}
KDE(x) = \frac{1}{|X_t|} \sum_{x_i \in X_t}e^{\frac{-||Z(x) - Z(x_i)||^2}{\sigma^2}},
\end{equation}

\noindent where $Z(x)$ denotes the logits vector given $x$ as the input, $X_t$ is a set of data of class $t$, and $\sigma$ is the kernel bandwidth. Note that the kernel function used is the Gaussian kernel and evaluated in the logits space instead of the input space. The use of the logits is inspired from the work by~\citet{pmlr-v28-bengio13} and~\citet{DBLP:journals/corr/GardnerKLUWH15}, which showed how the manifold learned by the network becomes increasingly linear and flatter which makes the logit space easier to work with than the input space. An adversarial example $x'$ will have low KDE value if $x'$ is far from the target class manifold, and thus can be detected using threshold based approach. In other words, $x$ is considered adversarial if $KDE(x)$ is less than a certain threshold value. However,~\citet{DBLP:journals/corr/FeinmanCSG17} suggested that this method does not work well when $x'$ is near the target class manifold.

\citet{DBLP:journals/corr/FeinmanCSG17} followed the work from~\citet{Gal:2016:DBA:3045390.3045502} by using dropout~\citep{JMLR:v15:srivastava14a} at inference time as a way to measure uncertainty. They quantify the uncertainty $U(x)$ of the network by performing $N$ stochastic passes by applying dropout during inference. Formally, $U(x)$ is defined as:

\begin{equation} \label{equations:bayesian_uncertainty}
U(x) = \frac{1}{N} \sum_{i=1}^N f(x)_i^T f(x)_i - \bigg(\frac{1}{N}\sum_{i=1}^N f(x)_i\bigg)^T \bigg(\frac{1}{N}\sum_{i=1}^N f(x)_i\bigg),
\end{equation}

\noindent where $f(x)_i$ denotes the $i$-th stochastic prediction. Intuitively, the uncertainty measure should be large when dealing with adversarial inputs.

\citet{DBLP:journals/corr/FeinmanCSG17} showed how the uncertainty distributions are different between adversarial and normal examples. This was indicated by the ratio $\frac{U(x')}{U(x)}$ that is larger than 1 in most cases when evaluated on MNIST and CIFAR10 against FGSM~\citep{43405} (see Section~\ref{FGSM}), BIM~\citep{DBLP:journals/corr/KurakinGB16,45816} (see Section~\ref{BIM}), JSMA~\citep{DBLP:journals/corr/PapernotMJFCS15} (see Section~\ref{JSMA}), and C\&W~\citep{DBLP:journals/corr/CarliniW16a} (see Section~\ref{CW}) attacks. KDE was also shown to be able to detect adversaries, but especially indicative when evaluated against a variant of BIM that stops the attack process following a fixed number of iterations. This is indicated by the ratio $\frac{KDE(x')}{KDE(x)}$ that are less than 1 in most cases. 

Based on these findings, the authors proposed a combined threshold based detection method, using both metrics to perform adversarial detection. This is done by putting threshold values on the uncertainty measurement and on the negative log of the kernel density estimates of a sample. The combined method was shown to be superior against FGSM, BIM, JSMA, and C\&W on MNIST, CIFAR10, and SVHN.

\citet{DBLP:journals/corr/CarliniW17} evaluated this method and concluded that the KDE method alone does not work well on CIFAR10 and can be circumvented with modified C\&W attacks by including an additional objective term defined by $\textrm{max}(-\log(KDE(x'))-\epsilon,0)$ both in whitebox and blackbox settings using SBA~\citep{DBLP:journals/corr/PapernotMGJCS16} (see Section~\ref{SUBSTITUTE}). They also showed that BUE can be circumvented by C\&W attacks on the expectations of different models sampled during dropout. However, they noted that the distortions required to generate adversarial examples that fool BUE are also larger compared to other detection methods that they evaluated. As a result,~\citet{DBLP:journals/corr/CarliniW17} concluded that BUE was the hardest detection method to fool compared to the other methods. It is also relatively straightforward to implement as an add-on to an existing network architecture.

\subsubsection{Feature Squeezing} \label{SQUEEZE} 

\begin{figure}[h]
\centering
\includegraphics[width=0.6\textwidth]{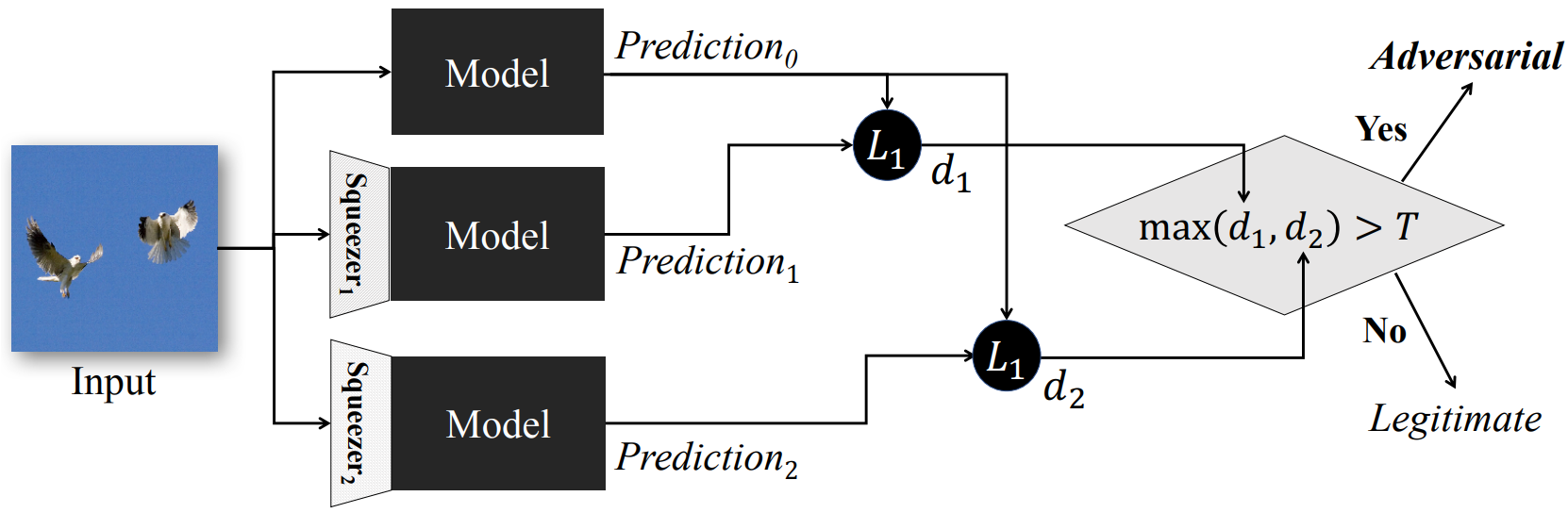}
\caption{Illustration of feature squeezing (taken from~\citet{DBLP:journals/corr/XuEQ17}).}
\label{fig:feature_squeezing}
\end{figure}

\citet{DBLP:journals/corr/XuEQ17} argued that the dimensionality of input features is often  unnecessarily large, yielding a large attack surface. They proposed a detection strategy in which they compare the predictions between squeezed and unsqueezed inputs. As the name suggests, the goal of feature squeezing is to remove unnecessary features from an input. Two feature squeezing methods were evaluated: color bit-depth reduction and spatial smoothing, both with local and non-local smoothing. The input is labelled as adversarial if the $L_1$ difference between the model's prediction on squeezed and unsqueezed inputs is larger than a certain threshold value $T$. Note that this method is independent of the defended model, and thus can be used together with other defense techniques. Figure~\ref{fig:feature_squeezing} illustrates this method.

For color depth reduction,~\citet{DBLP:journals/corr/XuEQ17} experimented with various $i$-bit depth where $1 \leq i \leq 8$. For local smoothing, median filtering was found to be more effective than Gaussian and mean smoothing. In the case of non-local smoothing, non-local means denoising~\citep{ipol.2011.bcm_nlm} was used. There is no strong motivation for this choice of squeezers; future work on the most effective squeezers is still needed. They also suggested using multiple feature squeezing methods as part of the detection model, since different squeezers are more effective on certain types of adversaries than others. Given $n$ squeezers $\{s_1,...,s_n\}$, the output of the detector is defined as

\begin{equation} \label{equations:feature_squeeze}
  Detector(x) =
  \begin{cases}
  \textrm{Adversarial} & \text{if $\max \Big(||f(x) - f(s_1(x))||_1,..., ||f(x) - f(s_n(x))||_1 \Big) > T$} \\
  \textrm{Benign} & \text{otherwise}
  \end{cases},
\end{equation}

\noindent where $f(x)$ is the output vector given by the softmax layer of a deep neural network.

\citet{DBLP:journals/corr/XuEQ17} demonstrated that feature squeezing can also be used to increase adversarial robustness. This is similar to the work from~\citet{guo2018countering} which also attempted to enhance adversarial robustness through various input transformations (see Section~\ref{TRANSFORM-DEFENSE}). Furthermore,~\citet{DBLP:journals/corr/XuEQ17} showed that the robustness can be improved by coupling feature squeezing with adversarial training~\citep{43405} (see Section~\ref{ADV_TRAINING}). However, they acknowledged that feature squeezing alone is not a reliable defense, and thus focus on adversarial detection instead.

Feature squeezing was shown to be able to detect adversarial examples generated by FGSM~\citep{43405} (see Section~\ref{FGSM}), BIM~\citep{DBLP:journals/corr/KurakinGB16,45816} (see Section~\ref{BIM}), DeepFool~\citep{DBLP:journals/corr/Moosavi-Dezfooli15} (see Section~\ref{DEEPFOOL}), JSMA~\citep{DBLP:journals/corr/PapernotMJFCS15} (see Section~\ref{JSMA}), and C\&W~\citep{DBLP:journals/corr/CarliniW16a} (see Section~\ref{CW}) attacks on MNIST, CIFAR10, and ImageNet models when the attacker is not aware of the defense strategy used. To be more specific,~\citet{DBLP:journals/corr/XuEQ17} found that color depth reduction is especially effective against $L_2$ and $L_{\infty}$ attacks, but less effective against $L_0$ attacks. On the other hand, spatial smoothing method was found to be effective against $L_0$ attack. Intuitively, $L_0$ attacks cause perturbations that can be thought of as salt-and-pepper noise, in which median filtering has often been used to remove this type of noise. These findings support the need of using joint squeezers configuration to take into account different types of adversarial examples.

\citet{DBLP:journals/corr/XuEQ17} also considered a case where the attacker is aware of the detection method. In this setting, the adversary needs to find adversarial examples that fool the classifier and minimize the difference in $L_1$ score between squeezed and unsqueezed inputs. The defense can then be circumvented by iterative optimization based attacks, such as C\&W attacks, by folding in the $L_1$ constraint. As a result, they suggested to introduce randomness, for example by randomizing the threshold value such that $T'=T \pm rand(0.1)$, in order to make it harder for the attacker to predict the behavior of the squeezers. Although feature squeezing method is powerful against static adversaries,~\citet{DBLP:journals/corr/XuEQ17} acknowledged that this method is possibly not immune to adversarial examples in situation where an attacker has full knowledge of the squeezers.

\subsubsection{Reverse Cross-Entropy} \label{RCE} 

\citet{DBLP:journals/corr/PangDZ17} proposed a detection method by introducing a new objective function called the Reverse Cross-Entropy (RCE). For $C$ classes, RCE loss is defined as

\begin{equation} \label{equations:rce}
\mathcal{L}_{RCE} = -y_r \log f(x), 
\end{equation}

\noindent where $y_r$ is the reversed label defined as

\begin{equation} \label{equations:reverse_label}
  \ y_{r_{(i)}} =
  \begin{cases}
  0 & \text{if \ $i = y$ \ (true label)} \\
  \frac{1}{C-1} & \text{otherwise}
  \end{cases}.
\end{equation}



\noindent They noted that training a model to minimize $\mathcal{L}_{RCE}$ produces a reverse classifier where the probability assigned to the correct class will be the lowest. Thus,~\citet{DBLP:journals/corr/PangDZ17} suggested to negate the logits such that $f(x) = softmax(-Z(x))$.


Once the model is trained, one may perform adversarial detection using threshold based detection.~\citet{DBLP:journals/corr/PangDZ17} performed the detection by computing the Kernel Density Estimate (KDE)~\citep{DBLP:journals/corr/FeinmanCSG17} (see Section~\ref{KERNEL_BAYESIAN}). They also introduced a new metric called the non-maximum element or non-ME defined as

\begin{equation} \label{equations:non_me}
\textrm{non-ME}(x) = -\sum_{i \neq \hat{y}} \hat{f}(x)_i \log (\hat{f}(x)_i), 
\end{equation}

\noindent where $\hat{f}(x)_i$ denotes the normalized non-maximum elements in $f(x)$. The detection is performed by setting a threshold value $T$. If the chosen metric (e.g., KDE or non-ME) is below the threshold value, the input is considered as adversarial. However,~\citet{DBLP:journals/corr/PangDZ17} found that detection using KDE outperforms non-ME in most cases.

This method was shown to be reliable when evaluated on MNIST and CIFAR10 datasets against FGSM \citep{43405} (see Section~\ref{FGSM}), BIM/ILLCM~\citep{DBLP:journals/corr/KurakinGB16, 45816} (see Section~\ref{BIM}), JSMA~\citep{DBLP:journals/corr/PapernotMJFCS15} (see Section~\ref{JSMA}), C\&W~\citep{DBLP:journals/corr/CarliniW16a} (see Section~\ref{CW}), and modified C\&W that takes into account the use of KDE; we refer this as MCW. Furthermore, this method was also evaluated in blackbox settings against MCW-SBA~\citep{DBLP:journals/corr/PapernotMGJCS16} (see Section~\ref{SUBSTITUTE}) and showed how the adversarial examples generated by SBA suffer from poor transferability. Even when adversarial examples that fool both detector and classifier models were found, the adversarial examples usually exhibit larger distortions and thus are visually visible to human.

Finally, the use of RCE not only allows one to perform adversarial detection, but also increases robustness of a model in general compared to the use of standard cross-entropy as the objective function. They also showed the comparison of t-SNE embeddings between a model trained with RCE and non-RCE objectives, and showed how the t-SNE~\citep{vanDerMaaten2008} visualization of the model that was trained with RCE objective achieved higher separability. The use of new objective functions such as RCE which constrains the ability of attacker to generate adversarial examples seems to be an interesting research area to be explored.

\pagebreak

\section{Transferability of Adversarial Examples} \label{TRANSFER} 



\subsection{What is transferability?}

Transferability is a phenomenon where adversarial examples or perturbations generated to fool a particular model can be used to fool others. Many machine learning models, including deep networks, are subject to this phenomenon~\citep{42503,DBLP:journals/corr/PapernotMG16,DBLP:journals/corr/PapernotMGJCS16}. Transferability can be carved into several different categories: (1) transferability between different models (e.g., between VGG16~\citep{Simonyan14c} and ResNet~\citep{He2016DeepRL} in the case of deep neural networks) trained on the same dataset or different subsets of the same dataset, (2) transferability between different machine learning techniques (e.g., between SVMs and deep neural networks), and (3) transferability between models that perform different tasks (e.g., between semantic image segmentation and object detection models~\citep{DBLP:journals/corr/XieWZZXY17}). Understanding the transferability phenomenon is critical to creating safer machine learning models. Several adversarial attacks like the SBA~\citep{DBLP:journals/corr/PapernotMGJCS16} (see Section~\ref{SUBSTITUTE}) work by exploiting this phenomenon. In fact, the transferability property has also been exploited to enable adversarial attacks on real world image classification~\citep{DBLP:journals/corr/PapernotMG16,DBLP:journals/corr/PapernotMGJCS16} and detection~\citep{zhang2018camou} systems.

\subsection{Where does transferability come from?}

\citet{43405} suggested that the reason why adversarial examples can fool deep neural networks is due to the linearity behavior when using activation functions like the Rectified Linear Units (ReLU)~\citep{pmlr-v15-glorot11a} or Maxout~\citep{Goodfellow:2013:MN:3042817.3043084}. Based on this view,~\citet{43405} argued that adversarial examples exist in the ``contiguous regions of the 1-D subspace defined by the fast gradient sign method'' which was contrary to the belief that adversarial examples lie in the low probability regions in the data manifold~\citep{42503}. This view suggests that there exist many adversarial examples which have higher chance to be transferable because different models will end up learning similar functions, and that the direction of perturbations is the key factor that allows transferable adversarial examples.

\citet{DBLP:journals/corr/TramerPGBM17} studied the subspace dimensionality of adversarial examples and asked if there is an intersection between the adversarial subspace of two different models. They introduced a technique called Gradient Aligned Adversarial Subspace (GAAS) to estimate the number of orthogonal adversarial directions. They found that there were many orthogonal adversarial directions, a large percentage of which were transferable. Furthermore, they found that the separating distance between the decision boundaries of two models was on average smaller than the distances between any inputs to the decision boundaries, even on adversarially trained models. This finding suggests that the adversarial subspaces of different models overlap, which is in agreement with the findings from~\citet{DBLP:journals/corr/LiuCLS16}, explaining why many adversarial examples are transferable between models.

\citet{tsipras2018robustness} investigated the property of robust classifier and found that robust classifiers tend to only take advantage of robust features (i.e., features that have high correlation with the ground truth label), as opposed to weak features (i.e., features that have low correlation with the ground truth label) during training. Based on this,~\citet{tsipras2018robustness} argue that classifiers that rely on weak features will suffer from the same adversarial examples that were created by perturbing these features. This hypothesis was also reinforced by~\citet{ilyas2019adversarial} that shows how different models tend to exploit similar \textit{non-robust features} in a dataset, and thus adversaries generated for one model can be transferred to the other models.

\subsection{What affects transferability?}

There are several dimensions that appear to affect transferability. Firstly, model type in turn influences the type of transferability observed.~\citet{DBLP:journals/corr/PapernotMG16} found empirically that deep neural networks and kNN are more robust to cross-technique transferability, but more vulnerable to intra-technique transferability. On the other hand, linear regression, SVM, decision trees, and ensemble methods are more robust to intra-technique transferability, but more vulnerable to cross-technique transferability.

Secondly, attack strength matters.~\citet{45816} found that stronger adversarial examples that can penetrate through a robustly defended model are less transferable to other models, while adversarial examples generated to attack undefended model are more transferable. They argued that adversarial examples generated to penetrate specific defense method may have ``overfitted'' to fool that particular model.~\citet{45816} also found that larger magnitude perturbations often increases the transferability rate.

Thirdly, non-targeted attacks transfer better than targeted ones.~\citet{DBLP:journals/corr/LiuCLS16} investigated the transferability property on large scale datasets like ImageNet and found that transferable non-targeted adversarial examples are easier to find than targeted ones. They further showed that decision boundaries of different models align with each other, hinting why adversarial examples are transferable. They found that the direction of the gradient of the loss function with respect to the input of two different models are orthogonal, and thus that fast gradient based approaches are unlikely to find targeted adversarial examples that transfer. To overcome this, they perform ensemble adversarial attack, successfully generating targeted adversaries that fool several different models.


\pagebreak

\section{Conclusion and Future Work} \label{CONCLUSION} 


We have seen various ways to generate adversarial examples in both whitebox and blackbox settings. While whitebox attack is especially important for evaluating new defenses in the worst-case scenario, works in blackbox attack are as important since they involve more realistic threat model that may happen in the real world. We also discussed several methods to defend against adversarial examples either by trying to robustify a model, or identifying adversarial samples so they can be rejected. However, the majority of these defenses boil down to some form of gradient masking; hiding or disrupting the gradients available to an attacker. The work from~\citet{obfuscated-gradients} dramatically demonstrated the fragility of such approaches. One of the interesting directions to explore is to investigate the effectiveness of Bayesian inference as an adversarial defense method. It may also be interesting to explore other existing algorithms to perform both adversarial attack and defense. For example, one may further explore the applications of reinforcement learning, meta-learning, and genetic algorithms in the field of adversarial machine learning.

Recent work aiming to improve resistance to adversarial attacks resulted in more generally robust models, which learned features that were immediately intuitive to human users~\cite{tsipras2018robustness,etmann2019connection,engstrom2019learning}.~\citet{ilyas2019adversarial} showed how machine learning models can be trained to be robust by just relying on some \textit{robust features}, while models that are trained on \textit{non-robust features} are highly vulnerable to adversarial examples. On somewhat similar notes,~\citet{geirhos2018imagenettrained} found that convolutional neural nets are biased towards textures, and how increasing the bias towards shape may improve the robustness of the model. A follow up work by~\citet{pmlr-v97-zhang19s} further showed that adversarially trained model is less biased towards textures and more towards shapes or edges. This emphasizes the fruitfulness of asking why machine learning models fail, and exploring those failure modes.


Although there have been many different defense strategies proposed, many of these defense methods have been found to be vulnerable to specific type of attacks, and raises a concern on whether these defenses were evaluated unfairly~\citep{DBLP:journals/corr/CarliniW17,DBLP:journals/corr/abs-1711-08478,obfuscated-gradients,pmlr-v80-uesato18a} which may have led to false robustness. Gradient masking and obfuscated gradients seems to be phenomenons that many defense methods often rely, \emph{intentionally or not}.~\citet{obfuscated-gradients} suggested several tips to identify gradient masking: 1) if single step attacks perform better than iterative attacks, 2) if blackbox attacks perform better than whitebox attacks, 3) if unbounded perturbations do not reach 100\% attack success rate, and 4) if random sampling successfully finds adversarial examples while gradient based attacks do not. 

We have seen the difficulty of correctly evaluating robustness of defense methods. New defenses not only need to be evaluated against existing strongest adversaries, but also need to be evaluated on adaptive attacks (i.e., evaluated with the assumption that an attacker is aware of the defense strategy used and will seek to overcome it). For example, if the defense strategy aims to perform non-differentiable operation or gradient minimization, then the defense needs to be evaluated against attacks that do not rely on calculating the exact gradient such as BPDA~\citep{obfuscated-gradients} (see Section~\ref{BPDA}) or SPSA~\citep{pmlr-v80-uesato18a} (see Section~\ref{SPSA}). Defenses also need to be evaluated against multiple types of adversaries (e.g., $L_{\infty}$, $L_{0}$, and $L_{2}$ and other types of adversaries) as defense such as adversarial training have been shown to only be robust against certain types of adversaries~\citep{tramèr2018ensemble,Schott2018a,tramer2019adversarial,kang2019transfer}. Furthermore, the vast majority of defenses are limited to empirical evaluations and do not offer robustness guarantee to unknown attacks. A future direction is to investigate properties that can theoretically guarantee robustness to adversarial attacks, similar to the WRM~\citep{sinha2018certifiable} (see Section~\ref{WRM}), coupled with empirical evaluations. It is important to note that these methods typically offer guarantee robustness only when the difference between the input and its adversarial counterpart is small under a certain metric. However, our understanding in perceptual similarity metric is still very limited. Common perceptual similarity metric such as the mean squared error is definitely not an ideal one~\citep{4775883}. As we discussed in Section~\ref{STA}, attacks like stAdv~\citep{xiao2018spatially} may still be able to penetrate through defenses that focus only on defending against $L_p$ adversaries.~\citet{carlini2019evaluating} discuss about how to evaluate robustness and some common pitfalls in robustness evaluation in more detail, thus we refer readers to their work for a more comprehensive discussion.

We also discussed about the transferability property, where adversarial examples can be transferred between different machine learning models, different machine learning techniques, and different tasks. In the case of neural networks, future research that evaluates the relation between memorization of deep networks and transferability property will be interesting. Understanding this phenomena may lead us to better understand the inner workings of existing machine learning models.

As machine learning exerts a greater influence in the real world, so to will adversarial attacks. With an increasing number of commercial machine learning systems, such as face recognition systems, autonomous driving, and medical imaging comes an increasing vulnerability to malevolent actors. Whilst many defense strategies have been proposed, most of the existing defenses focus on imperceptible perturbations. As discussed in Section~\ref{PHYSICAL}, some of the physically realizable attacks are in the form of visible but inconspicuous perturbations. To the best of our knowledge, there are no defense strategies that combat these types of adversaries as yet. Future works on methods to generate physically realizable adversaries and how to defend against them are thus needed. Moving forward, as training on simulated data becoming popular, a fully differentiable renderer that produces realistic looking images similar to the work done by~\citet{liu2018beyond}, can be an excellent tool to evaluate the robustness of a model.

Finally, we want to re-emphasize that this paper focuses mainly on the topic of adversarial examples in the computer vision domain. Adversarial examples already exist in other domains such as in computer security~\citep{10.1007/978-3-319-66399-9_4}, speech recognition~\citep{DBLP:journals/corr/abs-1801-01944,DBLP:journals/corr/abs-1801-00554,DBLP:journals/corr/abs-1801-03339}, and text~\citep{zhao2018generating}. Continuing in these endeavors are important for the safety of machine learning implementations. We leave these as future works.

\acknowledgments 

\pagebreak

\bibliographystyle{IEEEtranN}

\bibliography{references}



\end{document}